\documentclass[journal]{IEEEtran}

\usepackage[nocompress]{cite}
\usepackage{amsmath,amssymb}
\usepackage{graphicx}
\usepackage{booktabs}
\usepackage{tabularx}
\usepackage{array}
\usepackage{threeparttable}
\usepackage{placeins}
\usepackage[hidelinks]{hyperref}
\newcolumntype{Y}{>{\raggedright\arraybackslash}X}

\graphicspath{{figures/}}

\newcommand{\appxsec}[1]{~(Appendix~\ref{#1})}
\newif\ifarxivbuild
\arxivbuildtrue
\providecommand{\ket}[1]{\lvert #1 \rangle}
\usepackage[table]{xcolor}
\usepackage{tikz}
\usetikzlibrary{positioning,calc,arrows.meta,fit}
\definecolor{qgClassical}{HTML}{0072B2}
\definecolor{qgClassicalTint}{HTML}{D6E8F3}
\definecolor{qgQuantum}{HTML}{CC79A7}
\definecolor{qgQuantumTint}{HTML}{F7EAF1}
\definecolor{qgAttack}{HTML}{D55E00}
\definecolor{qgAttackTint}{HTML}{F8E5D6}
\definecolor{qgChoice}{HTML}{E69F00}
\definecolor{qgChoiceTint}{HTML}{FBF0D6}
\definecolor{qgControl}{HTML}{009E73}
\definecolor{qgControlTint}{HTML}{D6EFE9}
\definecolor{qgReference}{HTML}{7F7F7F}
\definecolor{qgReferenceTint}{HTML}{EBEBEB}
\def\dline{0.3pt}

\definecolor{dgrey}{gray}{0.30}
\tikzset{
  dbox/.style={draw=black!70, line width=\dline, align=left, inner xsep=3pt,
    inner ysep=1.6pt, font=\footnotesize, anchor=north west},
  mclassical/.style={draw=qgClassical, fill=qgClassicalTint, line width=\dline},
  mquantum/.style={draw=qgQuantum, fill=qgQuantumTint, line width=\dline},
  mattack/.style={draw=qgAttack, fill=qgAttackTint, line width=\dline},
  mchoice/.style={draw=qgChoice, fill=qgChoiceTint, line width=\dline},
  mcontrol/.style={draw=qgControl, fill=qgControlTint, line width=\dline},
  mreference/.style={draw=qgReference, fill=qgReferenceTint, line width=\dline},
  cbar/.style={append after command={\pgfextra{\let\cbarnode\tikzlastnode}
    ([xshift=-0.75pt,yshift=-0.15pt]\cbarnode.north east) edge[qgControl, line width=1.5pt]
    ([xshift=-0.75pt,yshift=0.15pt]\cbarnode.south east)}},
  darrow/.style={-{Stealth[length=1.5mm,width=1.3mm]}, line width=0.5pt},
  dhead/.style={font=\footnotesize\bfseries, anchor=south west, inner sep=0pt},
  dnote/.style={font=\footnotesize, text=dgrey, anchor=south west, inner sep=0pt},
}
\newcommand{\dt}[2]{{\bfseries #1}\\[0.3pt]{\color{dgrey}#2}}
\newcommand{\mdots}[2]{{\color{#1}\foreach \i in {1,...,#2}{\textbullet}}}
\newcommand{\pbar}[2]{\tikz[baseline=-0.6ex]{\def\pbarx{0}\foreach \s/\c in {#2}{%
  \fill[\c] (\pbarx/100*#1,-0.9ex) rectangle ++(\s/100*#1-0.4pt,1.5ex);
  \pgfmathsetmacro{\nx}{\pbarx+\s}\global\let\pbarx\nx}}}
\newcommand{\mkey}[2]{\tikz[baseline=-0.5ex]\node[#1, minimum width=2.4mm, minimum height=2.4mm,
  inner sep=0pt, draw]{};~#2}

\newcommand{\NumRowsBinary}{78\,368}
\newcommand{\NumTrainRows}{47\,020}
\newcommand{\NumAttackShare}{71}
\newcommand{\NumNaturalShare}{29}
\newcommand{\NumGasRows}{38\,054}
\newcommand{\NumGasRawRows}{274\,628}
\newcommand{\NumGasFeatures}{16}
\newcommand{\NumGasAttackShare}{54}
\newcommand{\NumGasBestClassical}{0.761}
\newcommand{\NumGasBestQuantum}{0.748}
\newcommand{\NumGridSmallFeatures}{82}
\newcommand{\NumGridSmallBestClassical}{0.996}
\newcommand{\NumGridSmallBestQuantum}{0.988}
\newcommand{\NumGridLargeFeatures}{726}
\newcommand{\NumGridLargeBestClassical}{0.997}
\newcommand{\NumGridLargeBestQuantum}{0.993}
\newcommand{\NumGridRows}{5760}
\newcommand{\NumGridRowsGenerated}{8640}
\newcommand{\NumGridAttackShare}{50}

\renewcommand{\IEEEiedlistdecl}{\setlength{\IEEElabelindent}{0pt}}

\begin{document}

\title{Benchmarking Quantum Machine Learning for Power-System Attack Detection:\\
Evaluation Choices Can Decide the Outcome}

\author{Md~Rezwanul~Islam and ASM~Shihavuddin%
\thanks{(Corresponding author: Md Rezwanul Islam.)}%
\thanks{M. R. Islam is an independent researcher, Dhaka 1207, Bangladesh
(e-mail: rznislam@gmail.com).}%
\thanks{ASM Shihavuddin is with the Department of Electrical and Electronic
Engineering, University of Liberal Arts Bangladesh, Dhaka 1207, Bangladesh
(e-mail: shihav.uddin@ulab.edu.bd).}}

\maketitle

\begin{abstract}
Smart grids rely on machine-learning detectors to spot cyberattacks such as false data
injection, and quantum machine learning has been proposed for this task. No study we could
read tests quantum and classical detectors under white-box, transfer, and black-box attacks
and random label flips in one protocol. We compare simulated quantum-kernel support vector
machines and variational quantum classifiers with six tuned classical models on the public
Mississippi State University and Oak Ridge National Laboratory power-system attack dataset,
under equal tuning budgets. Our main finding is that evaluation choices can decide the outcome.
On this dataset, seven choices in how models are tested each changed a reported result
while the model families stayed fixed. Row-level
splits, used by the studies we checked, put near-copies of test rows in training: a tuned
tree ensemble scores 0.905 macro-averaged F1 on them and 0.594 on held-out files. On held-out files at matched dimensionality, a tuned classical and a tuned quantum model clear chance by little. A poorly fitted surrogate
makes attacks transfer over four times more strongly one way than the other, and on this
dataset a correct fit removes the difference. Under its own white-box attack, the quantum kernel's macro-averaged F1 falls from 0.56 to 0.01
(chance is 0.50) where a transferred attack in the same run left 0.52;
tree models admit
no gradient-based white-box attack, so their figures are upper bounds. These attack results are for
eight-dimension detectors trained under the row-level split. We repeated the main checks on a gas-pipeline
control-system dataset and on two simulated transmission grids. No quantum model we ran is more
accurate or consistently more robust on any of them, and the split
inflation is largest on the main dataset, where holding out whole files leaves few near-copies
of a test row in training. Of the seven choices, three reproduce on the gas data (two within noise) and two in part, and none of the four checked on the simulated grids reproduces in full. On the gas data the main study's standardization maps every ordinary
pressure reading to one value. With that feature log-scaled and every model retuned, six
of the eight matched detectors gain 0.17 to 0.20
macro-averaged F1; the gas counts above use this fix. On the simulated grids, the grids' own bad-data test flags every row perturbed
without respecting the power-flow physics, random noise included. Benchmarks should hold out whole files, give every model its own attack where
possible, and control for random noise. We release the code and a seven-item control
checklist.
\end{abstract}

\begin{IEEEkeywords}
Adversarial robustness, false data injection, quantum machine learning, smart grid
security.
\end{IEEEkeywords}

\section{Introduction}

\IEEEPARstart{S}{tate} estimation and protection logic in modern power systems consume streams of phasor
measurement unit (PMU), relay, and log data. False data injection attacks (FDIA) corrupt
those streams and can bypass bad-data detection entirely~\cite{liu2011false}. Machine
learning (ML) detectors are the standard response. Yet a detector is itself a model
exposed to adversarial inputs: an attacker who can shape measurements to evade state
estimation can shape them to evade the detector too.
\begin{figure}[!t]
\centering
\includegraphics[width=\columnwidth]{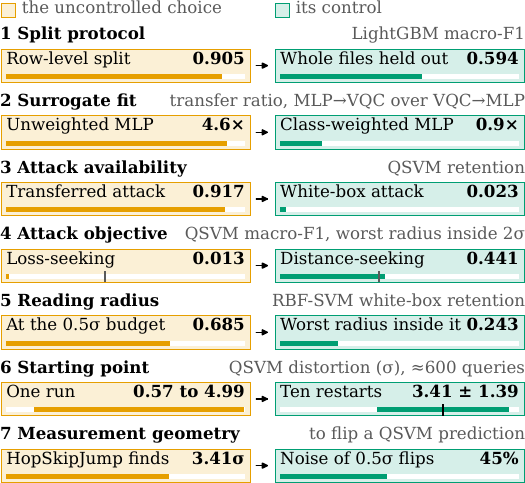}
\caption{Evaluation checklist. Each of seven choices changed a reported result on the
MSU/ORNL binary task with the model families fixed. Left: the reading without its control;
right (green): the control, which a benchmark should also run. QSVM: quantum-kernel support vector machine;
retention: the share of its clean macro-F1 a model keeps under attack (for the QSVM, 0.89 is chance); transfer ratio: the
macro-F1 that attacks crafted on the \ifarxivbuild multilayer perceptron (MLP) take from the
variational quantum classifier (VQC)\else MLP take from the VQC\fi, over the reverse; $\sigma$: one
training-split standard deviation. Row 4: the tick is the constant-predictor floor, the score
of a detector that labels every row an attack. The first attack drives the QSVM far below it;
the second leaves it just above\ifarxivbuild, with 96\% of rows given one label\fi. Rows 6 and 7: HopSkipJump
at about 600 queries per point; in row 7, noise flips rows that the attack places beyond $0.5\sigma$.}
\label{fig:checklist}
\end{figure}

\ifarxivbuild\begin{figure*}[!t]
\centering
\begin{tikzpicture}[node distance=1.6mm]
\pgfmathsetlengthmacro{\C}{(\textwidth-20mm-1pt)/6}
\pgfmathsetlengthmacro{\T}{\C-7.4pt}
\coordinate (k1) at (0,0);
\foreach \i [count=\j from 2] in {1,...,5} \coordinate (k\j) at ($(k\i)+(\C+4mm,0)$);
\foreach \c/\n/\h in {k1/1/1. Data, k2/2/2. Split protocols, k4/2/3. Models, k6/1/4. Attacks} {
  \draw[line width=0.8pt] (\c) -- ++({\n*\C+(\n-1)*4mm},0);
  \node[dhead, anchor=base west] at ($(\c)+(0,-3.7mm)$) {\h};
}
\node[dbox, mreference, text width=\T, minimum height=17.5mm, below=6.2mm of k1, anchor=north west] (b1)
  {\dt{MSU/ORNL data}{\NumRowsBinary{} rows, 128 features, 15 files\\
  \pbar{0.96*\T}{\NumAttackShare/qgAttack, \NumNaturalShare/qgReference!55}\\
  attack \NumAttackShare\% of rows}};
\node[dbox, mchoice, text width=\T, minimum height=17.5mm, below=6.2mm of k2, anchor=north west] (b2)
  {\dt{Row-level split}{60/20/20, the common choice\\
  \pbar{0.96*\T}{60/qgChoice, 20/qgChoice!60, 20/qgChoice!30}}};
\node[dbox, mcontrol, text width=\T, minimum height=17.5mm, below=6.2mm of k3, anchor=north west] (b3)
  {\dt{Whole files held out}{the control for the leak}};
\node[dbox, mclassical, text width=\T, minimum height=17.5mm, below=6.2mm of k4, anchor=north west] (b4)
  {\dt{Six classical \mdots{qgClassical}{6}}{128 features, or PCA to 8, 12, 16}};
\node[dbox, mquantum, text width=\T, minimum height=17.5mm, below=6.2mm of k5, anchor=north west] (b5)
  {\dt{Two quantum \mdots{qgQuantum}{2}}{QSVM and VQC, one qubit per dimension}};
\node[dbox, mattack, text width=\T, minimum height=17.5mm, below=6.2mm of k6, anchor=north west] (b6)
  {\dt{Attacks and noise}{white-box where gradients exist; transfer; black-box; label flips; noise}};
\foreach \a/\b in {b1/b2, b3/b4, b5/b6}
  \draw[darrow] ($(\a.east)+(0.4mm,0)$) -- ($(\b.west)+(-0.4mm,0)$);
\node[fit=(b1)(b2)(b3)(b4)(b5)(b6), inner sep=0pt] (lowest) {};
\coordinate (lg) at ($(lowest.south west -| k1)+(0,-2.6mm)$);
\draw[line width=0.8pt] (lg) -- ++(6*\C+20mm-1.2pt,0);
\node[anchor=north west, inner sep=0pt, font=\footnotesize] at ($(lg)+(0,-1.6mm)$)
  {\textbf{Color key:}\quad \mkey{mclassical}{classical model}\quad
  \mkey{mquantum}{quantum model}\quad \mkey{mattack}{attack}\quad
  \mkey{mchoice}{evaluation choice}\quad \mkey{mcontrol}{control that checks it}\quad
  \mkey{mreference}{data, reference}};
\end{tikzpicture}
\caption{Benchmark at a glance. Every model gets the same 50 tuning trials, and the
classical models are retrained on ten seeds.
Colors keep one meaning in every exhibit (key at the bottom; the replication figure has
its own)\ifarxivbuild.
Figs.~\ref{fig:threat}, \ref{fig:data}, and~\ref{fig:models} detail the stages\else.
Table~\ref{tab:attacks} details the attacks\fi.}
\label{fig:overview}
\end{figure*}\fi

Quantum machine learning (QML) has been proposed for exactly such settings, with evidence that
quantum classifiers can resist adversarial perturbations better than classical
ones~\cite{west2023towards,west2023benchmarking}. That evidence comes almost entirely from image
data. On the public Mississippi State University (MSU) and Oak Ridge National Laboratory (ORNL)
power-system attack dataset~\cite{hink2014machine,pan2015developing}, QML work reports accuracy
or tests one defended network against gradient-based evasion~\cite{ngo2026qupid}. Other QML work
on power systems uses simulated grids~\cite{li2026qstaformer}. No study we could read combines
white-box, transfer, and decision-based black-box attacks and random label flips across quantum
and classical detectors in one protocol (Table~\ref{tab:related}).
We ask three research questions (RQs):
\begin{itemize}[\IEEEsetlabelwidth{RQ3}]
\item[RQ1] Do quantum models match tuned classical baselines on detection accuracy?
\item[RQ2] Do quantum and classical models differ in adversarial robustness and attack
transferability, and can one shared attack protocol measure the difference?
\item[RQ3] How sensitive are the quantum results to encoding, qubit count, depth, shots, and
training-set size?
\end{itemize}

Answering them raised a prior question: how much of each answer does the evaluation itself
decide? Our main finding is that evaluation choices can decide the benchmark's answers.
Fig.~\ref{fig:checklist} lists seven such choices that each changed a result on the MSU/ORNL dataset\ifarxivbuild, and Fig.~\ref{fig:overview} lays out the
benchmark\fi. Choices 2 to 7, on how each attack is built
and reported, bear on RQ2. Choice 1, the split protocol, bears on RQ1. The dataset ships as 15
source files of near-duplicate rows. On our strongest model, tuned LightGBM, the row-level split inflates the macro-averaged F1 score (macro-F1)
by 0.31 over the whole-file split. On held-out files at matched dimensionality, tuned XGBoost and
the tuned quantum kernel clear the stratified-random floor (the score of a random guess at the
class shares, about 0.50) by only $+0.032$ and $+0.023$\ifarxivbuild{} under the 2000-row cap.
Their difference is not robustly significant: it changes sign on 7 of 20 file
partitions\fi. The attack experiments keep the row-level split, where the matched models have
accuracy to lose; the kernel contrast and the noise control hold on held-out files
(Section~\ref{sec:limitations}). Our contributions:
\begin{enumerate}
\item \emph{A seven-item evaluation checklist with measured sizes} (Fig.~\ref{fig:checklist}).
Checklists for adversarial evaluation and for leakage
exist~\cite{carlini2019evaluating,kapoor2023leakage}; ours measures seven such choices on grid
attack data. Each choice changed a
reported result with the model families fixed, and each has a control that catches it. For
example, a transferred attack leaves the quantum-kernel support vector machine (QSVM) near the
stratified-random floor it started just above, so it looks robust; its own white-box attack
cuts its retention (the share of clean macro-F1 a model keeps under attack) to $0.023$.\ifarxivbuild{} Tree families admit no gradient-based white-box attack, so their retention figures are upper bounds.\fi
\item \emph{An equal-budget, seed-replicated QML accuracy benchmark} on this dataset. Under the
row-level split, neither the QSVM nor a variational quantum classifier (VQC) shows an advantage
over the best tuned classical models, and a
positive control shows that the benchmark detects a quantum win where the labels favor one.
\item \emph{A test of which checklist items reproduce}, on a gas-pipeline control-system dataset~\cite{morris2014icsdata} and on
two simulated transmission grids with false data injection
(Sections~\ref{sec:replication} and~\ref{sec:physics}),
where we also attack under the power-flow physics. Three of the seven items reproduce on the gas
data (two within noise), and none of the four checked on the grids reproduces in full.\ifarxivbuild{} The white-box
contrast between the two kernel machines reproduces on the other data at most in part.\fi
\item \emph{QuantumGridBench}, an open, seeded, configuration-driven reproducibility package in
which every reported number traces to a stored run artifact\ifarxivbuild{} (the artifacts are
not part of the release)\fi.
\end{enumerate}

\section{Related Work}\label{sec:related}

Table~\ref{tab:related} places this work against six lines of prior work. Tabular benchmarks
already answer the accuracy question on generic data, and they find that the dataset dominates
the outcome~\cite{bowles2024better,kakavand2026benchmarking}. The question must thus be asked
per domain. The evidence for a quantum robustness advantage comes from image data; we test
whether that direction holds on public grid data.

\begin{table*}[!t]
\centering
\caption{Prior work and what it leaves open}
\label{tab:related}
\footnotesize
\setlength{\tabcolsep}{4pt}
\renewcommand{\arraystretch}{1.12}
\begin{tabularx}{\textwidth}{@{}>{\raggedright\arraybackslash}p{0.17\textwidth}YY@{}}
\toprule
Line of work & What it reports & What it leaves open, and where we test it\\
\midrule
Classical detection on this
dataset~\cite{hink2014machine,tuyizere2023machine,naeem2025classification,alihodzic2026cyberphysical,farsi2026detection}
& \ifarxivbuild Accuracy or macro-F1 of 0.89 to 0.94, across differing tasks and feature sets.
Every file supplies both training and test rows: ten random folds inside each file, a split of
the pooled rows, or a split inside each file. We read these five, not a census, and reran none.
On other FDIA data, evasion and poisoning attacks degrade classical
detection~\cite{sayghe2020survey}, and recent work remains largely classical, including pipelines
that combine principal component analysis and an autoencoder~\cite{tufail2025hybrid}\else Macro-F1 or accuracy of 0.89--0.94, every file in
both training and test (five studies, none rerun); on other FDIA data, attacks degrade classical
detection~\cite{sayghe2020survey}\fi
& How much of the headline the split contributes (Section~\ref{sec:split})\\
QML on this
dataset~\cite{cirillo2025quantum,ngo2026qupid,pornobi2026hybrid,adhikari2026quantumdiffusion}
& \ifarxivbuild The earliest we found: a quantum autoencoder on four features, level with a
classical baseline at about 72\% accuracy. QUPID: a variational network with a differential-privacy
noise defense, reporting quantum resilience under gradient-based evasion, with baselines at
literature hyperparameters and a 70/30 partition that describes no grouping, duplicate, or
group-aware check. Two 2026 accuracy studies, a hybrid network and a QSVM over quantum-diffusion
latents, report no tuned classical zoo, leakage check, or adversarial evaluation in their
abstracts\else The earliest we found, a quantum autoencoder, ties a classical baseline; QUPID, a
noise-defended variational network, faces gradient-based evasion against baselines at literature
settings; two 2026 accuracy studies (one a QSVM) report no tuned set of baselines, leakage check, or adversarial
evaluation\fi
& Beyond QUPID's own test: transferability, kernels, black-box attacks and label flips, seeded
statistics, and whether a ranking survives a whole-file split
(Sections~\ref{sec:split}--\ref{sec:blackbox})\\
Tabular QML
benchmarks~\cite{bowles2024better,kakavand2026benchmarking}\ifarxivbuild{} and quantum one-class
kernels~\cite{cultice2025quantum}\fi
& No generic quantum-kernel accuracy advantage; outcomes are dominated by dataset
choice\ifarxivbuild. On three water-infrastructure datasets, quantum-kernel one-class support
vector machines score 13.3\% higher F1 on average than classical one-class kernels, with no tree
or neural
baselines\fi
& The question on grid-security data, against tuned baselines with trees and networks
(Section~\ref{sec:rq1})\ifarxivbuild; the leading tabular foundation model~\cite{qu2026tabiclv2}
run through both split protocols\fi\\
Kernel theory~\cite{huang2021power,thanasilp2024exponential,shaydulin2022importance,canatar2023bandwidth,slattery2023numerical,florezablan2025similarity}
& \ifarxivbuild The geometric difference is necessary but not sufficient for
advantage~\cite{kubler2021inductive}; exponential concentration is a separate failure mode;
bandwidth tuning is required, but it drives the geometric difference down and pushes the kernel
toward classical behavior. Accuracy and the geometric difference have been seen to move in
opposite directions, without a mechanism~\cite{egginger2024hyperparameter}\else Geometric difference necessary, not sufficient; concentration a
separate failure mode; tuned bandwidths approach classical kernels\fi
& \ifarxivbuild Which screening quantity predicts accuracy on real labels, and a spectral-degeneracy
confound in the geometric difference\else Whether the kernels align with the real labels\fi{} (Section~\ref{sec:diag})\\
Adversarial
QML~\cite{weber2021optimal,du2021quantum,west2023towards,west2023benchmarking,wendlinger2024comparative,marchiori2025attaq,montalbano2025quantum,yu2024quantum,li2026qstaformer,nowmi2026sok}
& \ifarxivbuild Early vulnerability results~\cite{lu2020quantum,liu2020vulnerability}. Robustness
certificates and a bound that improves with depolarizing noise. Studies have reported empirical
robustness advantages on image data. Transfer across the quantum--classical boundary in both
directions on images, with directional asymmetries and no robustness-by-design effect. ATTAQ: white-box attacks and transfer in both directions on nongrid data,
quantum models more vulnerable; its abstract reports no decision-based black-box or poisoning
attack. A white-box attack on a fidelity kernel over 600 biomedical images, attributed to kernel
concentration. Grid stability models attacked. A survey: image- and malware-centric, variational
models only, no decision-based attacks, and an accuracy--robustness tradeoff on the encoding and
depth axis\else Robustness advantages reported on images, and two-way transfer there; ATTAQ,
white-box and transfer on nongrid data, no decision-based black-box or
poisoning attack in its abstract; attacked grid stability models; a survey\fi
& Decision-based black-box attacks and label flips, kernels and trees, on grid attack data. On
the MSU/ORNL data the transfer ratio follows the surrogate (Section~\ref{sec:surrogate})\ifarxivbuild;
concentration is measured absent (Section~\ref{sec:diag}); a kernel-bandwidth sweep is
consistent with the tradeoff (Section~\ref{sec:tuning}); where the obfuscated-gradients
story~\cite{athalye2018obfuscated} has black-box attacks expose false robustness, here the
black-box attack is itself fooled (Section~\ref{sec:blackbox})\fi\\
Evaluation practice~\cite{carlini2019evaluating,kapoor2023leakage,yin2026crossmodal}\ifarxivbuild{} and~\cite{arp2022dos,hoque2026interpretable}\fi
& \ifarxivbuild The standard checklist for adversarial-robustness evaluations\else The standard adversarial-evaluation checklist\fi; leakage \ifarxivbuild across\else in\fi{} 294 papers \ifarxivbuild in\else from\fi{}
17 fields; a 0.33 macro-F1 inflation on this dataset between random and event-level splits of a
random forest\ifarxivbuild. Data-snooping and sampling pitfalls across security ML; a 0.03
time-blocked gap on another cyber-physical dataset, with a physics-feasible query attack that
takes a random forest on the IEEE 14-bus system from 0.97 to 0.33\fi
& The size of each checklist choice on grid attack data, with its control
(Fig.~\ref{fig:checklist}); duplicate removal, tuned models, and the
leak's length (Section~\ref{sec:split})\\
\midrule
\textbf{This work}
& \multicolumn{2}{>{\hsize=\dimexpr2\hsize+2\tabcolsep\relax\raggedright\arraybackslash}X@{}}{Eight
model families under one protocol: white-box, transfer, decision- and score-based black-box
attacks, label flips, and sensor noise; accuracy under row-level and whole-file
splits\ifarxivbuild, robustness under the row-level split with whole-file replications\fi; up to ten seeds; a gas
pipeline and two simulated grids (\ifarxivbuild Figs.~\ref{fig:overview}
and~\ref{fig:threat}; \fi Tables~\ref{tab:attacks} and~\ref{tab:repdata})}\\
\bottomrule
\end{tabularx}
\end{table*}

\section{Methodology}
\ifarxivbuild Figs.~\ref{fig:threat}, \ref{fig:data}, and~\ref{fig:models} give each stage of the protocol, and
Table~\ref{tab:attacks} lists the models each attack reached; the
text below adds only what these exhibits cannot carry. \else Table~\ref{tab:attacks} lists the attacks and the models each one reached. \fi

\ifarxivbuild\begin{figure*}[!t]
\centering
\begin{tikzpicture}[node distance=1.6mm]
\pgfmathsetlengthmacro{\C}{(\textwidth-12mm-1pt)/3}
\pgfmathsetlengthmacro{\T}{\C-7.4pt}
\coordinate (c1) at (0,0);
\coordinate (c2) at ($(c1)+(\C+6mm,0)$);
\coordinate (c3) at ($(c2)+(\C+6mm,0)$);
\foreach \c/\h in {c1/Threat model, c2/Attacks, c3/Targets and metrics} {
  \draw[line width=0.8pt] (\c) -- ++(\C,0);
  \node[dhead] at ($(\c)+(0,-4.4mm)$) {\h};
}
\node[dbox, mattack, text width=\T, below=6.5mm of c1, anchor=north west] (goal)
  {\dt{Goal: misclassification}{a missed attack or a false alarm}};
\node[dbox, mattack, text width=\T, below=of goal.south west, anchor=north west] (cap)
  {\dt{Capability: an $\ell_\infty$ ball in standardized PCA space}{$\varepsilon \in \{0.01,
  0.05, 0.1, 0.2, 0.3, 0.5\}\,\sigma$ ($\sigma$: training-split standard deviation per
  component); end states to $2\sigma$ and beyond}};
\node[dbox, mattack, text width=\T, below=of cap.south west, anchor=north west] (know)
  {\dt{Knowledge depends on the model}{white-box: MLP, VQC, both kernel machines, logistic regression; gray-box
  (MLP surrogate, same data, scaler, and PCA): trees, and every model under the shared protocol
  but the MLP and the VQC; black-box (queries): every family}};
\node[dbox, mattack, text width=\T, below=of know.south west, anchor=north west] (train)
  {\dt{Training time: label poisoning}{random flips of 5, 10, or 20\% of the training labels,
  with no attacker knowledge}};
\node[dbox, mreference, text width=\T, below=of train.south west, anchor=north west] (phys)
  {\dt{No power-flow consistency enforced}{perturbations mapped back to the 128 raw
  channels\ifarxivbuild; one attack recomputes a relay's derived channels\fi; the simulated grids
  add a physics-respecting attacker}};
\node[dbox, mattack, text width=\T, below=6.5mm of c2, anchor=north west] (pgd)
  {\dt{White-box PGD}{20 steps of $0.25\varepsilon$ from one random start; MLP and VQC;
  500 test points, ten seeds}};
\node[dbox, mattack, cbar, text width=\T, below=of pgd.south west, anchor=north west] (kern)
  {\dt{White-box kernel attack}{the QSVM decision function differentiated through the
  simulator; the tuned RBF-SVM as matched control}};
\node[dbox, mattack, cbar, text width=\T, below=of kern.south west, anchor=north west] (tran)
  {\dt{Gray-box transfer}{MLP surrogate as tuned and with balanced weights (and an RBF-SVM
  surrogate); every set against every model}};
\node[dbox, mattack, cbar, text width=\T, below=of tran.south west, anchor=north west] (hsj)
  {\dt{Decision-based black-box}{HopSkipJump, about 600 queries (to about 10\,900 on three models), five
  or ten restarts, 100 points;
  random-perturbation control at matched radius}};
\node[dbox, mattack, text width=\T, below=of hsj.south west, anchor=north west] (sq)
  {\dt{Score-based black-box}{Square search, 600 queries per point at 0.1, 0.3, $0.5\sigma$;
  three seeds}};
\node[dbox, mattack, text width=\T, below=of sq.south west, anchor=north west] (pois)
  {\dt{Poisoning and benign noise}{label flips, then retraining; Gaussian sensor noise at
  three signal-to-noise ratios}};
\node[dbox, mreference, text width=\T, below=6.5mm of c3, anchor=north west] (fam)
  {\dt{All eight model families}{six classical \mdots{qgClassical}{6} (logistic regression,
  RBF-SVM, random forest, XGBoost, LightGBM, MLP) and two quantum \mdots{qgQuantum}{2}
  (QSVM, VQC)}};
\node[dbox, mreference, text width=\T, below=of fam.south west, anchor=north west] (met)
  {\dt{Metrics}{clean and attacked macro-F1; retention; macro-F1 minus the stratified-random
  floor}};
\node[dbox, mreference, text width=\T, below=of met.south west, anchor=north west] (bbr)
  {\dt{Black-box readings}{median $\ell_\infty$ distortion; random-flip rate; rays that cross
  more than one boundary}};
\node[dbox, mreference, text width=\T, below=of bbr.south west, anchor=north west] (op)
  {\dt{Operator view}{detection and false-alarm rates at a validation-fitted 1\%
  false-positive rate}};
\node[dbox, mreference, cbar, text width=\T, below=of op.south west, anchor=north west] (end)
  {\dt{End-state analysis}{disabled or inverted; loss- against distance-seeking objective;
  bandwidth and $C$ sweeps}};
\node[dbox, mcontrol, text width=\T, below=of end.south west, anchor=north west] (rep)
  {\dt{Replication under whole-file training}{white-box kernel contrast; random-flip control}};
\foreach \a/\b in {fam/met, met/bbr, bbr/op, op/end, end/rep} \draw[darrow] (\a.south) -- (\b.north);
\coordinate (lg) at ($(pois.south west -| c1)+(0,-3mm)$);
\draw[line width=0.8pt] (lg) -- ++(3*\C+12mm-1.2pt,0);
\node[anchor=north west, inner sep=0pt, font=\footnotesize] at ($(lg)+(0,-1.6mm)$)
  {\mkey{mattack}{attack}\quad \mkey{mcontrol}{control}\quad \mkey{mreference}{data,
  reference}\quad \tikz[baseline=-0.5ex]{\node[dbox, mreference, cbar, minimum width=4.2mm,
  minimum height=2.4mm, inner sep=0pt, anchor=center]{};}~holds a control that changed a
  conclusion (Fig.~\ref{fig:checklist})};
\end{tikzpicture}
\caption{Threat model, attacks, and what each attack is measured on. Attacks are crafted in
the models' standardized input space for every family.}
\label{fig:threat}
\end{figure*}\fi
\begin{table*}[!t]
\centering
\caption{Attacks and the models each one reached (binary task, eight dimensions). Orange: the
attack ran; green: the control; pink rows: quantum models}
\label{tab:attacks}
\footnotesize
\setlength{\tabcolsep}{4pt}
\newcommand{\hit}[1]{\cellcolor{qgAttackTint}#1}
\newcommand{\ctl}[1]{\cellcolor{qgControlTint}#1}
\newcommand{\none}{--}
\newcommand{\sub}[1]{\leavevmode{\color{dgrey}#1}}
\begin{tabularx}{\textwidth}{@{}l>{\raggedright\arraybackslash}Xllllll@{}}
\toprule
 & Own white-box attack & Transferred PGD & HopSkipJump & Random noise & Square search &
Evolution strategies\\
Model & \sub{500 points} & \sub{from the unweighted MLP} & \sub{100 points, restarts} &
\sub{the control} & \sub{500 points} & \sub{600 queries}\\
\midrule
Logistic regression & \hit{loss gradient, ten starts} & \hit{yes} & \hit{five} & \ctl{yes} & \hit{yes} & \none\\
RBF-SVM & \hit{kernel gradient, ten starts} & \hit{yes} & \hit{five} & \ctl{yes} & \hit{yes} & \none\\
Random forest & \none{} (no gradient-based one) & \hit{yes} & \hit{five} & \ctl{yes} & \hit{yes} & \hit{yes}\\
XGBoost & \none{} (no gradient-based one) & \hit{yes} & \hit{ten} & \ctl{yes} & \hit{yes} & \hit{yes}\\
LightGBM & \none{} (no gradient-based one) & \hit{yes} & \hit{ten} & \ctl{yes} & \hit{yes} & \hit{yes}\\
Unweighted MLP & \hit{PGD, ten seeds} & the source & \hit{ten} & \ctl{yes} & \hit{yes} & \ctl{positive control}\\
Class-weighted MLP & \hit{PGD, ten seeds, $0.5\sigma$ only} & \hit{yes} & \none & \none & \none & \none\\
\rowcolor{qgQuantumTint}QSVM & \hit{kernel gradient, ten starts} & \hit{yes} & \hit{ten} & \ctl{yes} & \hit{yes} & \none\\
\rowcolor{qgQuantumTint}VQC & \hit{PGD, ten seeds} & \hit{yes} & \hit{five} & \ctl{yes} & \hit{yes} & \none\\
\bottomrule
\end{tabularx}
\end{table*}

\subsection{Threat Model}\label{sec:threat}
\ifarxivbuild
The attacker's goal is \emph{misclassification}: to perturb measurements so that an attack sample is
classified as natural (evasion) or a natural sample as an attack (a false alarm). Our attacks perturb every row they are run on, natural rows included, so
an attacked macro-F1 counts both harms (Section~\ref{sec:rq2}). The attacker's knowledge depends on the model: white-box with full
gradients for the multilayer perceptron (MLP), the VQC, both kernel machines, and logistic
regression. It is gray-box for the three tree ensembles, and under the shared protocol for the
kernel machines and logistic regression too: an MLP surrogate shares their training data,
scaler, and principal component analysis (PCA). It is black-box, through about 600 queries per
point (class scores for the Square search and the evolution-strategies attack, labels alone for
HopSkipJump, per point it could start on), for every model family; a budget sweep gives
HopSkipJump up to about 10\,900 queries per point on three models
(Section~\ref{sec:blackbox}). The capability is an $\ell_\infty$-bounded perturbation
\emph{in the standardized PCA coordinate system}, which is the models' input space, in units of
the per-component training-split standard deviation $\sigma$ (Fig.~\ref{fig:threat}); the
full-feature detectors of Sections~\ref{sec:console} and~\ref{sec:discussion} are attacked in their own 128 standardized
features. This is
deliberately an abstract capability: a ball on 8--16 principal components corresponds to
correlated perturbations of all 128 raw channels, and some channels, such as relay status flags
and log counters, are discrete and cannot be perturbed by fractions. Query access and per-sample
perturbation are evaluation instruments that upper-bound a deployed attacker. A ball of radius
$\varepsilon$ contains every smaller ball, so an attacker bounded by $\varepsilon$ may use any
smaller perturbation. We therefore read every white-box attack at the tested radius, at or below its
budget, that leaves the lowest mean score over the ten attack starts, or the ten refits where a
model is refit per seed~\cite{carlini2019evaluating}. This reading is still an upper bound on
what the attacker can do. An attacker may pick the radius per input\appxsec{app:split}. A score that
rises with the radius, as the kernel machines' do, also means the 20-step attack has not
converged at the larger radii. That checklist reads such a curve as a fault of the attack, and
the worst-radius reading is the least it asks. The radius is chosen on the starts it is read
on: for the QSVM the two lowest radii differ by less than one standard deviation over the starts.
Where a control ran one radius only, its reading is an upper bound on that worst case, and we
name the radius. Transferred and black-box results keep the radius of the shared protocol
(Section~\ref{sec:attackprotocol}), except in the lowest-score column of the robustness table
(Section~\ref{sec:rq2}), which reads the two score-based attacks at their worst tested radius; on the
MSU/ORNL data the two readings of a transferred attack differ by at most 0.013 in retention. A
score-based attacker could choose its radius too. The natural-evolution-strategies attack of
Section~\ref{sec:rq2} does its greatest damage to the three tree ensembles at $0.3\sigma$, 0.08
to 0.11 macro-F1 more than at the bound, so its readings at the bound understate it by that
much. At
training time, an
attacker without knowledge of the models flips 5, 10, or 20\% of the training labels at random. On
the simulated grids, an attacker who knows the measurement model perturbs the estimated state
instead (Section~\ref{sec:physics}).
\else
The attacker's goal is \emph{misclassification}: an attack row classified as natural (evasion)
or a natural row as an attack. Our attacks perturb every test row, so an attacked macro-F1
counts both (Section~\ref{sec:rq2}).
Knowledge is white-box, gray-box through a multilayer perceptron (MLP) surrogate, or black-box
through about 600 queries per point an attack could start on (scores or, for HopSkipJump,
labels), by model as Table~\ref{tab:attacks} lists. The capability is an $\ell_\infty$-bounded perturbation of the
standardized principal component analysis (PCA) coordinates. Its unit is $\sigma$, each
component's training-split standard deviation (for the full-feature detectors of
Section~\ref{sec:console}, that of each of the 128 standardized features). The radius is
$\varepsilon \in \{0.01, 0.05, 0.1, 0.2, 0.3, 0.5\}\sigma$, or up to $2\sigma$ in the attack-objective runs of
Section~\ref{sec:rq2}. It is deliberately abstract: it moves most raw channels together, which upper-bounds a
deployed attacker (Section~\ref{sec:realizability}). We report each white-box attack at the tested radius, inside its budget,
that leaves the lowest mean score (Section~\ref{sec:rq2}). An attacker who picks the radius per
input cannot raise that score. At training time, an attacker without
knowledge of the models flips 5, 10, or 20\% of the training labels at random.
\fi

\subsection{Dataset and Splits}\label{sec:data}
\ifarxivbuild
\begin{figure}[!t]
\centering
\begin{tikzpicture}[node distance=2.2mm]
\pgfmathsetlengthmacro{\W}{\columnwidth-7.4pt}
\pgfmathsetlengthmacro{\H}{(\columnwidth-4pt)/2-7.4pt}
\node[dbox, mreference, text width=\W] (src) at (0,0)
  {\dt{15 released source files (MSU/ORNL testbed)}{two generators, two lines, and four relays,
  each with a PMU; 128 features; 36 or 37 of the 37 scenarios per file; no timestamp column}};
\node[dbox, mreference, text width=\W, below=of src.south west, anchor=north west] (smp)
  {\dt{Each file: an order-preserving 1\% sample of one recording}{inferred from row order;
  adjacent rows are near-copies}};
\node[dbox, mreference, text width=\W, below=of smp.south west, anchor=north west] (prep)
  {\dt{Pool files, remove duplicates, impute infinite impedances}{imputation, scaler, and PCA
  are fit on the training split only}};
\node[dbox, mreference, text width=\W, below=of prep.south west, anchor=north west] (rows)
  {\dt{\NumRowsBinary{} rows: attack against natural}{\NumAttackShare\%/\NumNaturalShare\%;
  a three-class variant (no event, natural, attack) replicates the diagnostics\\
  \pbar{0.97*\W}{\NumAttackShare/qgAttack, \NumNaturalShare/qgReference!55}}};
\foreach \a/\b in {src/smp, smp/prep, prep/rows} \draw[darrow] (\a.south) -- (\b.north);
\coordinate (rule) at ($(rows.south west)+(0,-6.2mm)$);
\draw[line width=0.8pt] (rule) -- ++(\columnwidth-1.2pt,0);
\node[dhead] (sh) at ($(rule)+(0,1.1mm)$) {Split protocols};
\node[dnote, right=1.5mm of sh.south east, anchor=south west] {same transforms; only the
  split indices change};
\node[dbox, mchoice, text width=\H, minimum height=14.2mm, below=2.2mm of rule, anchor=north west]
  (rl) {\dt{Row-level 60/20/20}{stratified; fixed indices shared by every model
  (\NumTrainRows{} training rows)}};
\node[dbox, mchoice, text width=\H, minimum height=14.2mm, right=4pt of rl.north east,
  anchor=north west] (wf) {\dt{Within-file}{random rows inside each file: the rule the
  benchmark was introduced with}};
\node[dbox, mcontrol, text width=\H, minimum height=14.2mm, below=4pt of rl.south west,
  anchor=north west] (ho) {\dt{Whole files held out}{5 and 20 file partitions; the
  row-level tuning is reused, and retuning is checked separately}};
\node[dbox, mcontrol, text width=\H, minimum height=14.2mm, right=4pt of ho.north east,
  anchor=north west] (tm) {\dt{Temporal and middle-band}{positional splits inside each
  scenario block; the middle 60\% trains}};
\node[dbox, mcontrol, text width=\W, below=5.5mm of ho.south west, anchor=north west] (aud)
  {\dt{Leakage audit}{share of test rows with a training neighbor closer than 0.1
  (standardized); score against file-order distance gives the leak's length}};
\draw[darrow] (aud.north |- ho.south) -- (aud.north);
\node[dbox, mreference, text width=\W, below=of aud.south west, anchor=north west] (out)
  {\dt{Accuracy reported per protocol}{robustness runs use the row-level split; the white-box
  kernel contrast and the random control are repeated under whole-file training}};
\draw[darrow] (aud.south) -- (out.north);
\end{tikzpicture}
\caption{Data preparation and the split protocols. Every protocol uses the same transforms;
only the split indices change.}
\label{fig:data}
\end{figure}
\fi
We use the public dataset~\cite{hink2014machine}\ifarxivbuild{} (Fig.~\ref{fig:data})\fi. Its features come from
PMU phasors, relays, control panels, and Snort logs on a hardware-in-the-loop testbed, with
scenario-assigned labels. We study how detectors of attacks on its measurement and log
data are evaluated, not the protection scheme itself. \ifarxivbuild Holding whole files out is the coarsest
block holdout the release exposes. \fi The scenarios include data injection, command injection,
and relay-setting changes, so we describe the task as cyberattack detection with FDIA focus.
\ifarxivbuild\else Each of the 15 released files is an order-preserving 1\% sample of what we read as a single recording (an inference from row order). We pool the files and remove duplicates, which leaves \NumRowsBinary{} rows of 128 features, \NumAttackShare\% of them attacks. We impute infinite impedance values from the training split, then partition the rows 60/20/20 with stratification and fixed indices shared by every model. A whole-file split control, each partition holding out three of the 15 files, measures the leakage of this row-level split (Section~\ref{sec:split}); all fitted transforms are fit on the training split only. \fi

\subsection{Replication Data}\label{sec:repdata}
\ifarxivbuild\else\begin{table}[!t]
\centering
\caption{Replication data: rows kept of those recorded or generated; best matched-contest
macro-F1 (eight dimensions, row-level split, gas pressure log-scaled); held-out unit of the
split check}
\label{tab:repdata}
\footnotesize
\setlength{\tabcolsep}{4pt}
\begin{tabular}{@{}lccc@{}}
\toprule
 & Gas pipeline & 14-bus grid & 118-bus grid\\
\midrule
Rows kept & \NumGasRows{} of \NumGasRawRows & \NumGridRows{} of \NumGridRowsGenerated & \NumGridRows{} of \NumGridRowsGenerated\\
Features & \NumGasFeatures & \NumGridSmallFeatures & \NumGridLargeFeatures\\
Attack rows & \NumGasAttackShare\% & \NumGridAttackShare\% & \NumGridAttackShare\%\\
Best classical & \NumGasBestClassical & \NumGridSmallBestClassical & \NumGridLargeBestClassical\\
Best quantum & \NumGasBestQuantum & \NumGridSmallBestQuantum & \NumGridLargeBestQuantum\\
Held-out unit & time blocks & days & days\\
\bottomrule
\end{tabular}
\end{table}
\fi
\ifarxivbuild
We repeat the pipeline on two further sources (Table~\ref{tab:repdata}). Each replication check has a threshold that was
written down before its result was read; we call these thresholds registered
(Section~\ref{sec:replication} and its figure's caption note the cells read by a stricter or a
later rule). The
first is the gas-pipeline dataset from the same maintainer~\cite{morris2014icsdata}: 16 features
of a laboratory gas pipeline's control-system traffic with a binary attack label, deduplicated
before a stratified 60/20/20 split, with missing fields filled by the training median. Under
the main study's pipeline, unchanged (the \emph{main-study pipeline}), standardization maps every
ordinary pressure reading to one value, because injected pressure values set the scale. After a
first registered reading under it, we passed the pressure feature through a fixed signed
logarithm, retuned every model at the same budget, and reran every gas check of the figure in
Section~\ref{sec:replication} with its threshold
unchanged. The gas readings of that figure use that pipeline, and the section says
where the main-study pipeline's readings differ. Two further gas checks, marked in the
table of Appendix~\ref{app:further}, ran under the main-study pipeline only. The second is false data injection that we generate on the IEEE
14-bus and 118-bus test systems~\cite{christie1999pstca}, as shipped with
pandapower~\cite{thurner2018pandapower}, with an alternating-current power flow.
\else
We repeat the pipeline on the gas-pipeline dataset from the same
maintainer~\cite{morris2014icsdata}, control-system traffic of a laboratory pipeline with a
binary attack label, and on false data injection that we generate on
the IEEE 14-bus and 118-bus systems~\cite{christie1999pstca} with an alternating-current power
flow in pandapower~\cite{thurner2018pandapower} (Table~\ref{tab:repdata}). Each replication check has a threshold that
was written down before its result was read; we call these thresholds registered
(Section~\ref{sec:replication} and its figure's caption note the exceptions). On the gas data, under the main study's pipeline, unchanged (the
\emph{main-study pipeline}), injected pressure values up to $3.4 \times 10^{38}$ set the scale of
the standardization, so every ordinary pressure reading maps to one value. We then passed
pressure through a fixed signed logarithm, retuned every model, and reran the gas checks; the
gas readings of Section~\ref{sec:replication} use that pipeline.
\fi
Each grid has 90 simulated days of 15-minute snapshots\ifarxivbuild. Loads follow the weekly, daily, and
hourly peak-load percentages of the IEEE Reliability Test System (archived in~\cite{christie1999pstca}),
interpolated to 15 minutes, with a per-bus autoregressive deviation of standard deviation 3\%;
generation scales with total load\fi. A day holds 12 two-hour episodes in random order: 4
normal, 4 stealthy, and 4 whose injection is not stealthy, which we drop. A row is one
snapshot's measurement vector $z$; a stealthy row is $z_a$,
generalizing~\cite{liu2011false}:
\begin{equation}\label{eq:fdia}
z = h(s) + e, \qquad z_a = z + h(s + c) - h(s).
\end{equation}
Here $s$ is the state (bus voltage magnitudes and nonreference angles), and $h$ gives the
features: voltage magnitudes, power injections, and from-end line flows. The
noise $e$ is Gaussian, with a standard deviation of 1\% of
the channel's base-case magnitude or of its measurement type's median nonzero magnitude,
whichever is larger. In
a stealthy episode, $c$ shifts up to half of the angles, by 0.005--0.02~rad or 0.02--0.05~rad of
either sign or by a normal draw of standard deviation 0.14~rad; the three sizes take turns.
Stealthy rows pass the chi-square bad-data test of the
weighted-least-squares state estimator (1\% false-alarm rate\ifarxivbuild{} by design; it flags 1.2\% and 1.3\% of all generated normal rows\fi) as often as normal rows. A
physics-consistent attack on a detector moves a row by $h(\hat{s} + d) - h(\hat{s})$, where
$\hat{s}$ is the row's state estimate. Each entry of $d$ is bounded by $\varepsilon$ times that
state's training standard deviation, which is this attack's $\sigma$. The attack it is compared
with ignores the physics and moves each measurement by at most $\varepsilon$ training standard
deviations (Section~\ref{sec:physics}).
\ifarxivbuild
Normal and stealthy rows are balanced (5760 per grid, split 60/20/20 by row, with
days as the held-out group). The same eight model families, tuning budget, PCA-8 regime, and attacks run
on every source; the simulated grids add the Square search~\cite{andriushchenko2020square} at
0.1, 0.3, and $0.5\sigma$ (the $\sigma$ of Section~\ref{sec:threat}).
\begin{table}[!t]
\centering
\caption{Replication data: rows kept of those recorded or generated; best matched-contest
macro-F1 (eight dimensions, row-level split, gas pressure log-scaled); held-out unit of the
split check}
\label{tab:repdata}
\footnotesize
\setlength{\tabcolsep}{4pt}
\begin{tabular}{@{}lccc@{}}
\toprule
 & Gas pipeline & 14-bus grid & 118-bus grid\\
\midrule
Rows kept & \NumGasRows{} of \NumGasRawRows & \NumGridRows{} of \NumGridRowsGenerated & \NumGridRows{} of \NumGridRowsGenerated\\
Features & \NumGasFeatures & \NumGridSmallFeatures & \NumGridLargeFeatures\\
Attack rows & \NumGasAttackShare\% & \NumGridAttackShare\% & \NumGridAttackShare\%\\
Best classical & \NumGasBestClassical & \NumGridSmallBestClassical & \NumGridLargeBestClassical\\
Best quantum & \NumGasBestQuantum & \NumGridSmallBestQuantum & \NumGridLargeBestQuantum\\
Held-out unit & time blocks & days & days\\
\bottomrule
\end{tabular}
\end{table}

\fi

\subsection{Models, Tuning, and Diagnostics}\label{sec:models}
\ifarxivbuild
\begin{figure}[!t]
\centering
\begin{tikzpicture}[node distance=2.2mm]
\pgfmathsetlengthmacro{\W}{\columnwidth-7.4pt}
\pgfmathsetlengthmacro{\H}{(\columnwidth-4pt)/2-7.4pt}
\node[dbox, mreference, text width=\W] (tr) at (0,0)
  {\dt{Training split (\NumTrainRows{} rows)}{scaler fit on the training split only}};
\node[dbox, mclassical, text width=\H, minimum height=17.6mm, below=5mm of tr.south west,
  anchor=north west] (ra) {\dt{Regime (a): all 128 features}{no reduction; the full training
  split (RBF-SVM: a stratified 10\,000-row cap)}};
\node[dbox, mreference, text width=\H, minimum height=17.6mm, right=4pt of ra.north east, anchor=north west]
  (rb) {\dt{Regime (b): PCA to 8, 12, or 16}{rescaled to $[-\pi,\pi]$; qubits $=$ dimensions;
  a stratified 2000-row cap, forced by the $O(N^2)$ kernel cost, for every model}};
\node[dbox, mclassical, text width=\H, minimum height=29.6mm, below=of ra.south west, anchor=north west]
  (cz) {\dt{Classical zoo \mdots{qgClassical}{6}}{logistic regression, RBF-SVM, random forest,
  XGBoost, LightGBM, two-layer MLP; class weights (the MLP takes balanced sample weights; its
  unweighted fit is kept as the transfer surrogate); runs in both regimes}};
\node[dbox, mquantum, text width=\H, minimum height=29.6mm, below=of rb.south west, anchor=north west]
  (qm) {\dt{Quantum models \mdots{qgQuantum}{2}}{QSVM: fidelity kernel with RY-angle or ZZ map,
  encoding bandwidth tuned; regime (b) only. VQC: angle encoding and strongly entangling layers, depth 2, 4,
  or 6; class-weighted loss. Exact statevector simulation}};
\draw[darrow] (ra.north |- tr.south) -- (ra.north);
\draw[darrow] (rb.north |- tr.south) -- (rb.north);
\draw[darrow] (ra.south) -- (cz.north);
\draw[darrow] (rb.south) -- (qm.north);
\node[dbox, mreference, text width=\W, below=5mm of cz.south west, anchor=north west] (tune)
  {\dt{Equal-budget tuning for every model}{Optuna, 50 trials on validation macro-F1, over
  seeds 0, 1, and 2 (classical) or seed 0 (VQC; the QSVM is deterministic); the quantum models
  are scored on a 2000-row validation subset; the winner is retrained on seeds 0 to 9}};
\draw[darrow] (cz.south) -- (cz.south |- tune.north);
\draw[darrow] (qm.south) -- (qm.south |- tune.north);
\node[dbox, mreference, text width=\W, below=of tune.south west, anchor=north west] (test)
  {\dt{Test split touched once per final configuration}{mean $\pm$ std over ten seeds;
  Wilcoxon signed-rank tests with Holm correction; paired bootstrap (10\,000 resamples) for
  the deterministic QSVM}};
\node[dbox, mquantum, text width=\W, below=of test.south west, anchor=north west] (diag)
  {\dt{Kernel diagnostics against the tuned RBF kernel}{KTA and geometric difference $g$,
  Eq.~(\ref{eq:diag}); label-alignment positive control}};
\draw[darrow] (tune.south) -- (test.north);
\draw[darrow] (test.south) -- (diag.north);
\end{tikzpicture}
\caption{Feature regimes, models, and tuning. Every model gets the same tuning budget and the
same test protocol.}
\label{fig:models}
\end{figure}
\fi
Classical models run in two regimes\ifarxivbuild{} (Fig.~\ref{fig:models})\fi: (a) all 128 features, and (b)
PCA reductions to 8, 12, or 16 dimensions, where each quantum model uses one qubit per
dimension. Regime (b) is the \emph{matched contest}: every model at the same dimensionality and
the same training cap. The study has eight model families, six classical and two quantum; we say
\emph{family} for one of the eight and write quantum or classical models for the two groups. The
six classical families are logistic regression, a radial
basis function (RBF) support vector machine (RBF-SVM), random forest, XGBoost, LightGBM, and a
two-layer MLP. They are tuned with Optuna~\cite{akiba2019optuna}\ifarxivbuild; the VQC is trained with Adam
and early stopping, and the class-weighted MLP is the accuracy model at eight dimensions
(Section~\ref{sec:rq1}).\else{} under an \emph{equal budget} of 50 trials per model on validation macro-F1 over seeds \{0,1,2\}, and the best configuration is retrained on all ten seeds. Class weights
handle the imbalance, except in the MLP as first tuned, whose library takes none. We keep that
\emph{unweighted MLP} as the surrogate that transferred attacks are crafted on, because
Section~\ref{sec:surrogate} shows what its fit does to transfer results. The \emph{class-weighted
MLP}, trained with balanced sample weights, is the accuracy model at eight dimensions
(Table~\ref{tab:rq1}). In regime (b) every model is capped at 2000
stratified training rows, which the Gram matrix's $O(N^2)$ cost forces.

The quantum models get the same 50 trials, scored on 2000 validation rows, the VQC at one seed
(the QSVM is deterministic). The VQC uses angle encoding with strongly entangling layers (depth
2, 4, or 6), Adam with early stopping, and class-weighted loss.\fi{} The QSVM uses the fidelity kernel
$k(x,x')=|\langle\phi(x)|\phi(x')\rangle|^2$ with an RY-angle map (one rotation about the
$y$-axis per feature) or a ZZ map\ifarxivbuild{} (Fig.~\ref{fig:circuits})\else, in which
products of feature pairs set entangling rotations\fi. Each PCA component is min-max
scaled to $[-\pi, \pi]$. Under this scaling the ZZ-map QSVM scores higher at eight dimensions,
and it is the QSVM of every evasion experiment. The encoding bandwidth, a scale
factor on these angles, sets the kernel's length scale and is
tuned, as the literature requires~\cite{shaydulin2022importance,canatar2023bandwidth}. Circuits are
simulated with exact statevectors in PennyLane~\cite{bergholm2018pennylane}\ifarxivbuild; a
shot-noise study (analytic, 1024, and 4096 shots) and a depolarizing-noise ablation bridge toward
device conditions\fi.

Following~\cite{huang2021power}, two diagnostics compare each quantum kernel\ifarxivbuild{}
$K_Q$\fi{} with a comparison RBF kernel\ifarxivbuild{} $K_C$\fi{} whose width is tuned on validation beside it. The kernel--target alignment (KTA) measures how well a kernel's similarity structure lines up with the labels (1 is a perfect match). The geometric difference\ifarxivbuild{} $g$\fi{} measures how far the quantum kernel is from the classical one\ifarxivbuild. The two are defined by
\begin{equation}\label{eq:diag}
\mathrm{KTA} = \frac{\langle \bar K, \bar K_y\rangle_F}{\|\bar K\|_F\,\|\bar K_y\|_F},
\quad
g = \sqrt{\bigl\|K_Q^{1/2} K_C^{-1} K_Q^{1/2}\bigr\|_\infty},
\end{equation}
where $\bar K$ is the doubly centered Gram matrix, and $K_y = yy^\top$ for $y \in \{\pm1\}^n$
(one-hot Gram for three classes). Both kernels are trace-normalized to $n$, and $K_C$ is
ridge-regularized ($\lambda = 10^{-6}$). \else; \cite{islam2026benchmarkingarxiv} gives both formulas. \fi As a positive control we also build labels on which a quantum kernel should win: for each
quantum Gram matrix, labels that follow the direction that maximizes its geometric difference
from the classical kernel.\ifarxivbuild{} If the pipeline can detect a quantum win, it must appear there.\fi

\ifarxivbuild
\begin{figure}[!t]
\centering
\includegraphics[width=\columnwidth]{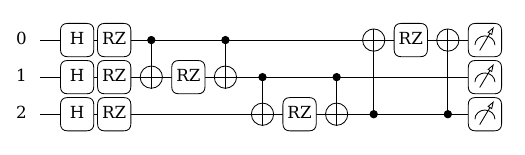}\\[2pt]
\includegraphics{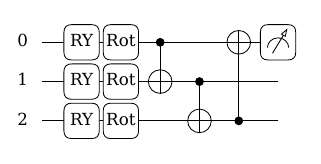}
\caption{Model circuits, shown for three qubits (the experiments use 8 to 16). Top: one of the
two repetitions of the QSVM's ZZ feature map: a Hadamard (H) gate and an $R_Z(2x_i)$ rotation per
qubit, then an $R_Z(2x_ix_j)$ rotation between two controlled-NOT (CNOT) gates for each
neighboring pair on a ring. Bottom: the VQC, an RY angle encoding, one strongly entangling layer
of single-qubit rotations (Rot) and CNOTs, and a Pauli-Z readout on qubit 0.}
\label{fig:circuits}
\end{figure}
\fi

\subsection{Attacks and Statistics}\label{sec:attackprotocol}
\ifarxivbuild Fig.~\ref{fig:threat} lists every attack, its budget, and its targets, and
Table~\ref{tab:attacks} the models each one reached. \fi Projected gradient descent
(PGD)~\cite{madry2018pgd}, 20 steps of $0.25\varepsilon$ from one random start, runs
white-box on the unweighted MLP and the VQC, with quantum gradients taken from the simulator\ifarxivbuild, and so does the fast gradient sign method
(FGSM)~\cite{goodfellow2015fgsm}\fi. Every crafted attack set is evaluated against every model\ifarxivbuild,
which yields a source$\times$target transfer matrix\fi. We call PGD at $0.5\sigma$, white-box on the unweighted MLP and the VQC and transferred from
that MLP to every other model, the \emph{shared protocol}. The QSVM's white-box attack is crafted on the
differentiated kernel and evaluated on the \ifarxivbuild deployed kernel path (the differentiated
kernel agrees with an exact reference to $2\times10^{-5}$ and with finite differences to
$4\times10^{-7}$; the deployed path sits $2\times10^{-3}$ from it); on the
gas data it is evaluated on the exact path (Section~\ref{sec:replication})\else
kernel as deployed (one fitted model, ten attack starts; gas data:
Section~\ref{sec:replication})\fi. For choice 4 the QSVM also receives
a distance-seeking attack at radii up to $2\sigma$ (200 test points). It does not raise the loss; it lowers each input's mean kernel value against the model's
support vectors. We run HopSkipJump\ifarxivbuild{} (HSJ)\fi~\cite{chen2020hopskipjump} at about 600 queries
per point it could start on, via the Adversarial
Robustness Toolbox (ART)~\cite{nicolae2018art}; because ART draws its initial adversarial point
from an unseeded generator, we report it over five or ten restarts (Section~\ref{sec:blackbox})\ifarxivbuild. Three
models also receive it at about 2800 and 10\,900 queries per point, in a sweep that
Section~\ref{sec:blackbox} reports. Where
no draw is adversarial for an input, ART returns the input unchanged. On the MSU/ORNL data that
happened on most inputs
for the RBF-SVM and the random forest, so for those two we read the distortion over the inputs the
attack could start on and report their share. For the other models it started on at least 98\% of
the inputs where the share was recorded, and we read the distortion over all of them in the
published runs and over the inputs the attack flipped in the query sweep. On the gas
data it started on 90\% to 96\% of the inputs for the two quantum models, on all of them for
logistic regression, and on 51\% to 63\% for the other classical models. The registered gas rule
reads the median over all inputs, and the QSVM keeps its rank over the started inputs alone. ART's
step-size search has no bound: it repeats one query forever when its candidate equals the current
point in single precision and that point is not adversarial. Our wrapper ends the search there
and counts the event. The search stalled one gas run, which we stopped and reran with the wrapper. Of the
reported runs, only one middle-budget restart of the query sweep (Section~\ref{sec:blackbox})
contains the event, once. The
score-based Square search~\cite{andriushchenko2020square} runs against every family, and natural evolution strategies~\cite{ilyas2018blackbox} at a matched 600-query budget attack the
trees, with the unweighted MLP as the attack's positive control. Three defenses are trained and
then attacked (Section~\ref{sec:rq3}). Two are for the MLP and the VQC: one round of adversarial augmentation with
PGD examples at $0.5\sigma$, and min-max training that replaces every minibatch by its 10-step PGD
attack at $0.5\sigma$. The third is a QUPID-type input-noise defense\else. The score-based Square
search~\cite{andriushchenko2020square} runs against every family, natural evolution
strategies~\cite{ilyas2018blackbox} attack the trees at a matched 600-query budget, and three
defenses are trained and then attacked (Section~\ref{sec:rq3})\fi. The QSVM is deterministic given its training subset, so we judge
it by its margin over ten draws of that subset, and a paired percentile bootstrap
over test points (Section~\ref{sec:rq1}) checks each margin. \ifarxivbuild We report macro-F1,
area under the precision-recall curve, balanced accuracy, area under the receiver operating
characteristic curve (ROC-AUC), Brier score, and expected calibration error, plus training time
and per-sample inference latency. The headline task is binary; a three-class variant replicates
the diagnostic findings\appxsec{app:threeclass}.\else We report macro-F1 and per-sample
inference latency. The headline task is binary; the complete
version~\cite{islam2026benchmarkingarxiv} adds five metrics and a three-class variant.\fi{} Headline
comparisons use paired Wilcoxon signed-rank tests across seeds with Holm correction, and the test
split is touched once per final configuration. With fixed splits, these tests measure
seed-to-seed variability, not resampling variance.

\section{Results}\label{sec:results}

\ifarxivbuild Sections~\ref{sec:split} and~\ref{sec:surrogate} to~\ref{sec:blackbox} take the seven choices of
Fig.~\ref{fig:checklist} in its order. RQ1 is answered
in Section~\ref{sec:rq1}, RQ2 in Sections~\ref{sec:surrogate} to~\ref{sec:blackbox}, and RQ3 in
Section~\ref{sec:rq3}. Appendix~\ref{app:further} lists further controls behind
each subsection.\fi

\subsection{Choice 1: The Split Protocol Decides the Headline Number}\label{sec:split}
\subsubsection{The Leak}
Under the row-level split, 45.8\% of test rows have a near-copy in training,
almost always a row a few positions away in their file (Fig.~\ref{fig:splitleak}(c)), so almost half of
the test rows are rows the model has in effect seen. Holding whole files out removes most of them.
\subsubsection{What the Leak Is Worth}
\begin{figure*}[!t]
\centering
\includegraphics[width=\textwidth]{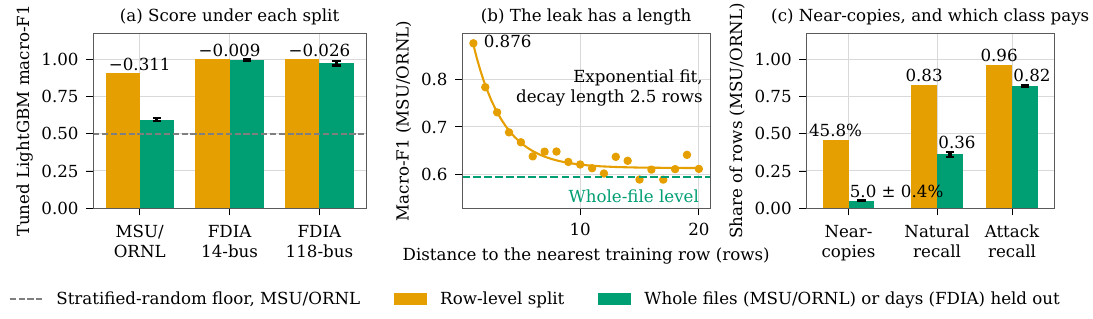}
\caption{Split protocol and its leak. (a) Tuned full-feature LightGBM under the row-level
split and with whole groups held out: MSU/ORNL source files (five partitions, $0.594 \pm 0.008$) and, for the
simulated grids, day folds (five folds). (b) LightGBM's MSU/ORNL
macro-F1 against the distance from a test row to its nearest training row, with the
exponential fit and the whole-file level (trained on the middle 60\% of each block of
consecutive rows of one scenario). (c) MSU/ORNL under both splits: test rows with a
training row closer than 0.1 (Euclidean, standardized full-feature space; under the row-level split, in
99\% of cases a row of the same file within ten positions, in 71\% the adjacent row), and recall of each class. Error bars: standard deviations.}
\label{fig:splitleak}
\end{figure*}

With whole files held out, the same tuned LightGBM loses 0.31 macro-F1
(Fig.~\ref{fig:splitleak}(a)), with only the split indices changed. The natural class carries
most of the loss: the detector labels most natural events as attacks
(Fig.~\ref{fig:splitleak}(c)). The dataset's
original rule, random rows inside each file, \ifarxivbuild scores 0.906 and leaks the same
way\else leaks the same way (macro-F1 0.906)\fi.\ifarxivbuild{} An independent study reports
a 0.33 inflation between random and event-level splits of a random forest~\cite{yin2026crossmodal}.\fi{}
Retuning on held-out files recovers $+0.009 \pm 0.008$, so the loss is the protocol's.
\subsubsection{The Leak Has a Length}
The inflation fades with distance to the nearest training row, with a decay length of 2.5
released rows (Fig.~\ref{fig:splitleak}(b))\ifarxivbuild, about 250 recorded samples. This one curve predicts the scores of five other splits
within 0.02\fi. The leak is therefore local: keeping test rows a few positions from every
training row would remove most of it. Blocked cross-validation rests on the same distance
effect~\ifarxivbuild\cite{roberts2017crossvalidation,ploton2020spatial};
we add its length on a cyberattack benchmark and its power to predict other splits\else\cite{roberts2017crossvalidation};
we add its length on a cyberattack benchmark. A random-forest study reports a similar inflation
on this dataset~\cite{yin2026crossmodal}\fi.
\ifarxivbuild
\subsubsection{The Simulated Grids as a Contrast}
The simulated grids' rows have no near-copies: no test row has a
training row closer than 0.1, and holding out whole days costs 0.009 and 0.026.
\fi
\subsubsection{The Matched Contest on Held-Out Files}
Under the 2000-row cap, both models clear
the stratified-random floor by little: tuned XGBoost by $+0.032$ (95\% bootstrap interval $[0.026, 0.039]$) and the
ZZ-map QSVM by $+0.023$ ($[0.013, 0.033]$), over 20 partitions (\ifarxivbuild seed 0 of 3; the
other two agree within 0.003\else first of 3 seeds\fi). Their difference changes sign on 7 of the 20.
\ifarxivbuild
\subsubsection{Leak Exploitation Differs by Family}
The slope of score against the share of test
rows with a near training neighbor differs $27\times$ across five tuned classical families, in the
order of their row-level scores (Spearman 0.90). This protocol therefore ranks capacity to exploit the
leak.
\subsubsection{Files Are Replicates}
A model trained on the other files of the held-out group scores
0.565, against 0.586 for one trained on as many rows from the training files. The effective replication count for a cross-file claim is therefore 15 files, not 78\,368 rows (Table~\ref{tab:split}). On
file-disjoint folds, where every file is tested exactly once, LightGBM reaches $0.601 \pm 0.013$
and the matched quantum--classical contrast is $+0.000 \pm 0.024$, with 3 of 5 folds positive.
\fi

\ifarxivbuild
\begin{table*}[!t]
\centering
\begin{threeparttable}
\caption{Split protocol as an evaluation choice: identical tuned models, identical
transforms, only the split indices change}
\label{tab:split}
% auto-generated by qgridbench.eval.tables -- do not edit by hand
\begin{tabular}{lrrr}
\toprule
 & Row-level split & Within-file split & Whole files held out \\
\midrule
Full 128 features, tuned LightGBM & 0.905 & 0.906\,$\pm$\,0.00012 & 0.594\,$\pm$\,0.008 \\
PCA-8 (2000-row cap), tuned XGBoost & 0.596 & 0.586\,$\pm$\,0.003 & 0.523\,$\pm$\,0.013 \\
PCA-8 (2000-row cap), tuned ZZ-map QSVM & 0.564 & 0.556\,$\pm$\,0.013 & 0.508\,$\pm$\,0.012 \\
\quad Contrast (classical $-$ quantum) & $+0.031$ & $+0.031\,\pm\,0.011$ & $+0.015\,\pm\,0.015$ \\
\midrule
Stratified-random floor & 0.501 & 0.500\,$\pm$\,0.00030 & 0.499\,$\pm$\,0.001 \\
Test rows with train neighbor $<0.1$ & 45.8\% & 46.0\,$\pm$\,0.1\% & 5.0\,$\pm$\,0.4\% \\
\bottomrule
\end{tabular}

\begin{tablenotes}[flushleft]
\footnotesize
\setlength{\labelsep}{0pt}
\item[] The within-file column is the rule the benchmark was introduced with (mean$\,\pm\,$std
over three partitions). The whole-file column is mean$\,\pm\,$std over five held-out partitions at
one training seed. The last row is the fraction of test rows with a train neighbor closer than
$0.1$ in standardized full-feature space.
\end{tablenotes}
\end{threeparttable}
\end{table*}
\fi

\subsection{RQ1: No Quantum Model We Ran Leads a Matched Row-Level Contest}\label{sec:rq1}

\ifarxivbuild
\begin{table*}[!t]
\centering
\begin{threeparttable}
\caption{RQ1 at matched dimensionality (PCA-8, binary task, test split, ten seeds)}
\label{tab:rq1}
% auto-generated by qgridbench.eval.tables -- do not edit by hand
\begin{tabular}{lrrrrrr}
\toprule
Model & Macro-F1 & Balanced accuracy & AUPRC & ROC-AUC & Brier & ECE \\
\midrule
Logistic regression$^{b}$ & 0.511\,$\pm$\,0.000 & 0.538\,$\pm$\,0.000 & 0.746\,$\pm$\,0.000 & 0.540\,$\pm$\,0.000 & 0.496\,$\pm$\,0.000 & 0.038\,$\pm$\,0.000 \\
RBF-SVM$^{b}$ & 0.578\,$\pm$\,0.000 & 0.576\,$\pm$\,0.000 & 0.762\,$\pm$\,0.000 & 0.605\,$\pm$\,0.000 & 0.466\,$\pm$\,0.000 & 0.084\,$\pm$\,0.000 \\
Random forest$^{b}$ & 0.586\,$\pm$\,0.002 & 0.582\,$\pm$\,0.001 & 0.793\,$\pm$\,0.001 & 0.635\,$\pm$\,0.001 & 0.408\,$\pm$\,0.000 & 0.030\,$\pm$\,0.001 \\
XGBoost$^{a}$ & 0.597\,$\pm$\,0.002 & 0.595\,$\pm$\,0.002 & 0.784\,$\pm$\,0.001 & 0.630\,$\pm$\,0.001 & 0.438\,$\pm$\,0.001 & 0.069\,$\pm$\,0.002 \\
LightGBM$^{b}$ & 0.593\,$\pm$\,0.001 & 0.590\,$\pm$\,0.001 & 0.784\,$\pm$\,0.001 & 0.632\,$\pm$\,0.000 & 0.454\,$\pm$\,0.000 & 0.130\,$\pm$\,0.001 \\
MLP$^{b}$ & 0.522\,$\pm$\,0.015 & 0.542\,$\pm$\,0.008 & 0.754\,$\pm$\,0.008 & 0.559\,$\pm$\,0.012 & 0.492\,$\pm$\,0.013 & 0.044\,$\pm$\,0.014 \\
\rowcolor{qgQuantumTint}QSVM (RY-angle) & 0.527\,$\pm$\,0.000 & 0.548\,$\pm$\,0.000 & 0.751\,$\pm$\,0.000 & 0.565\,$\pm$\,0.000 & 0.519\,$\pm$\,0.000 & 0.112\,$\pm$\,0.000 \\
\rowcolor{qgQuantumTint}QSVM (ZZ) & 0.564\,$\pm$\,0.000 & 0.569\,$\pm$\,0.000 & 0.752\,$\pm$\,0.000 & 0.584\,$\pm$\,0.000 & 0.515\,$\pm$\,0.000 & 0.117\,$\pm$\,0.000 \\
\rowcolor{qgQuantumTint}VQC$^{b}$ & 0.514\,$\pm$\,0.007 & 0.520\,$\pm$\,0.006 & 0.740\,$\pm$\,0.006 & 0.536\,$\pm$\,0.007 & 0.484\,$\pm$\,0.004 & 0.059\,$\pm$\,0.013 \\
\rowcolor{qgReferenceTint}Majority class & 0.415\,$\pm$\,0.000 & 0.500\,$\pm$\,0.000 & 0.710\,$\pm$\,0.000 & 0.500\,$\pm$\,0.000 & 0.579\,$\pm$\,0.000 & 0.290\,$\pm$\,0.000 \\
\rowcolor{qgReferenceTint}Stratified random & 0.501\,$\pm$\,0.005 & 0.501\,$\pm$\,0.005 & 0.710\,$\pm$\,0.000 & 0.501\,$\pm$\,0.000 & 0.412\,$\pm$\,0.000 & 0.000\,$\pm$\,0.000 \\
\bottomrule
\end{tabular}
% a = best model on mean macro-F1;
% b = significantly below the best (Holm-corrected p < 0.05).

\begin{tablenotes}[flushleft]
\footnotesize
\setlength{\labelsep}{0pt}
\item[] $^{a}$Best model in the regime. $^{b}$Significantly below the best (Holm-corrected Wilcoxon
test, $p<0.05$). QSVM rows are deterministic across seeds and are compared by margin. Pink
rows: quantum models; gray: chance baselines.\ifarxivbuild{} The stratified-random floor is shown
because several models sit near it.\fi{} AUPRC: area under the precision-recall curve; ECE:
expected calibration error.
\end{tablenotes}
\end{threeparttable}
\end{table*}
\fi
\begin{table}[!t]
\centering
\begin{threeparttable}
\caption{RQ1: test macro-F1 in the matched contest (binary task, row-level split, means over
seeds)}
\label{tab:rq1dims}\ifarxivbuild\else\label{tab:rq1}\fi
\footnotesize
% auto-generated by qgridbench.eval.tables -- do not edit by hand
\begin{tabular}{lrrr}
\toprule
 & \multicolumn{3}{c}{Dimensions (qubits)} \\
Model & 8 & 12 & 16 \\
\midrule
Logistic regression & 0.511$^{b}$ & 0.511$^{b}$ & 0.513$^{b}$ \\
RBF-SVM & 0.578$^{b}$ & 0.575$^{b}$ & 0.581$^{b}$ \\
Random forest & 0.586$^{b}$ & 0.599$^{b}$ & 0.598$^{b}$ \\
XGBoost & 0.597$^{a}$ & 0.606$^{a}$ & 0.607$^{a}$ \\
LightGBM & 0.593$^{b}$ & 0.604$^{b}$ & 0.606 \\
MLP & 0.522$^{b}$ & 0.473$^{b,c}$ & 0.459$^{b,c}$ \\
\rowcolor{qgQuantumTint}QSVM (RY-angle) & 0.527 & 0.541 & 0.574 \\
\rowcolor{qgQuantumTint}QSVM (ZZ) & 0.564 & 0.564 & 0.567 \\
\rowcolor{qgQuantumTint}VQC & 0.514$^{b}$ & 0.510$^{b}$ & -- \\
\rowcolor{qgReferenceTint}Majority class & 0.415 & 0.415 & 0.415 \\
\rowcolor{qgReferenceTint}Stratified random & 0.501 & 0.501 & 0.501 \\
\bottomrule
\end{tabular}
% a = best model at that dimensionality; b = significantly below it (Holm-corrected
% p < 0.05); c = the unweighted MLP fit.

\begin{tablenotes}[flushleft]
\footnotesize
\setlength{\labelsep}{0pt}
\item[] $^{a}$Best model at that dimensionality. $^{b}$Significantly below it (Holm-corrected
Wilcoxon test over ten seeds, $p<0.05$). A dash: not run. Logistic regression,
the RBF-SVM, and the QSVM do not vary with the seed; the QSVM rows are compared by margin.
Seed standard deviations: trees under 0.003, MLP 0.015--0.044, VQC 0.007--0.009. $^{c}$The unweighted MLP; the class-weighted fit was run
at eight dimensions only. Pink: quantum models; gray: the constant-predictor (majority class) and
stratified-random floors.\ifarxivbuild{} Table~\ref{tab:rq1} and
Appendix~\ref{app:rq1dims} give six metrics with their seed spreads.\fi
\end{tablenotes}
\end{threeparttable}
\end{table}

\ifarxivbuild
\begin{figure}[!t]
\centering
\includegraphics[width=\columnwidth]{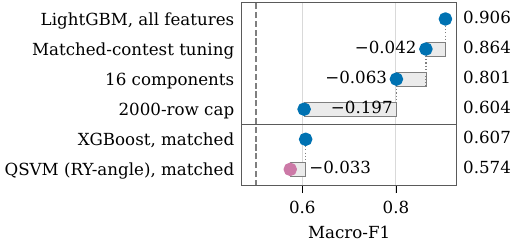}
\caption{Where the accuracy goes between the full-feature setting and the matched contest
(MSU/ORNL binary task, row-level split). Gray bar: the cost of a step; marker: the score it
leaves; dashed: the stratified-random floor. The top four rows are LightGBM at one seed, from a
separate run (0.906 here, 0.905 in Fig.~\ref{fig:checklist}), and each step changes one thing: the hyperparameters\ifarxivbuild{} (from the full-feature set to the
set tuned at PCA-8 under the cap)\fi, the features, then the training rows\ifarxivbuild{}
(\NumTrainRows{} to 2000)\fi. \ifarxivbuild The matched-contest scores\else Matched-contest scores\fi{} are ten-seed means.}
\label{fig:bridge}
\end{figure}
\fi

Under the row-level split, tuned XGBoost beats the best quantum kernel by a small but real margin
(\ifarxivbuild Tables~\ref{tab:rq1} and~\ref{tab:rq1dims}\else Table~\ref{tab:rq1dims}\fi; min-max scaling): 0.032--0.042 macro-F1 before rounding,
and a paired bootstrap excludes zero each time. The margin is not an accident of the 2000 training
rows drawn: \ifarxivbuild over ten draws at each dimensionality (the published one and nine more), with the tuned
settings held fixed, XGBoost leads every time, by\else over ten draws at each dimensionality (the published one and nine more), with tuned settings held fixed, XGBoost leads every time, by\fi{}
0.023--0.055\ifarxivbuild{} (means 0.030, 0.040, and 0.037 at 8, 12, and 16 dimensions against the published best
kernel; both models' scores move by about 0.02 with the draw)\appxsec{app:further}\fi.\ifarxivbuild\else{} Section~\ref{sec:preempt}
gives what the 2000-row cap costs a classical model.\fi
\ifarxivbuild
\subsubsection{The Cap Costs More Than the Gap}
On all 128
features LightGBM reaches 0.906 (Fig.~\ref{fig:bridge}). Moving it into the matched regime at 16 components costs 0.302
macro-F1 in three steps, each with everything else fixed (one seed): the hyperparameters tuned in
the PCA regime cost 0.042, PCA 0.063, and the fidelity kernel's 2000-row cap, which we apply to
classical models too, 0.197. The cap alone is about six times the quantum--classical gap at that
dimensionality under the min-max scaling (about eight under the quantile scaling). With the full-feature hyperparameters held fixed instead, PCA costs 0.051 and the
cap 0.257\appxsec{app:further}. Quantum kernels come close to classical models only inside the
regime their own sample complexity forces on both.
\fi
\ifarxivbuild
\subsubsection{A Positive Control Shows the Pipeline Is Not Blind}
On the engineered labels of
Section~\ref{sec:models}~\cite{huang2021power,egginger2024hyperparameter}, the pipeline detects a
quantum win in all eight kernel configurations (six binary, two three-class; $+0.004$ to $+0.092$ over freshly retuned
classical models, five resamples; the weakest is marginal). The win grows with the logarithm of the geometric
difference $g$ ($r = +0.85$, $p = 0.007$). Thus the pipeline is not blind to a quantum win.
\subsubsection{The Real Labels Align Weakly With Every Kernel}\label{sec:diag}
Under the min-max scaling the quantum kernels' KTA
is at most 0.015 on Gram matrices where the engineered labels reach 0.046--0.5. The comparison
RBF kernel also aligns weakly (0.005--0.019), so under the min-max scaling
low alignment marks the task at these dimensionalities and does not by itself explain the classical
lead. On the real
labels KTA, not the geometric difference $g$, predicts a quantum kernel's accuracy. Across 54 bandwidth
configurations it follows validation macro-F1 for every feature map and dimensionality
(Spearman $+0.65$ to $+0.97$), while the correlation of $g$ with macro-F1 changes sign between
them, because spectral degeneracy inflates $g$ at
both bandwidth extremes (pooled $\rho = +0.8$). The low alignment is not exponential
concentration~\cite{thanasilp2024exponential}, which we measured and did not find:
off-diagonal variance falls only 0.05--0.09 $\log_2$ per added qubit, against $-1.0$ under
concentration.
\else
\subsubsection{A Positive Control Shows the Pipeline Is Not Blind}\label{sec:diag}
On engineered labels that follow each quantum kernel's geometry
(Section~\ref{sec:models})~\cite{huang2021power}, the pipeline detects a quantum win in all
eight kernel configurations, six binary and two three-class. The wins are $+0.004$ to $+0.092$
over freshly retuned classical models (five resamples; the weakest is marginal), so the
pipeline is not blind to one. On the real
labels every kernel aligns weakly: under the min-max scaling the quantum kernels' KTA is at most 0.015, and that of the
comparison RBF kernel is 0.005--0.019, so alignment alone does not explain the classical lead. The low
alignment is not exponential concentration~\cite{thanasilp2024exponential}, which we measured
and did not find.
\fi
\ifarxivbuild
\subsubsection{Newer Tabular Models Show the Same Split Effect}
TabICLv2~\cite{qu2026tabiclv2}, an in-context tabular foundation model with no training or
tuning, an FT-Transformer~\cite{gorishniy2021revisiting} at literature defaults, and
TabM~\cite{gorishniy2025tabm} score 0.63--0.77 under the row-level split, against 0.905 for
tuned LightGBM (Fig.~\ref{fig:tabular}). With whole files held out all four score 0.59--0.60, and the FT-Transformer is
within 0.002 of LightGBM ($0.596 \pm 0.014$ against $0.594 \pm 0.008$ over five partitions). The gap rises
with the row-level score, and LightGBM regularized to the FT-Transformer's row-level score has a
gap of 0.029, so no architecture story is needed.
\fi
\subsubsection{The Angle Scaling Moves the Quantum-Kernel Scores}\label{sec:anglescaling}
The quantum scores of Table~\ref{tab:rq1dims} use the min-max scaling of
Section~\ref{sec:models}. The extremes of heavy-tailed components set its range, so on six
of the eight the central 90\% of the training rows fall in 2--7\% of the angle range.
With each component mapped through its training quantiles instead, and the kernel retuned at the
same budget, the RY-angle QSVM gains 0.053 at eight dimensions and the ZZ map 0.013, under the
0.015 we registered as a real move \ifarxivbuild (0.527 to 0.580 and 0.564 to 0.577; a clipped
min-max scaling moves them by $+0.038$ and $-0.003$)\else (Table~\ref{tab:otherchecks}, Angle scaling)\fi.
XGBoost, which a monotone transform of a feature does not change, then leads the best quantum
kernel by 0.017, under the 0.02 margin by which Section~\ref{sec:replication} calls a lead (a
reading made after the result; at 12 and 16 dimensions the rule was registered first).
At 12 dimensions the two kernels gain 0.040 and 0.019, and at 16 only 0.008 and 0.013, each under
the registered 0.015; XGBoost's lead holds at 12 and 16 (0.023 and 0.025).
The VQC and the other classical models were not rerun, and every attack result below uses the
min-max scaling.
\subsubsection{Answer to RQ1}
Under the row-level split, no quantum model we ran leads. Under the min-max scaling the tuned classical
models lead at every matched dimensionality; under the quantile scaling their lead falls inside
our 0.02 margin at eight dimensions (read after the result) and holds at 12 and 16. On held-out files neither clearly leads
(Section~\ref{sec:split}), and we interpret them as too close to each other to rank.

\ifarxivbuild
\begin{figure}[!t]
\centering
\includegraphics[width=\columnwidth]{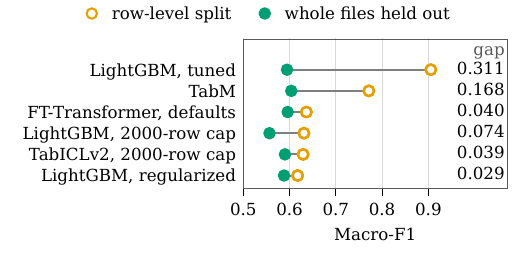}
\caption{Newer tabular models under both splits (MSU/ORNL binary task, all 128 features).
TabICLv2 and the LightGBM beside it train on 2000 rows. The regularized LightGBM is the capacity
setting nearest the FT-Transformer's row-level score. Held-out scores are means over five file
partitions.}
\label{fig:tabular}
\end{figure}
\fi

\subsection{Choice 2: The Surrogate's Fit Sets the Transfer Ratio}\label{sec:surrogate}
Unless stated, the MSU/ORNL attack results of Sections~\ref{sec:surrogate} to~\ref{sec:blackbox} describe
eight-dimension detectors trained under the row-level
split (Section~\ref{sec:limitations}). They are weak detectors: on the attacked rows they score
0.47--0.63 clean, against a stratified-random floor of 0.50 (Fig.~\ref{fig:controls}(a)), and
the two kernel machines and the three tree models flag 54--65\% of natural rows before any
attack. A high retention can therefore mean a score at chance.
\subsubsection{An Apparent Asymmetry}
At $0.5\sigma$, attacks crafted on the
unweighted MLP take 4.6 times as much macro-F1 from the VQC as attacks crafted on the VQC take from that
MLP\appxsec{app:surrogate}. We call this number the \emph{transfer ratio}.
\subsubsection{The Cause Is the Surrogate's Fit}
\ifarxivbuild Our MLP is scikit-learn's \texttt{MLPClassifier},
which takes no \texttt{class\_weight} argument, so it alone among the models was fitted
with no imbalance handling on a 71\%-positive task.\else The unweighted MLP alone among the
models was fitted with no imbalance handling (Section~\ref{sec:models}), on a 71\%-positive
task.\fi{} It predicts the positive class 96.6\% of the
time and scores 0.031 below the stratified-random floor, so an attack finds few of its predictions
left to change\ifarxivbuild: VQC-crafted examples damage it by only 0.058\fi.
The control is a class-weighted MLP: it does the same damage to the classical targets
(0.204 against 0.208)\ifarxivbuild, and the damage its examples do to the VQC falls from 0.268
to 0.211. The damage it takes from VQC-crafted examples rises to 0.242, so\else, and\fi{} the transfer ratio falls to
$0.9\times$ (Fig.~\ref{fig:checklist}, row 2).\ifarxivbuild{} A second, independently
written implementation of that fit gives $1.0\times$.\fi
\ifarxivbuild
\subsubsection{A Dose--Response}
With a surrogate of a different function class, the tuned RBF-SVM,
in the MLP's place, the transfer ratio is $0.006\times$. Ordered by how far each surrogate clears
the stratified-random floor on the attacked rows ($-0.031$, $+0.035$, $+0.090$), the ratio declines monotonically,
also after dividing by the damage each surrogate does to the classical targets ($1.29 \to 1.04 \to 0.008$).
\fi
\subsubsection{What the Ratio Measures}
\ifarxivbuild\else A third surrogate, the tuned RBF-SVM, sits furthest above the stratified-random floor
of the three and gives $0.006\times$. \fi The ratio's direction depends on the surrogate, so it measures the
surrogate's competence, and we claim no transfer asymmetry in either
direction; with the RBF-SVM surrogate, function class and competence cannot be
separated.\ifarxivbuild{} The mechanism of the artifact stays open: a shift in the surrogate's
class prior is measured false\appxsec{app:surrogate}.\fi{} All transfer results here are for an
attacker who ignores the grid's physical relations (Section~\ref{sec:physics}).

\subsection{Choices 3 to 5: Attack Availability Moves the Robustness Ranking}\label{sec:rq2}

\begin{figure*}[!t]
\centering
\ifarxivbuild
\includegraphics[width=\textwidth]{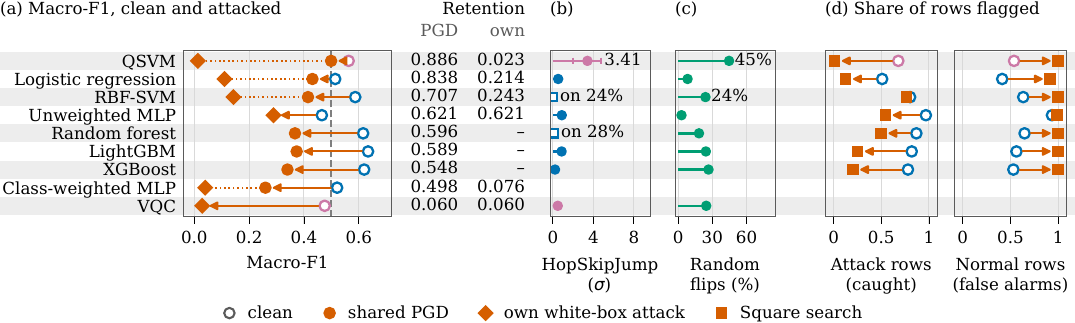}
\caption{Robustness at $\varepsilon = 0.5\sigma$, read with its controls (binary task, eight
dimensions). (a) PGD on a 500-row stratified test subset, ten seeds; rows in order of retention
under PGD. The two columns give the retention under PGD and under the model's own white-box
attack. Dashed: the stratified-random floor; the QSVM's transferred reading sits on it, and the VQC
is below it before any attack. PGD is white-box for the unweighted MLP (the surrogate) and the
VQC, and transferred from that MLP to every other row. A dash: the model admits no
gradient-based white-box attack, so its retention is an upper bound. The kernel machines' and
logistic regression's own attacks: separate single-model runs (ten attack starts each), read
at the worst radius inside the budget ($0.3\sigma$ and $0.1\sigma$ for the
kernel machines, $0.5\sigma$ for logistic regression); the class-weighted
MLP's ran at $0.5\sigma$ only. (b), (c) On 100 test points: the decision-based attack
(mean $\pm$ std over restarts: ten for QSVM, MLP, LightGBM, and XGBoost, five
otherwise) and its random-noise control (three draws, one
seed); not measured for the class-weighted MLP. Hollow square in (b): the attack found a starting point
on that share of the points only. (d) Clean and under the Square
search~\cite{andriushchenko2020square} (500 test rows, three seeds). The class-weighted MLP was
not run, and the VQC's per-class reading predates its final configuration and is not drawn.}
\else
\includegraphics[width=\textwidth]{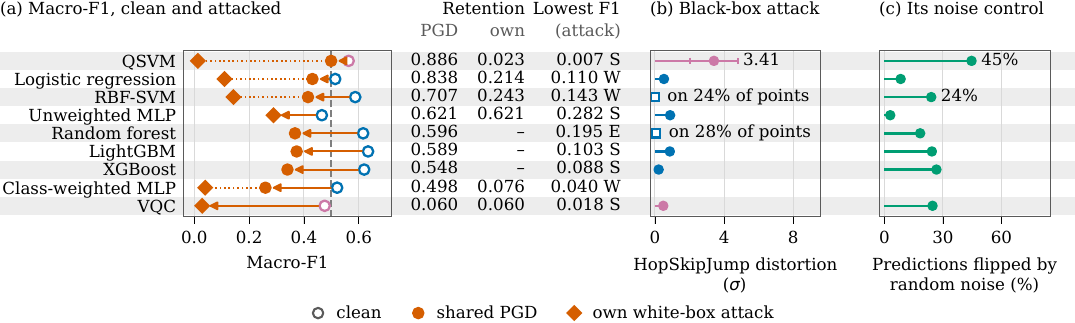}
\caption{Robustness under the shared protocol at $\varepsilon = 0.5\sigma$ (binary task, eight
dimensions; 500 test rows, so clean scores differ from Table~\ref{tab:rq1dims}; ten-seed means). (a) Macro-F1 clean,
under PGD, and (diamond) under the model's own white-box attack at its worst tested radius
(kernel machines and logistic regression: separate single-model runs; class-weighted MLP: $0.5\sigma$ only, an upper
bound). PGD is transferred from the unweighted MLP except on that MLP and the VQC. Columns: the two retentions and the lowest macro-F1 over the PCA-space attacks run (P: PGD at $0.5\sigma$; W: own white-box; S: Square search; E: evolution strategies; W, S, E at their worst tested radius; S, E: three seeds); dashed: the stratified-random floor; a dash: the model admits no gradient-based white-box attack,
so the retention is an upper bound. (b) HopSkipJump median
distortion on 100 rows, mean $\pm$ std over restarts; a hollow square: the attack started on
that share of the points only. (c) The same rows under random noise of $0.5\sigma$; (b), (c), S, and E were not measured
for the class-weighted MLP.}
\fi
\label{fig:controls}
\end{figure*}

\ifarxivbuild
\begin{table*}[!t]
\centering
\begin{threeparttable}
\caption{Robustness at $\varepsilon=0.5\sigma$, read with its controls (binary task, eight
dimensions, 500-row stratified test subset, ten seeds)}
\label{tab:robust}
\setlength{\tabcolsep}{4.5pt}% nine columns: the default gap overfills the text width
% auto-generated by qgridbench.eval.tables -- do not edit by hand
\begin{tabular}{lrrrrrrrr}
\toprule
Model & Clean F1 & PGD@0.5$\sigma$ & F1$-$floor & Retention & White-box retention & Lowest F1 & Random flip rate & HSJ median $\ell_\infty$ ($\sigma$) \\
\midrule
\rowcolor{qgQuantumTint}QSVM & 0.563 & 0.499 & $-$0.001 & 0.886 & 0.023 (0.3$\sigma$) & 0.007 (S) & 45\% & 3.413\,$\pm$\,1.386 \\
Logistic regression & 0.514 & 0.431 & $-$0.069 & 0.838 & 0.214 & 0.110 (W) & 8\% & 0.520\,$\pm$\,0.009 \\
RBF-SVM & 0.587 & 0.415 & $-$0.085 & 0.707 & 0.243 (0.1$\sigma$) & 0.143 (W) & 24\% & 0.013\,$\pm$\,0.001 (24\%) \\
MLP (unweighted) & 0.465 & 0.289 & $-$0.211 & 0.621 & 0.621 & 0.282 (S) & 3\% & 0.877\,$\pm$\,0.015 \\
Random forest & 0.616 & 0.367 & $-$0.133 & 0.596 & --- & 0.195 (E) & 18\% & 0.062\,$\pm$\,0.042 (28\%) \\
LightGBM & 0.634 & 0.373 & $-$0.127 & 0.589 & --- & 0.103 (S) & 24\% & 0.859\,$\pm$\,0.120 \\
XGBoost & 0.620 & 0.340 & $-$0.160 & 0.548 & --- & 0.088 (S) & 27\% & 0.212\,$\pm$\,0.048 \\
MLP (class-weighted) & 0.521 & 0.259 & $-$0.241 & 0.498 & 0.076 & 0.040 (W) & --- & --- \\
\rowcolor{qgQuantumTint}VQC & 0.476 & 0.029 & $-$0.472 & 0.060 & 0.060 & 0.018 (S) & 25\% & 0.481\,$\pm$\,0.067 \\
\bottomrule
\end{tabular}

\begin{tablenotes}[flushleft]
\footnotesize
\setlength{\labelsep}{0pt}
\item[] Pink rows: quantum models. Retention ranks the QSVM first, but its 0.886 is a score at
chance ($-0.001$ from the stratified-random floor), and the VQC is below chance on this subset
before any attack (0.476 against 0.500). The random flip rate and HSJ columns use 100 test points;
the random noise is three draws at one seed.\ifarxivbuild{} Clean macro-F1 differs from
Table~\ref{tab:rq1} because the attack columns use stratified test subsets.\fi{} PGD is white-box
for the unweighted MLP and the VQC, and transferred from the unweighted MLP for every other row,
the class-weighted MLP included. F1$-$floor: macro-F1 under PGD minus the stratified-random floor.
Lowest F1: the lowest macro-F1, at up to $0.5\sigma$, over the attacks on the model in the
standardized PCA space that Table~\ref{tab:attacks} lists, at their reported settings, with the
attack in parentheses (P: the shared PGD at $0.5\sigma$; W: its own white-box attack; S: the Square
search~\cite{andriushchenko2020square}; E: the evolution-strategies
attack~\cite{ilyas2018blackbox}, run on the trees and the unweighted MLP). W, S, and E are read
at their worst tested radius; S and E were tested at 0.1, 0.3, and $0.5\sigma$ with about 600
queries per point. Each attack is a mean over its own seeds: ten for PGD, three for S and E.
The VQC's Square reading is from a run on this table's configuration; the class-weighted MLP
received neither S nor E. XGBoost's fit moves with the thread count, so its S and E runs
attacked fits whose clean scores differ from this table's by 0.004 to 0.018 per seed.
Dashes: no white-box attack was run, so those retentions are upper bounds; for the class-weighted
MLP, the control was not measured. White-box retention is read at the worst tested radius inside
$0.5\sigma$, printed in parentheses where it is not $0.5\sigma$ itself; read at $0.5\sigma$ the
QSVM keeps 0.064 and the RBF-SVM 0.685. The two kernel machines' readings are from a second run
and logistic regression's from a third (one fitted model, ten attack starts each). The class-weighted MLP's own attack ran at $0.5\sigma$
only, so its cell is an upper bound on its worst case. Random flip rate: share of predictions flipped by random noise
of the same radius. HSJ: mean$\,\pm\,$std over five restarts, or ten for the QSVM, the MLP, LightGBM, and XGBoost\ifarxivbuild{}
(Section~\ref{sec:blackbox})\fi. The ten come from two runs (August and November 2026); the XGBoost
fit depends on the thread count and differs slightly between them. A share in parentheses is the
share of the 100 points on which the attack found a starting point, and that cell is read over
those points. The other cells are read over all 100 points; where the share was recorded, the
attack found a starting point on at least 98\% of them.
\end{tablenotes}
\end{threeparttable}
\end{table*}
\fi

\ifarxivbuild
\begin{figure}[!t]
\centering
\includegraphics[width=\columnwidth]{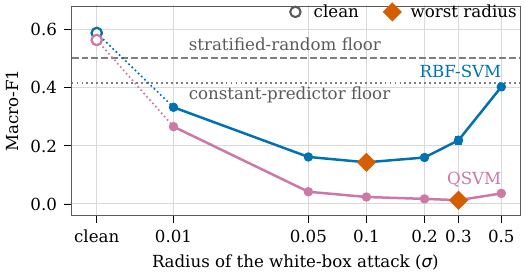}
\caption{Two kernel machines under their own white-box attack at each tested radius inside
the $0.5\sigma$ budget (MSU/ORNL binary task, eight dimensions; mean over ten attack
starts, whose standard deviation is smaller than the markers). The lowest score lies inside the budget. At the budget the RBF-SVM returns to within 0.013
of the constant-predictor floor (dotted), because its output is nearly constant. Dashed: the
stratified-random floor.}
\label{fig:radius}
\end{figure}
\fi

Under PGD at $\varepsilon = 0.5\sigma$
(Fig.~\ref{fig:controls}(a)\ifarxivbuild{} and Table~\ref{tab:robust}\fi), retention
ranks the QSVM first, at 0.886, and the VQC last, so the two quantum models seem to sit at
opposite ends. The ranking shows which models the protocol can attack, not which resist: the
QSVM receives only a transferred attack, which leaves its score at the stratified-random floor.
\subsubsection{Attack Availability (Choice 3)}
Differentiating the kernel through the simulator
cuts the QSVM's retention to 0.023, at $0.3\sigma$\ifarxivbuild{} (macro-F1 $0.013 \pm 0.006$)\fi. In that same
single-model run the transferred attack leaves \ifarxivbuild 0.917\else the score near the stratified-random floor\fi{} (Fig.~\ref{fig:checklist}, row 3); the
shared protocol's number above is a mean over ten seeds of the surrogate. A competent surrogate does not close the gap:
crafted on the class-weighted MLP or on the RBF-SVM, the transferred attack leaves the QSVM 0.74
and 0.70 of its clean score\appxsec{app:surrogate}.\ifarxivbuild{} The transfer ordering itself depends
on the surrogate: with an XGBoost surrogate the QSVM keeps 0.691, below logistic regression
(0.940) and the unweighted MLP (0.858).\fi
\subsubsection{The Matched Classical Kernel}
\ifarxivbuild The identical attack leaves the tuned RBF-SVM 0.243 of
its clean score, at $0.1\sigma$ (paired over ten attack starts, each model at its
worst radius: $p = 0.002$, rank-biserial $-1.0$). Both kernel machines end below the
stratified-random floor under their own white-box attacks, and under this 20-step attack the QSVM
scores the lower of the two at every tested radius.\else The identical attack leaves the tuned RBF-SVM a retention of 0.243, at
$0.1\sigma$. Both kernel machines end below the stratified-random floor, the QSVM the lower of
the two (Fig.~\ref{fig:controls}(a)).\fi
\subsubsection{No Ranking of Quantum Against Classical}
White-box attacks break classical models
too: the class-weighted MLP keeps 0.076 at $0.5\sigma$\ifarxivbuild, the only radius at which its attack
ran (an upper bound on its worst case; Table~\ref{tab:robust})\else{} (Fig.~\ref{fig:controls}(a))\fi. Low white-box
retention is therefore not a quantum trait, and the two kernel machines are the only matched
white-box pair we have. Under the shared protocol all 12
quantum--classical pairs differ in attacked macro-F1 after Holm correction, in both directions, so
we make no family-level robustness claim.\ifarxivbuild{} The white-box contrast survives whole-file
training. Refit on whole-file splits and attacked identically, the QSVM lands 0.47--0.49 below the
stratified-random floor and the RBF-SVM 0.35--0.36 below, each at its worst radius. The QSVM
is the lower in 20 of 20 attack seeds over two partitions.\fi
\subsubsection{Models Given No White-Box Attack}
Tree families admit no gradient-based white-box attack, and we ran no tree-specific
one~\cite{kantchelian2016evasion}, so their
retention figures are upper bounds, as the QSVM's transferred figure was. Such bounds are loose: logistic regression, itself near the
stratified-random floor, keeps 0.214 under its own attack, a quarter of what the transferred
attack reported (Fig.~\ref{fig:controls}(a)). Score-based black-box attacks, at about 600 queries per
point, leave all three tree models below the stratified-random floor, XGBoost at 0.088
(\ifarxivbuild Table~\ref{tab:robust}\else Fig.~\ref{fig:controls}(a)\fi).\ifarxivbuild{} The equal-budget evolution-strategies
attack~\cite{ilyas2018blackbox} does 0.077--0.172 more damage to the trees at $0.5\sigma$ than the transferred attacks
report (three seeds against the ten-seed transfer damage).\fi
\ifarxivbuild
\subsubsection{Two End States}
An attacked model can
end \emph{inverted}, with macro-F1 far below the constant-predictor floor (about 0.415, the score
of a detector that labels every row an attack), or \emph{disabled}, with constant
predictions and macro-F1 at that floor. We name a state only when it holds over three or more
tested radii (9 of 114 recorded configurations change label under a single-radius reading).
\fi
\subsubsection{Attack Objective (Choice 4)}
The white-box attack above ascends the classification
loss and \ifarxivbuild inverts the QSVM: 0.013 at its worst radius\else\emph{inverts} the QSVM, to 0.013
macro-F1, far below the constant-predictor floor (Fig.~\ref{fig:checklist}, row 4)\fi. \ifarxivbuild A \emph{distance-seeking} attack instead
pushes each input away from the model's support vectors in kernel space. It never takes\else The \emph{distance-seeking} attack
of Section~\ref{sec:attackprotocol} never takes\fi{} the QSVM below the
constant-predictor floor at any tested radius up to $2\sigma$\ifarxivbuild; its lowest score is 0.441, at
$2\sigma$ (Fig.~\ref{fig:checklist}, row 4)\fi. One
white-box attacker produces two failures, confidently wrong labels or one label for
96\% of rows, and an attacked score describes the objective as much as the
model.\ifarxivbuild{} Inside $0.5\sigma$ the distance-seeking attack leaves 0.49 to 0.57. In the first run of this comparison (three attack starts, $2\sigma$) the
distance-seeking objective cuts the QSVM's retained kernel mass $6.2\times$ further, to 0.027 of
clean\appxsec{app:objreg}.\fi
\subsubsection{Reading Radius (Choice 5)}
An attacker allowed a radius can use any smaller
one, so we report each white-box attack at its worst tested radius inside the
budget~\cite{carlini2019evaluating}\ifarxivbuild{} (Fig.~\ref{fig:radius})\fi. Reported at the $0.5\sigma$ bound instead, the RBF-SVM keeps 0.685
(Fig.~\ref{fig:checklist}, row 5) and the QSVM 0.064. At the bound the RBF-SVM's output is nearly
constant (within 0.013 of the constant-predictor floor), which scores higher than its inverted predictions at $0.1\sigma$, so a reading at the
bound alone flatters a detector that a smaller perturbation inverts.\ifarxivbuild{} Both
kernels score lower at the worst radius than at the bound in ten of ten attack starts. The MLP's
and the VQC's curves fall with the radius, so their two readings agree.\fi
\ifarxivbuild
\subsubsection{Kernel Width and Regularization}\label{sec:widthreg}
We also varied the RBF
kernel's width and the QSVM's regularization constant $C$, and neither prevents the inversion.
At its worst radius every RBF width is inverted (0.27 to 0.42 below the
constant-predictor floor)\appxsec{app:lengthscale}, and so is the QSVM at every $C$ from 1 to the
tuned 835 (0.37 to 0.40 below, under loss ascent at $0.3\sigma$). Both settings change what the
reading at the largest radius, $2\sigma$, shows, so neither is a checklist item. The full-budget reading of $C$ is this. At $2\sigma$ under
the distance-seeking attack, the attacked decision spread runs from 0.11 to 0.70 of clean as $C$
runs from 1 to the tuned 835 (ten attack starts)\appxsec{app:objreg}. The macro-F1 stays within
0.03 of the constant-predictor floor at every $C$. On 2000 validation rows $C = 10$ costs 0.069.
The random-noise control at the worst radius leaves 0.49 to 0.53 at every $C$, so the inversion
is the attack's work.
\fi
\subsubsection{Retention Is Not Attack Recall}
Retention ranks what a detector keeps of its score,
not which attacks it still catches: under the shared protocol the two orders barely agree
(Spearman 0.10; Fig.~\ref{fig:replication}, row C). Retention also hides the form of the
failure: the transferred attack lowers the three tree models' attack recall by only 3--8 points
and raises the RBF-SVM's, and it makes all four flag 98--100\% of natural rows.\ifarxivbuild{} Under the strongest common
attacker, the score-based Square search~\cite{andriushchenko2020square}, the QSVM's attack recall
falls from 0.68 to 0.01, the trees' by 37--58 points, and the RBF-SVM's by four points while it
flags every normal row (Fig.~\ref{fig:controls}(d)).\fi

\subsection{Choices 6 and 7: At About 600 Queries per Point, the Decision-Based Black-Box Attack Is Itself Fooled}\label{sec:blackbox}
HopSkipJump at identical settings reports the QSVM as the
hardest target in nine of its ten restarts: its median $\ell_\infty$ distortion is $(3.41 \pm 1.39)\sigma$, against
$0.01$--$0.88\sigma$ for classical models (Fig.~\ref{fig:controls}(b)).\ifarxivbuild{} For the
RBF-SVM and the random forest the attack found a starting point on only 24\% and 28\% of the
points. On the others none of its 100 random draws left the class the model already predicted, and
the library returns such an input unchanged. Their distortions, $0.013$ and $0.062\sigma$, are read
over the points it could start on, and all but a few of those are points the model labels
normal.\fi
\subsubsection{Starting Point (Choice 6)}
\ifarxivbuild The attack starts from an unseeded random point in the
library we used\else The attack's starting point is unseeded
(Section~\ref{sec:attackprotocol})\fi, so one run is one draw: its answer for the QSVM moves from $0.57$ to
$4.99\sigma$ over ten restarts (Fig.~\ref{fig:checklist}, row 6).\ifarxivbuild{} A single run can thus place the
QSVM inside the classical range or far above it. We report it as a mean and
spread over restarts. That median mixes two groups: the 33 rows the QSVM labels natural
(median $0.02\sigma$) and the 67 it labels attack ($6.45\sigma$ in five new
restarts at the same settings; Table~\ref{tab:hsjsweep}). It is the latter's lower quartile (the
mean of the 17th and 18th smallest of the 67 distortions when the attack flips all 100 rows) and
varies 42\% across those restarts; that group's own median varies 13\% (standard
deviation over mean).\else{} The median mixes two groups of rows:
the 33 that the QSVM labels natural, which a very small perturbation flips (median
$0.02\sigma$), and the 67 that it labels attack ($6.45\sigma$ in five more restarts), both
read after the result of the sweep below.\fi
\subsubsection{Measurement Geometry (Choice 7)}
Random $\ell_\infty$ noise of $0.5\sigma$
flips 45\% of the QSVM's predictions\ifarxivbuild, the most of any model, against 24\% for the
RBF-SVM\fi{} (Fig.~\ref{fig:controls}(c)). It flips 28 of those 67 rows in at least two of three
draws, so each has an adversarial point \ifarxivbuild at $0.5\sigma$; the attack places 22 to 27 of the 28
farther away. On those rows its distortions are not the smallest ones
(Fig.~\ref{fig:checklist}, row 7). A weaker rule reads the same way: at least one draw
flips 51 of the 67 rows, and the attack places 41 to 45 of them farther away. Every one of
these noise points lies inside the box that the attack searches. The control replicates on held-out files (10 of 10
partitions, QSVM-to-RBF-SVM flip ratio 1.54--2.26), while the distortion magnitude varies $40.7\times$
across partitions with no covariate we could find.\else within $0.5\sigma$, and a
minimum-distortion attack should report one at least that close. It does not: the attack places
22 to 27 of the 28 farther away (Fig.~\ref{fig:checklist}, row 7). Here the attack, not a
defense, produces the false sense of security~\cite{athalye2018obfuscated}. In a sweep to about 10\,900 queries per point, our registered rule reads the
600-query result as a budget artifact: the median over both groups, $0.42\sigma$, is under the
registered threshold, the $0.88\sigma$ above. Read row by row after that result, as the count
above was, the attack still places 18 to 22 of the 28 farther away, which suggests the error
remains.\fi
\ifarxivbuild
\subsubsection{Larger Query Budgets (Choices 6 and 7)}
We registered a sweep of the query budget before running it, with its rule read at the
largest budget: five new restarts of the QSVM, the MLP, and LightGBM at about 2800 and
10\,900 queries per point and, by an amendment made before any result at the largest budget
existed, five new restarts at about 600 (Table~\ref{tab:hsjsweep}). At about 10\,900 queries per
point the attack still places 18 to 22 of the 28 beyond $0.5\sigma$ and puts the median of the 67 rows at
$4.08\sigma$ (the MLP's 94 such rows: $0.74\sigma$; LightGBM's 74: $0.59\sigma$). The
registered median over both groups falls from 2.70 to $0.42\sigma$, under the MLP's
$0.72\sigma$ and under the registered threshold of $0.88\sigma$, the top classical median of
Table~\ref{tab:robust}: by our registered rule, a budget artifact. The readings by the
model's clean label and row by row against the noise, here and above, were made after that
result. They suggest that the error remains: the median over both groups fell because the attack
now places 28\% of the 67 rows within $0.5\sigma$, up from 15\%, and the statistic reads
near that group's lower quartile. The attack-labeled rows' own median fell by a factor of 1.6 (the
MLP's by 1.2, LightGBM's by 2.5). These readings use fields that the registered runs stored
and a noise run made earlier. They were not registered, so we give them as exploratory.
LightGBM's five restarts at about 600 queries read higher than its ten published ones (1.07
against $0.86\sigma$ over both groups), a difference we have not explained, so apart from that
threshold we compare the sweep's budgets with each other and not with
Table~\ref{tab:robust}. The attack's distortions are not all the smallest ones for LightGBM
either: at least one noise draw flips 13 of its 74 attack-labeled rows, and at the largest
budget the attack places 6.2 of them, on average, farther than $0.5\sigma$.
\begin{table*}[!t]
\centering
\begin{threeparttable}
\caption{HopSkipJump at three query budgets, read by the model's clean label (binary task,
eight dimensions, 100 test rows, five restarts; first three blocks: median $\ell_\infty$
distortion in $\sigma$, mean\,$\pm$\,std; last two blocks: mean count of rows)}
\label{tab:hsjsweep}
% auto-generated by qgridbench.eval.tables -- do not edit by hand
\begin{tabular}{lrrr}
\toprule
Queries per point & 594--626 & 2725--2782 & 10\,777--10\,880 \\
\midrule
\multicolumn{4}{l}{\emph{Median over both groups (registered)}} \\
QSVM & 2.70\,$\pm$\,1.13 & 1.71\,$\pm$\,1.00 & 0.42\,$\pm$\,0.16 \\
MLP & 0.87\,$\pm$\,0.01 & 0.77\,$\pm$\,0.01 & 0.72\,$\pm$\,$<$\,0.01 \\
LightGBM & 1.07\,$\pm$\,0.12 & 0.35\,$\pm$\,0.07 & 0.18\,$\pm$\,0.05 \\
\midrule
\multicolumn{4}{l}{\emph{Rows the model labels attack}} \\
QSVM (67 rows) & 6.45\,$\pm$\,0.85 & 5.18\,$\pm$\,0.30 & 4.08\,$\pm$\,0.30 \\
MLP (94 rows) & 0.89\,$\pm$\,0.01 & 0.79\,$\pm$\,0.01 & 0.74\,$\pm$\,$<$\,0.01 \\
LightGBM (74 rows) & 1.45\,$\pm$\,0.06 & 0.87\,$\pm$\,0.20 & 0.59\,$\pm$\,0.25 \\
\midrule
\multicolumn{4}{l}{\emph{Rows the model labels natural}} \\
QSVM (33 rows) & 0.020\,$\pm$\,0.003 & 0.015\,$\pm$\,0.001 & 0.012\,$\pm$\,0.001 \\
MLP (6 rows) & 0.145\,$\pm$\,0.004 & 0.130\,$\pm$\,$<$\,0.001 & 0.122\,$\pm$\,$<$\,0.001 \\
LightGBM (26 rows) & 0.024\,$\pm$\,0.009 & 0.015\,$\pm$\,0.004 & 0.013\,$\pm$\,0.003 \\
\midrule
\multicolumn{4}{l}{\emph{Attack-labelled rows that noise flips in at least two of its draws: how many the attack places beyond $0.5\sigma$}} \\
QSVM (28 rows) & 24.6 & 20.6 & 20.4 \\
MLP (0 rows) & -- & -- & -- \\
LightGBM (2 rows) & 1.6 & 1.2 & 1.0 \\
\midrule
\multicolumn{4}{l}{\emph{Attack-labelled rows that noise flips in any draw: how many the attack places beyond $0.5\sigma$}} \\
QSVM (51 rows) & 43.0 & 39.0 & 35.0 \\
MLP (2 rows) & 0.0 & 0.0 & 0.0 \\
LightGBM (13 rows) & 9.4 & 7.4 & 6.2 \\
\bottomrule
\end{tabular}

\begin{tablenotes}[flushleft]
\footnotesize
\setlength{\labelsep}{0pt}
\item[] Distortions are read over the rows the attack flipped: at least 98 of the 100 for the
QSVM, all 100 for the MLP and LightGBM. Only the median over both groups was registered; the other
blocks were read after its result. Noise: three draws of random $\ell_\infty$ noise at
$0.5\sigma$ on the same rows and the same fitted models. A row whose label a draw changes has
an adversarial point at exactly $0.5\sigma$, and every such point lies inside the box the
attack searches. The last two blocks give, as a mean over restarts, how many of those rows the
attack still places beyond $0.5\sigma$. One QSVM restart at the middle budget left one
attack-labeled row unflipped, so its counts are out of 27 and 50. In another restart at that
budget our wrapper ended one step-size search (Section~\ref{sec:attackprotocol}).
\end{tablenotes}
\end{threeparttable}
\end{table*}
\fi
\ifarxivbuild
\subsubsection{A Likely Cause}
HopSkipJump's bisection~\cite{chen2020hopskipjump} between a row and an adversarial point can
stop at a far crossing when its path crosses the decision boundary more than once. Along 26\% of
random perturbation rays inside the $0.5\sigma$ ball the QSVM's prediction changes more than
once, against 17\% for the tuned RBF-SVM\appxsec{app:crossings}. We did not count crossings
along the attack's own, longer paths, and LightGBM's rows show the same contradiction without a
kernel, so this cause remains a likely one. Here the attack, not a defense, produces the false sense of
security~\cite{athalye2018obfuscated}. The Square search agrees: it drives the QSVM to
macro-F1 0.007, the largest damage of all eight models. Decision-based robustness must be
reported over restarts, by the model's clean label, and next to a random-perturbation control.
\subsubsection{Fragility Is High Where Accuracy Is High}\label{sec:tuning}
The QSVM's random-flip rate across its bandwidth
sweep, a separate run, is 42\% at the accuracy optimum against 19--29\% where the model scores
at the stratified-random floor (100 test points, three noise draws, $C$ reselected per bandwidth). A narrower
bandwidth with the same accuracy to 0.001 flips 33\%, and the next wider one, which scores
0.517, flips 40\%, so the highest rate is the optimum's by two rows in a hundred. For
the RBF-SVM the accuracy optimum is not the fragility maximum, and narrower kernels are
monotonically more robust~\cite{fawzi2015analysis}. This sweep is consistent with the
accuracy--robustness tradeoff known for variational models~\cite{nowmi2026sok}, here on the
fidelity kernel's bandwidth.
\fi
\subsubsection{Answer to RQ2}
Quantum and classical models differ under the shared protocol, in
both directions, and that protocol cannot rank them: its ranking follows which models get their
own white-box attack, and how each attack is built and reported (choices 2 to 7).

\subsection{What Reproduces on Other Data}\label{sec:replication}
\ifarxivbuild We repeated the main checks, with hypotheses registered before each run, on the
gas-pipeline control-system dataset~\cite{morris2014icsdata} and on two simulated transmission
grids (IEEE 14-bus and 118-bus) with false data injection (Fig.~\ref{fig:replication}; rows 1 to
7 are the checklist, rows A to D the other findings). Budgets
match the main study, and so does the pipeline, with two gas exceptions described below (the
pressure feature and the precision of one scoring path).\else
Fig.~\ref{fig:replication} gives the registered readings on the replication sources
(Section~\ref{sec:repdata}); budgets match the main study.\fi
\ifarxivbuild\subsubsection{No Quantum Accuracy Advantage (Row A)}\fi
\ifarxivbuild At matched dimensionality, under the min-max scaling, the
best classical and quantum gas models are 0.012 apart before rounding, below
the 0.02 margin our registered rule needs to call a classical lead, so row A reads in part there.
On the grids the best model of each group scores
0.988 or more (differences of 0.008 and 0.004 between deterministic models).\else{} The
best classical and quantum gas
models are 0.012 apart before rounding, below the 0.02 margin our registered rule needs
to call a classical lead, so row A reads in part there; on the grids each group's best model
scores 0.988 or more (Fig.~\ref{fig:replication}, Table~\ref{tab:repdata}). Every row A cell
uses the min-max scaling (MSU/ORNL: 0.017 under the quantile scaling,
Section~\ref{sec:anglescaling}).\fi
\subsubsection{Split Inflation (Row 1)}
\ifarxivbuild The row-level split inflates the score by 0.31 on the
MSU/ORNL dataset, whose adjacent rows are near-copies. \fi On the gas data the split inflation
reaches its threshold by 0.001 (0.051 against 0.05), inside the spread of its five
held-out time blocks (0.013), and on the grids it
reproduces at most in part.
\ifarxivbuild
\subsubsection{The Kernel Contrast Reproduces at Most in Part (Row B)}
\else
\subsubsection{Kernel Contrast (Row B)}
\fi
On the gas data the QSVM ends below the RBF-SVM on all ten attack starts, but by 0.03 in
retention (0.124 against 0.152) where the registered threshold is 0.20\ifarxivbuild. On the
grids it ends below the RBF-SVM by less than their registered threshold (0.08 and 0.07
against 0.10). The
reading radius (row 5) reproduces in part on the gas data, where it moves the RBF-SVM's
retention, the registered statistic, by 0.10 (0.254 at $0.5\sigma$, 0.152 at $0.3\sigma$), and
not on the grids (0.02), against 0.44 on the MSU/ORNL
dataset. It also moves the QSVM on the 118-bus grid, by 0.07 in retention and from a disabled to a
partial end state, and by 0.04 or less on the other two sources. Under the shared protocol,
11 of the 12 quantum--classical pairs differ in attacked macro-F1 after Holm correction on the
gas data, with the quantum model ahead in 5. On the grids 11 of 12 (14-bus) and 7 of 12
(118-bus) differ, and none favors the quantum model: the VQC trails every classical model by
0.12 to 0.37, and the QSVM is within 0.05 of every classical model except logistic
regression, which holds the highest attacked score on both grids (only the MLP and the VQC
receive their own attack under this protocol, so the other scores are upper bounds).\else; on the grids it
also falls short of its threshold. Under the shared protocol, 5 of the 11 significant
quantum--classical pairs favor the quantum model on the gas data and none on the grids.\fi
\subsubsection{The Checklist (Rows 1 to 7)}
On the gas data three of the seven checklist items
reproduce (1, 3, and 4), two in part (5 and 6), and two do not (2 and 7). Item 6 passes its registered rule
there only through a change of model order between restarts\ifarxivbuild{} (the QSVM
and the VQC trade places, and so do two pairs of classical models)\fi. The QSVM's distortion
varies by 12\%, under the registered 15\%, so we
mark it as in part. Item 4 passes there on
black-box rankings. The Square search, which seeks the largest loss, ranks the QSVM the most
damaged. HopSkipJump, which seeks the smallest distortion, ranks it
the second-hardest target, by $0.004\sigma$ over the VQC, inside the spread between restarts
($0.012\sigma$)\ifarxivbuild, so its pass depends on the restart
draw. The gas thresholds were first registered
on an earlier list of eight (Section~\ref{sec:checklist}). Item 2 does not reproduce there
because the tuned surrogate is already well fitted: the refit moves the transfer ratio 1.12
times, from 0.37 to 0.33. We registered a ratio between 0.5 and 2 as no asymmetry and one
outside 0.33 to 3 as a new finding. At 0.331 it sits on that line: the runner, which tests one
third, read it as a new finding. The reading that follows was made after the result. Attacks crafted on the VQC cut
the MLP by 0.41; attacks crafted on the MLP cut the VQC, which starts 0.07 above the
stratified-random floor, by 0.14. Each attack leaves its target below that floor, and the VQC
had little to lose, so we claim no asymmetry between the model families from it. With the
RBF-SVM surrogate the ratio is 0.07, as on the MSU/ORNL data a surrogate whose attacks barely
reach the VQC.\else.\fi{} \ifarxivbuild Of the four items
checked on the simulated grids, none reproduces in full:
the split item reproduces in part (118-bus) or not at all (14-bus), and the surrogate,
attack-availability, and reading-radius items do not.
\else None of the four items checked on the simulated grids reproduces in full. Their rows have
no near-copies (no test row has a training row closer than 0.1) and their surrogates are well fitted.
\fi
We interpret each item as conditional: it matters where the condition it guards against is
present.
\ifarxivbuild
\subsubsection{Retention Order (Row C)}
Retention order departs from attack-recall order on every source (Spearman 0.07 on the gas
data, 0.10 on the MSU/ORNL data, and $-0.17$ on both grids).
\subsubsection{The Gas Pressure Feature}
The main-study pipeline cannot read it. Injected pressure
values up to $3.4 \times 10^{38}$ set its standard deviation, so standardization maps every
ordinary pressure reading to one value and, in the float32 inputs the models see, turns 68\% of
the training rows into label-conflicting duplicates. Under it, the matched gas models score
0.54--0.59 macro-F1 (Fig.~\ref{fig:gaspressure}; means over seeds). After that registered reading, we applied a fixed signed
logarithm, $\operatorname{sign}(x)\log(1+|x|)$, to pressure before the main-study pipeline,
retuned every model at the same budget, and reran every gas check. The matched
models then score 0.73--0.76, except logistic regression and the VQC (two of its ten seeds
reach 0.68--0.69). The random forest (0.761, seed standard deviation 0.001) leads the ZZ-map QSVM
(0.748, deterministic; min-max scaling) by 0.012 before rounding. The restored signal comes from the injected readings: pressure is recorded on
response packets, and all but one of the attack rows that carry it belong to a
response-injection attack.
\subsubsection{The Gas Readings Under the Main-Study Pipeline}
Every gas reading of Fig.~\ref{fig:replication} uses pressure log-scaled. Under the main-study
pipeline, which we ran first, four rows read differently. The split inflation is 0.042 (in part,
not yes). The accuracy gap is 0.033 (yes, not in part). The reading radius moves the RBF-SVM's
retention by 0.00 (no, not in part). The two kernel machines end at the same retention, 0.113
against 0.112 (no, not in part). The other six rows keep their verdicts, and two checklist
items reproduce in full and two in part. The pair test under the shared protocol also goes
both ways (11 of 12 pairs differ, the quantum model ahead in 7). The tree ensembles' fragility under that pipeline sits
on the duplicated rows, where HopSkipJump needs a
distortion about 150 to 290 times smaller than elsewhere, read over the rows on which it found an
adversarial point. With pressure log-scaled, the white-box run behind rows 3, 5, and B scores
the QSVM on the exact double-precision kernel path. The single-precision path as deployed missed
our agreement check against the exact path there (relative error 0.007 against a tolerance of
0.005), and the two paths'
predictions differ on at most 1 of 500 rows. Every other QSVM number in the gas column uses
the deployed path.
\fi

\ifarxivbuild\begin{figure*}[!tp]\else\begin{figure*}[!t]\fi
\centering
\includegraphics[width=\textwidth]{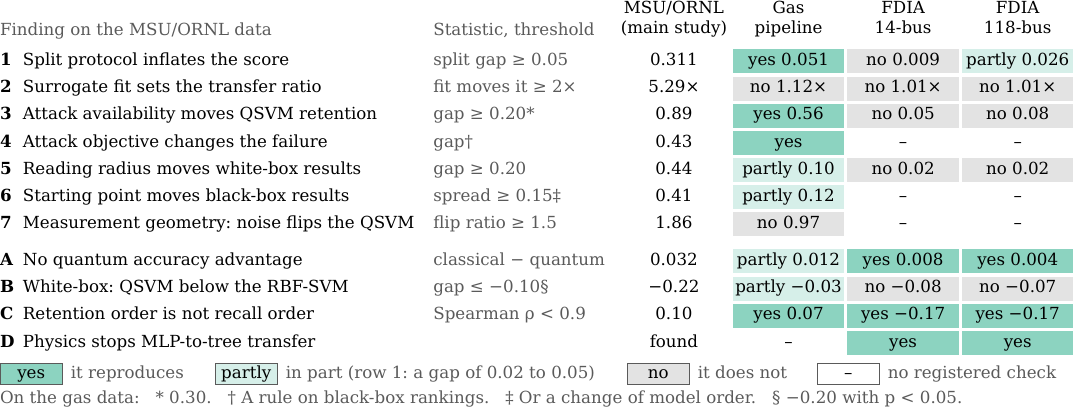}
\caption{Which MSU/ORNL findings reproduce on a gas-pipeline dataset and two simulated grids.
Rows 1 to 7 are the checklist of Fig.~\ref{fig:checklist}; rows A to D are other findings. Each
cell applies a registered threshold and prints its statistic. Row D is judged by a registered rule
with no single number, and its MSU/ORNL \ifarxivbuild cell prints ``found''\else cell is a
four-relay run reported in the complete version~\cite{islam2026benchmarkingarxiv}\fi. Seven cells are judged more strictly or later
than registered\ifarxivbuild: row 6 on the gas data passes its registered rule through a change
of model order only (the QSVM's distortion varies 12\% over restarts) and
is marked down to partly; on the grids the thresholds of row 5 were set before its reading but
after the attack runs it reads (two cells), row 2 is read by the gas rule (the grids' own registered
statistic, the transfer ratio under the class-weighted surrogate, reproduces), and row 1 gives
each grid a word where one joint verdict was registered (intermediate)\else: row 6 on the gas
data is marked down although its rule passed; on the grids row 5's thresholds were set after its
attack runs, row 2 is judged by the gas rule, and row 1 per grid. Statistics: row 2, the factor by
which the class-weighted fit moves the transfer ratio; row 6, the
standard deviation of the QSVM's distortion over restarts, divided by its mean; row 7, the QSVM's flip rate over the RBF-SVM's. Partly on the gas data: 0.05 to 0.20 (row 5), a classical lead
under 0.02 (A)\fi. Gap: the difference between the two readings of that checklist row.
White-box readings are at the worst tested radius; they were registered at $0.5\sigma$, where
no verdict differs.\ifarxivbuild{} Every row A cell uses the min-max scaling (MSU/ORNL: 0.017 under the quantile
scaling, Section~\ref{sec:anglescaling}).\fi{} The gas column uses pressure log-scaled
(Section~\ref{sec:repdata})\ifarxivbuild. Statistics: row-level minus held-out score (row
1), the factor by which the class-weighted fit moves the transfer ratio (2), QSVM
transfer minus white-box retention (3), the RBF-SVM's white-box retention at $0.5\sigma$ minus at
its worst radius inside it (5), the standard deviation of the QSVM's distortion over restarts
divided by its mean (6; the gas rule also passes when the order of the models changes between
restarts), the QSVM over RBF-SVM random-flip ratio (7), best classical
minus best quantum macro-F1 (A), QSVM minus RBF-SVM white-box retention (B; on the gas data yes
also needs row 3 to pass), and the Spearman correlation of retention with attack recall (C; a
reorder is below 0.9). Row D is read at $0.5\sigma$. Transferred retentions stay at $0.5\sigma$; on the gas data the
QSVM's is lower at $0.2\sigma$ (0.482 against 0.686), and with it read there row 3 gives 0.36,
still above its threshold. On the gas data a cell reads partly in row 5 between 0.05 and 0.20;
the text gives the rules of rows A and B\else; its white-box run scores the QSVM in double precision (the
single-precision kernel as deployed missed our precision check there and differs on at most 1
of 500 predictions)\fi.}
\label{fig:replication}
\end{figure*}

\ifarxivbuild
\begin{figure}[!t]
\centering
\includegraphics[width=\columnwidth]{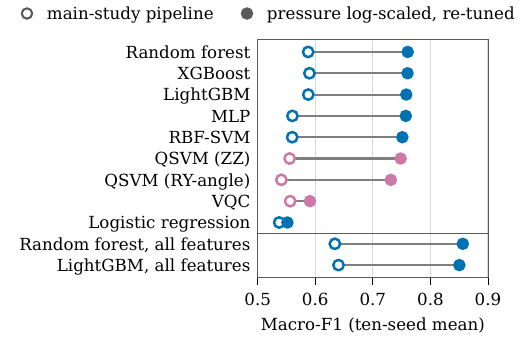}
\caption{Gas pressure feature: macro-F1 under the main-study pipeline (hollow) and with pressure
passed through a fixed signed logarithm and every model retuned (filled); ten-seed means. Above
the rule: matched models (PCA-8, 2000-row cap). Below it: all features.}
\label{fig:gaspressure}
\end{figure}
\fi

\ifarxivbuild\begin{table*}[!tp]\else\begin{table*}[!t]\fi
\centering
\caption{RQ3 and the other checks (binary task; MSU/ORNL data and eight dimensions unless stated)}
\label{tab:otherchecks}
\footnotesize
% auto-generated by experiments/build_further_controls.py -- do not edit by hand
\begin{tabularx}{\textwidth}{@{}>{\raggedright\arraybackslash}p{0.19\textwidth}>{\raggedright\arraybackslash}p{0.33\textwidth}Y@{}}
\toprule
Check & Setting & Reading (macro-F1 unless stated) \\
\midrule
\rowcolor{qgQuantumTint}[0pt][0pt]\multicolumn{3}{@{}l@{}}{\textbf{Settings of the quantum models (RQ3)}} \\
Shots & tuned ZZ QSVM, trained and tested at 1024 shots & loses $0.058$ $\pm$ $0.030$ (three sampling draws) \\\addlinespace[0.5pt]
Depolarizing noise & $p = 0.05$; ZZ QSVM on 400 training rows & loses $0.024$ \\\addlinespace[0.5pt]
Depth & VQC at depths 2, 4, and 6, one untuned setting & macro-F1 $0.45$, $0.47$, $0.44$ (seed spread up to $0.04$); gradient variance $0.00196$, $0.00192$, $0.00195$: no collapse \\\addlinespace[0.5pt]
Training rows & three models, 250 to 2000 rows (two shown) & full-feature XGBoost leads the VQC by $0.125$ at 250 rows and $0.118$ at 2000, and the RY-angle QSVM by $0.116$ and $0.121$ \\\addlinespace[0.5pt]
Angle scaling & quantile angle scaling, both QSVMs retuned & RY-angle $0.527$ to $0.580$; ZZ $0.564$ to $0.577$ (XGBoost $0.597$) \\\addlinespace[0.5pt]
\rowcolor{qgReferenceTint}[0pt][0pt]\multicolumn{3}{@{}l@{}}{\textbf{Other checks}} \\
Label flips & random, up to 20\% of training labels, five seeds & largest mean move $0.03$ (seed spread $0.02$) \\\addlinespace[0.5pt]
Sensor noise & Gaussian, 10~dB signal-to-noise ratio, ten seeds & costs the RBF-SVM $0.17$, random forest $0.13$, LightGBM $0.12$, XGBoost $0.09$, QSVM $0.08$; the other three models lose nothing \\\addlinespace[0.5pt]
Latency (simulated) & per sample, against logistic regression & slower by about 1900$\times$ (VQC), 31\,000$\times$ (RY-angle), 75\,000$\times$ (ZZ) \\\addlinespace[0.5pt]
\rowcolor{qgReferenceTint}[0pt][0pt]\multicolumn{3}{@{}l@{}}{\textbf{Defenses, attacked at $0.5\sigma$}} \\
Adversarial augmentation & one round on PGD examples, three seeds & the VQC gains $+0.012$, under its registered threshold of $0.05$ \\\addlinespace[0.5pt]
Min-max training & every minibatch replaced by PGD, ten seeds & the VQC gains $+0.344$ and passes its threshold of $0.05$, but unattacked the defended VQC and MLP score within 0.05 of the constant-predictor floor (a post-run cut) on 8 and 10 of 10 seeds \\\addlinespace[0.5pt]
Input-noise defense~\cite{ngo2026qupid} & Gaussian input noise; Square search; three seeds & the VQC gains $+0.091$ and passes its threshold of $0.05$; $+0.064$ of the gain is a higher clean score, and the defended VQC stays $0.31$ below the constant-predictor floor \\\addlinespace[0.5pt]
\rowcolor{qgReferenceTint}[0pt][0pt]\multicolumn{3}{@{}l@{}}{\textbf{Physical consistency on the simulated grids (14-bus, then 118-bus)}} \\
Bad-data test & share of rows flagged, every budget & 100\% and 100\% ignoring the physics; 1.2\% and 0.0\% consistent, the unperturbed rates \\\addlinespace[0.5pt]
Square search at $0.1\sigma$ & models with clean macro-F1 $\geq 0.98$; PCA $\sigma$ & $0.28$ to $0.33$; random noise of that size, $0.35$ to $0.91$ \\\addlinespace[0.5pt]
Consistent attack, $0.1\sigma$ & white-box on the QSVM and on the MLP; state $\sigma$ & QSVM $0.26$, $0.25$; MLP $0.30$, $0.30$; random offsets on those two, $0.37$ to $0.59$ \\\addlinespace[0.5pt]
\bottomrule
\end{tabularx}

\end{table*}

\subsection{Physical Consistency}\label{sec:physics}
\ifarxivbuild The simulated grids carry the state estimate behind every measurement, so an
attack can be made to respect the power-flow physics (Fig.~\ref{fig:fdia}).\fi
\subsubsection{The Bad-Data Test}
The simulated grids' own chi-square test\ifarxivbuild\else{} (Section~\ref{sec:repdata})\fi{} flags 100\% of rows perturbed without
respecting the physics, random noise included, at every budget. It flags rows built from a
consistent state offset as rarely as unperturbed rows (Table~\ref{tab:otherchecks}). Such rows are stealthy in the
sense of~\cite{liu2011false}, so the test validates our generator.
\subsubsection{Random Noise Is Enough}
In the $\sigma$ of Section~\ref{sec:threat}, random noise at $0.5\sigma$ alone takes every nonlinear
classical model and the QSVM to the constant-predictor floor (0.333 on these balanced grids) by making them flag normal rows. That budget measures sensitivity to noise, so we read attacks
at $0.1\sigma$\ifarxivbuild. There, the Square search takes every model with
clean macro-F1 of 0.98 or more to 0.28--0.33, the QSVM lowest on both grids, while random noise of
the same size leaves 0.35--0.91\else, where the Square search still does more damage to every
accurate model than noise does (Table~\ref{tab:otherchecks})\fi.
\subsubsection{Physics Stops Transfer; White-Box Fragility Stays}
Attacks that respect the physics stop
transferring from gradient-based surrogates to tree models at $0.5\sigma$ (Fig.~\ref{fig:replication}, row D). At $0.1\sigma$ they
keep a fifth (14-bus) to a third (118-bus) of the unconstrained transfer\ifarxivbuild. Each attack is read in its own $\sigma$ (Section~\ref{sec:repdata})\else{} (each attack in its own $\sigma$)\fi. Crafted on an XGBoost
surrogate, such an attack still transfers to the other tree models on the 118-bus grid. White-box fragility survives the physics for
quantum and classical models alike\ifarxivbuild. The consistent white-box attack at $0.1\sigma$ takes the QSVM to 0.26
and 0.25 on the two grids, the MLP to 0.30 on both, and the VQC to 0.45 and 0.14, against 0.37--0.79
under consistent random offsets of the same size. The RBF-SVM sits at the floor under either\else{}
(QSVM 0.25--0.26 at $0.1\sigma$; Table~\ref{tab:otherchecks})\fi. The QSVM and the MLP are
accurate detectors there (clean macro-F1 of 0.988 or more), so white-box fragility is not an
artifact of the weak MSU/ORNL detectors.
\ifarxivbuild
\subsubsection{One Relay on the MSU/ORNL Data}
A consistency check on the relays' sequence
channels, set to a 1\% false-alarm rate, flags every row of the two kernel detectors' white-box
attacks that ignore the physics and every random row of the same radius\appxsec{app:further}.
An attack on one relay's phasors, with every derived sequence and impedance channel recomputed,
passes that check: at most 0.8\% of its rows are flagged. It still takes both kernel detectors
below chance on all four relays. Its transfer from a gradient-based surrogate to the trees falls
to the level of random noise.
\subsubsection{All Four Relays}
Attacked jointly in the same consistent way, transfer stays within
the surrogate's family. At $0.5\sigma$ a random-forest, XGBoost, or LightGBM surrogate costs the
other trees 0.07--0.24 macro-F1 beyond random noise, and the QSVM, RBF-SVM, logistic regression, and
MLP at most 0.06. At the same budget a QSVM surrogate's attack takes the QSVM itself to 0.006 but
costs no other model more than 0.033 beyond random noise.
\fi

\ifarxivbuild
\begin{figure*}[!t]
\centering
\includegraphics[width=\textwidth]{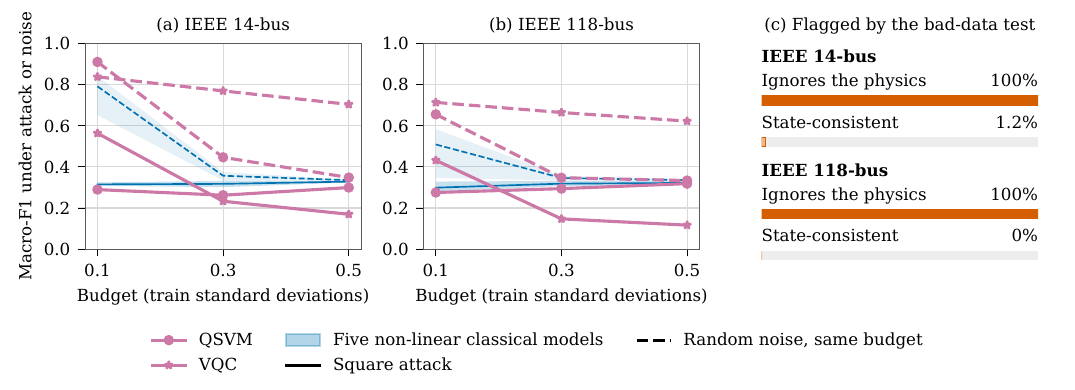}
\caption{Simulated grids. (a), (b) Macro-F1 under the Square black-box attack (solid) and under
random sign noise of the same budget (dashed), for the QSVM, the VQC, and the band of the five
nonlinear classical models; means over three seeds. (c) Share of MLP-crafted rows the chi-square
bad-data test flags, when the attack ignores the physics and when it is built from a consistent
state offset; the share does not change with the budget.}
\label{fig:fdia}
\end{figure*}
\fi

\subsection{What an Attack Does to an Operator's Console}\label{sec:console}
\begin{figure}[!t]
\centering
\includegraphics[width=\columnwidth]{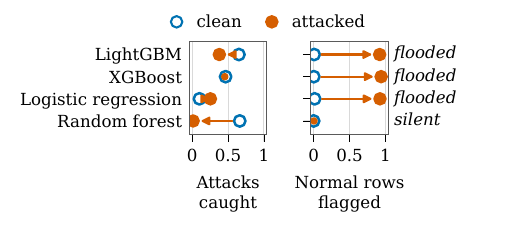}
\caption{What one transferred attack (PGD at $0.5\sigma$) does to an operator's console:
full-feature detectors at a 1\% alarm budget (a false-positive threshold fitted on validation
data), clean (hollow) and attacked (filled; a dot in a ring: unmoved; mean of three seeds).
Silent: almost no attack raises an alarm. Flooded: most
normal rows raise one.}
\label{fig:console}
\end{figure}
On the MSU/ORNL data under the row-level split, at a 1\% alarm budget\ifarxivbuild{} (a false-positive
threshold fitted on validation data)\fi, one transferred attack floods three detectors and silences the random
forest (Fig.~\ref{fig:console}\ifarxivbuild; Appendix~\ref{app:console}\fi)\ifarxivbuild: LightGBM, XGBoost, and
logistic regression flag 92--94\% of normal rows (from under 1.5\% clean), and the forest
catches 1\% of the attacks. The full-feature MLP, the surrogate, has no finite threshold that meets
the 1\% budget and is not drawn. On the simulated grids a flood of the same form appears at
$0.5\sigma$ under the models' default thresholds,
and random noise of that size produces it too: every accurate model flags 99--100\% of normal
rows while still catching 78--100\% of attacks\fi. \ifarxivbuild At a 5\% budget the forest floods too (all four then flag 80--96\% of normal rows), so
the failure an operator sees depends on the alarm budget, and macro-F1 shows neither.\else At a 5\% budget the forest floods too (all four flag 80--96\% of normal rows), so what an
operator sees depends on that budget; macro-F1 shows neither failure.\fi

\subsection{RQ3 and Other Checks}\label{sec:rq3}
\subsubsection{Answer to RQ3}
The encoding, the qubit count, and the angle scaling move the quantum results, and no
setting we ran puts a quantum model ahead of the best classical model (Table~\ref{tab:rq1dims},
Section~\ref{sec:anglescaling}). Under the min-max scaling the ZZ map leads the
RY-angle map by 0.038 macro-F1 at 8 qubits \ifarxivbuild and is about level with it at 16 (0.007 behind, single fits): from 8 to 16 qubits
the RY-angle kernel gains 0.048 and the ZZ kernel 0.003 (differences before rounding). Under the quantile scaling
the two maps are level at 8 qubits. \else (before rounding) and is about level with it at 16;
under the quantile scaling the two are level at 8. \fi\ifarxivbuild Table~\ref{tab:otherchecks}
has the other three: at one untuned training setting the VQC scores about the same at every
depth, below the stratified-random floor (tuned, at depth 6, it scores 0.514); 1024
shots cost the tuned QSVM 0.058; and full-feature XGBoost's lead over the two eight-qubit
quantum models it was compared with is about as large at 250 rows as at 2000.\else Table~\ref{tab:otherchecks}
has the other three: at one untuned setting, where it stays below the stratified-random floor,
the VQC scores about the same at every depth; 1024 shots cost the tuned QSVM 0.058; and
full-feature XGBoost's lead is about as large at 250 rows as at 2000.\fi
\subsubsection{Defenses and Other Checks}
No defense we
measured makes a model robust\ifarxivbuild{} (Table~\ref{tab:otherchecks}); the rest of this section gives the detail\else{} (Table~\ref{tab:otherchecks}, which also gives label flips, sensor noise, and latency)\fi.
\ifarxivbuild
\subsubsection{Shots and Noise}
Shot noise at 1024 and 4096 shots is 2--6\% of the
informative kernel spread across the Gram matrices we measured (4.6\% for the deployed kernel at
1024 shots).
At the ablation's setting (RY-angle QSVM, bandwidth 1.0, $C = 1$) it moves macro-F1 by under
0.01. The deployed ZZ fit ($C = 835$), trained and tested at 1024 shots, loses $0.058 \pm 0.030$
over three sampling draws\appxsec{app:further}. Depolarizing noise at $p = 0.05$ moves an eight-qubit ZZ
QSVM trained on 400 rows by $-0.024$ macro-F1.
\subsubsection{Trainability}
The VQC's gradients do not collapse with depth at eight qubits:
gradient-norm variance stays within 2\% of its mean across depths 2, 4, and 6, so its weak accuracy comes
from the model class, not from a failure to train. Depth does not buy accuracy either: at one
fixed training setting the test macro-F1 is 0.45, 0.47, and 0.44 at the three depths (ten seeds,
spread up to 0.04; Table~\ref{tab:otherchecks}). The plateau is on the qubit
axis\appxsec{app:trainability}.
\subsubsection{Data Efficiency}
In a comparison that is not matched (XGBoost on all
features at fixed settings against eight-qubit quantum models, the VQC at its
Table~\ref{tab:rq1} configuration), the classical lead over
the VQC is 0.125 macro-F1 at 250 samples and 0.118 at 2000. Over the RY-angle QSVM it is 0.116 and
0.121.
\subsubsection{Label Poisoning}
Random label flips (no attacker knowledge) up to 20\% move no
classical
family other than the MLP by more than about 0.023 in mean macro-F1 over five seeds. On average they
move the MLP and the RY-angle QSVM by about 0.03 against a seed spread of 0.02, and the VQC by
0.012 against a seed spread of 0.013.
\subsubsection{Sensor Noise}
Gaussian noise at a 10~dB signal-to-noise ratio costs the tree
ensembles 0.09--0.13 macro-F1, the RBF-SVM 0.17, and the QSVM 0.08, while logistic regression and
the near-chance MLP and VQC lose nothing.
\subsubsection{Latency}
Simulated quantum inference at eight qubits runs about $1900\times$ (VQC) to
$31\,000$--$75\,000\times$ (QSVMs) slower per sample than the fastest classical model.
\subsubsection{Defenses}
One round of adversarial
augmentation with PGD examples at $0.5\sigma$ changes attacked macro-F1 by $-0.253$ for the
unweighted MLP, which inverts further, and by $+0.012$ for the VQC, short of the registered 0.05
threshold. Min-max training that replaces every minibatch by its PGD attack at $0.5\sigma$
passes the registered threshold for the VQC ($+0.344$). Both defenses are read at $0.5\sigma$, the
worst of the three radii their runs attacked at. Yet on 8 of 10 seeds the defended VQC, and on
every seed the defended MLP, is a near-constant majority-class predictor within 0.05 of the
constant-predictor floor. Read against that floor, a rule we apply after the run, no defended
model resists the white-box attack better than predicting ``attack'' on every row.
\subsubsection{Input-Noise Defense}
A QUPID-type noise defense~\cite{ngo2026qupid} (Gaussian input noise at
training and prediction, its size chosen on validation) was selected only for the VQC and logistic
regression. Under the strongest attack we ran on it, the Square search (gradient attacks
under-attack it: 0.161 against 0.110), it passes its registered threshold for the VQC, $+0.091$ over the
undefended model. Most of that gain is a higher clean score ($+0.064$): Square's damage falls only
from 0.44 to 0.42, and the VQC's damage relative to the classical models does not move on that
axis (1.11 against 1.06). The defended VQC still scores 0.110 under attack, 0.31 below the
constant-predictor floor: less broken, not robust.
\fi

\section{Discussion and Limitations}\label{sec:discussion}

\subsection{What the Matched-Dimensionality Comparison Answers}\label{sec:preempt}
The matched contest asks whether the quantum feature map is competitive at matched
dimensionality and matched data. It does not ask whether utilities should replace full-feature
gradient boosting. \ifarxivbuild The matched-contest scores sit near 0.60, and at 16
components the
2000-row cap, which the fidelity kernel's $O(N^2)$ cost forces, accounts for
65\% to 83\% of the gap to the full-feature score, depending on which tuned hyperparameters are
held fixed, and PCA for 17\% to 21\% (Fig.~\ref{fig:bridge})\else The matched-contest scores sit near 0.60. At 16
components, with hyperparameters fixed, the 2000-row cap costs LightGBM 0.197 of the 0.302
macro-F1 below its full-feature score, six to eight times the quantum--classical gap; PCA costs
0.063 (one seed)\fi. The robustness comparisons inherit this scope. \ifarxivbuild A full-feature check shows the low
robustness is not a low-accuracy artifact: at 0.776 macro-F1 on all 128 features, white-box PGD
leaves the MLP at retention 0.119. That is read at $0.1\sigma$, its worst radius inside the
$0.5\sigma$ budget; at the bound it is 0.256.\fi

\ifarxivbuild
\subsection{The Checklist Is the Deliverable}\label{sec:checklist}
Fig.~\ref{fig:checklist} is the checklist: for each of the seven choices, the control in its green
column is what a benchmark should run, and Section~\ref{sec:replication} gives where each item
reproduces. Each choice changed a reported result in this study with the model
families fixed. The regularization constant $C$ is not among them. At the full $2\sigma$ budget it
sets how constant the QSVM's output is under a distance-seeking attack, but inside that ball loss
ascent inverts the QSVM at every $C$ we tried (Section~\ref{sec:widthreg}). Like the kernel length
scale, it changes what the largest radius shows, not what a benchmark should report.

The gas verdict was first registered, under the main-study pipeline, on a list of eight, which
held the kernel length scale
and the regularization. Six or more reproducing would have counted as a reproduction, and five
did, which read partial. The bins were not registered again for the list of seven or for the
log-scaled rerun, so we give the
count (Section~\ref{sec:replication}) without a verdict. In that count the starting-point
item is in part, although its registered rule passed: it passed through a change of model
order between restarts, while the QSVM's own distortion varies
12\% there, under the registered 15\%. The split item
reaches its threshold (0.051 against 0.05). The simulated grids have rows with no near-copies and
well-fitted surrogates, so the surrogate item does not reproduce there and the split item
reproduces at most in part. By its registered statistic, the RBF-SVM's retention at the bound minus
at its worst radius, the reading radius (item 5) reproduces in part on the gas data and not on
the grids. It
still moves the QSVM on the 118-bus grid, by 0.07 in retention and from a disabled to a partial
end state. The reading radius and the kernel length scale it replaced in this list show one
behavior, a kernel machine whose score does not fall steadily with the radius; the list keeps the
one that is a choice the evaluator makes.

The dangerous choices are the ones nobody makes knowingly: an argument an estimator does not
accept, a random draw a library does not expose, a split convention a field
inherits. Two of the cheapest controls caught two of the seven. Checking that
every baseline receives its documented protocol exposed the MLP's missing class weights (item 2).
Reproducing every attack from its seed exposed HopSkipJump's unseeded starting point (item
6).\fi

\ifarxivbuild
\subsection{What Any Fix Can Buy Against the 0.31 Gap}\label{sec:levers}
Which file a detector trains on matters about four times more than how many files it trains on
(single-file scores spread 0.056--0.069), and going from one training file to nine at a fixed row
budget gains only $+0.015 \pm 0.006$. Raw features beyond 16 components add $+0.028$, training
volume $+0.028$, and 16 instead of 8 dimensions $+0.013$. The choice of linear projection adds
nothing, and only a supervised encoder fit on training files clears the stratified-random floor
($+0.063$ against PCA's $+0.029$), measured but not adopted. Summed generously, the measured
levers close 0.08--0.12 of the 0.31 and are not additive, so more of the same telemetry does not
close a cross-file gap.
\fi

\subsection{Physical Realizability, Measured}\label{sec:realizability}
$\ell_\infty$ balls in PCA space do not guarantee grid-consistent measurements.\ifarxivbuild{} On the
MSU/ORNL data the scaler-plus-PCA pipeline is affine, so we map each perturbation back to the 128
raw channels. At $0.5\sigma$, 108 to 109 of them (three seeds) move by more than $0.1\sigma$ on
average over the attacked rows, so the threat model grants
coherent control of most of the measurement vector. An attacker who controls fewer channels
within the same bound in these coordinates chooses from a subset of these perturbations.\else{} Mapped back through the affine scaler and PCA, a $0.5\sigma$ perturbation
moves most of the 128 raw channels by more than $0.1\sigma$, so the threat model grants coherent
control of most of the measurement vector. An attacker who controls fewer channels within the
same bound in these coordinates chooses from a subset of these perturbations.\fi\ifarxivbuild{} On the simulated grids we can do more
(Section~\ref{sec:physics}). There the operator's
chi-square bad-data test already flags every row perturbed without respecting the physics, random
noise included, so the feature-space attacker of the main protocol attacks a
detector that the bad-data test would already protect. Attacks that respect the power-flow
physics pass that test. They stop transferring from gradient-based surrogates to tree models at
$0.5\sigma$ but keep a fifth to a third of the transfer at $0.1\sigma$, and under a white-box
attack they still break the QSVM and the MLP alike.\fi\ifarxivbuild{} The MSU/ORNL data carry no solar or other distributed-energy telemetry, so we
make no claim about such resources.\fi

\subsection{Limitations}\label{sec:limitations}
\ifarxivbuild
All attacks are standard published methods on public research data, and we introduce no new
offensive capability.
\fi
\ifarxivbuild\subsubsection{Split}\fi
Every attack experiment on the MSU/ORNL data except the whole-file
replications trains under the row-level split, so those sections describe detectors that
partly memorize near-duplicate rows. The comparisons still hold, at fixed models within
one split. The white-box kernel contrast and the noise control also replicate under
whole-file training\ifarxivbuild\else, where the QSVM lands 0.47--0.49 below the stratified-random
floor and the RBF-SVM 0.35--0.36 below, each at its worst radius\fi.\ifarxivbuild{} So does the full-feature MLP's white-box
fragility\appxsec{app:further}. The surrogate result keeps row-level scope by necessity,
because the unweighted MLP sits below the stratified-random floor at every quantum-reachable
dimensionality.\fi
\ifarxivbuild\subsubsection{Simulation}\else{} \fi
All quantum results are exact statevector simulations, so they
characterize the model class, not quantum execution\ifarxivbuild. Shot noise moves the
RY-angle QSVM by under 0.01 macro-F1 at the ablation's setting, and the deployed ZZ fit loses
$0.058 \pm 0.030$ at 1024 shots\appxsec{app:further}\fi.
\ifarxivbuild\subsubsection{Data}\fi
The gas readings of Fig.~\ref{fig:replication} use a pressure rescaling \ifarxivbuild that we \fi fixed after the
main-study pipeline's first
registered reading (Section~\ref{sec:repdata}); the thresholds were not
changed, and \ifarxivbuild Section~\ref{sec:replication} gives the readings under both
pipelines\else\cite{islam2026benchmarkingarxiv} gives both readings\fi.\ifarxivbuild{} Neither the laboratory gas data nor our simulated grids are
operational data. We did not run the VQC at 16 qubits, because its tuning cost would
have broken the equal-budget contract.\fi
\ifarxivbuild
\subsubsection{Defenses}
Every model is undefended in its shipped form, and no defense we
measured, a QUPID-type noise defense included, lifts an attacked model above the
constant-predictor floor (Section~\ref{sec:rq3}). The relabel control's labels are a validity instrument,
not a claim about grid labels.
\fi

\section{Conclusion}

\ifarxivbuild
Evaluation choices can decide the outcome. On the standard public power-system attack
dataset, seven such choices each changed a reported result with the model families fixed. The row-level split that the five studies of Table~\ref{tab:related} use inflates our
accuracy headline by 0.31 macro-F1 over the whole-file split. On held-out files at matched
dimensionality tuned classical and quantum models clear the stratified-random floor by little
($+0.032$ and $+0.023$), and
their difference flips sign on 7 of 20 partitions. Unless stated, the MSU/ORNL attack results that follow are for weak eight-dimension
detectors (0.47 to 0.63 clean) trained under the row-level split. Under the shared protocol the three models
ranked most robust receive no white-box attack. The quantum kernel that looks most
robust under MLP-transferred and decision-based attacks falls from 0.56 to 0.01 macro-F1 under its
own white-box attack, and both kernel machines end below the stratified-random floor under
theirs. Tree models admit no gradient-based white-box attack, so their retentions are
upper bounds. All 12 paired
comparisons of attacked macro-F1 under the shared protocol are significant across seeds, in both directions, so we make no
family-level robustness claim. Two results are general in form. Row-level splits of
recording-structured data score memorization of temporal neighbors, over a distance we measured
(2.5 released rows), the effect that blocked cross-validation is designed to
remove~\cite{roberts2017crossvalidation}; synthetic grid data without such neighbors show
almost no gap, and its size depends on the data (0.31 here, 0.051 on the gas data). A transfer asymmetry between families can be made by the surrogate alone: it fell
from $4.6\times$ to $0.9\times$ once the surrogate was fitted correctly. \ifarxivbuild Two more hold on this
dataset. At about 600 queries per point, minimum-distortion black-box attacks overstate quantum-kernel robustness without a
random-perturbation control (the effect does not reproduce on the gas data). At about 10\,900
queries per point the registered median no longer ranks the QSVM the hardest of the three models
swept (by the registered rule, a budget artifact), but a reading row by row against the noise,
made after that result, still finds the attack placing most of the rows the QSVM labels attack
that noise flips (18 to 22 of 28) beyond $0.5\sigma$.
Label alignment, not geometric difference, tracks quantum-kernel accuracy. \fi No quantum model we ran
is more accurate on a gas-pipeline dataset or on two simulated grids, and on the grids the best
model of each group scores 0.988 or more. None is consistently more robust there. On the gas data,
under the shared protocol, the QSVM leads five classical models and the VQC trails all six; only the
MLP and the VQC receive their own attack there, so the other scores are upper bounds. Under its own
white-box attack the QSVM keeps less of its clean score than the RBF-SVM (0.124 against 0.152;
Section~\ref{sec:replication}). On the grids, too, the QSVM keeps less of its clean score than the
RBF-SVM under its own white-box attack, and the operator's own
bad-data test flags every perturbation that ignores the physics, random noise included. The
checklist's items depend on the data. Three of the seven reproduce on the gas data (two of them by
margins inside their spread) and two in part;
none of the four checked on the simulated grids reproduces in full. For grid operators, nothing here favors a quantum detector
today: it is no more accurate, not consistently more robust, and far slower in simulation. Under
the row-level split a transferred attack can flood or silence a full-feature detector's alarms at
a 1\% alarm budget. Future
work includes validation on operational research microgrids (grid-connected facilities with solar
generation, battery storage, and artificial-intelligence-based control) and a federated extension
across distributed grid assets.
\else
Evaluation choices can decide the outcome: on the MSU/ORNL dataset, seven of them each changed a
reported result with model families fixed. The row-level split inflates our
accuracy headline by 0.31 macro-F1 over the whole-file split, and on held-out files at matched
dimensionality tuned classical and quantum models barely clear the stratified-random floor.
Unless stated, the MSU/ORNL attack results that follow are for weak eight-dimension detectors (0.47 to 0.63 clean) trained under the row-level split. A
poorly fitted surrogate alone produced a 4.6-fold transfer ratio. The shared protocol gives its three top-ranked models no white-box
attack: the quantum kernel, ranked first, falls from 0.56 to 0.01 under its
own white-box attack (trees admit no gradient-based one, so theirs are upper bounds). On this
dataset plain random noise refutes the black-box attack that, at about 600 queries per point,
reports it as the hardest target; that effect does not reproduce on the gas data. Under the
shared protocol all 12 quantum--classical pairs differ significantly across seeds, in both
directions, so we make no robustness claim for either group. No quantum
model we ran is more accurate on a gas-pipeline dataset or on two simulated grids. We ask the next
benchmark to run the seven-item checklist of Fig.~\ref{fig:checklist}. Its items
depend on the data: three reproduce on the gas data (two of them by margins inside their spread)
and two in part; none of the four checked
on the simulated grids reproduces in full. For grid
operators, nothing here favors a quantum detector today, and under the row-level split a
transferred attack can flood or silence a full-feature detector's alarms at a 1\% alarm
budget. The MSU/ORNL data carry no
distributed-energy telemetry and none of our data are operational, so future work includes
validation on operational research microgrids (grid-connected facilities with solar generation,
battery storage, and artificial-intelligence-based control) and a federated extension across
distributed grid assets.
\fi

\section*{Reproducibility and Data Availability}

Code, configurations, and seeded pipelines are released as QuantumGridBench~\cite{qgridbench2026code}.
The main dataset is the public \ifarxivbuild Mississippi State University and ORNL\else MSU/ORNL\fi{} power-system attack dataset,
retrieved on 13 August 2026 from the maintainer's \ifarxivbuild industrial control system (ICS) dataset
page\else dataset page\fi~\cite{morris2014icsdata}, which also hosts the gas-pipeline dataset. The simulated grids are
generated by the released scripts\ifarxivbuild. The\else, and the\fi{} accuracy benchmark runs over seeds $\{0,\dots,9\}$ with fixed
split indices\ifarxivbuild. Every reported number is read from a stored run artifact that records its
configuration, git commit, and lock-file hash, or from a merge that lists such runs as its
sources; runs since 16 August 2026 also record thread
counts. A run made from an uncommitted tree
records that fact, and the later ones record the changed paths too. The later runs also write
their per-seed metrics, per-row predictions, and attacked inputs\fi. \ifarxivbuild The extended exhibits are given in the Appendix. XGBoost's fit varies
with the thread count at a fixed seed (about $0.0012$ macro-F1 in the ten-seed mean across 1, 4,
10, and 24 threads, below its seed spread and every gap we claim).\else A longer preprint of this paper, the complete
version~\cite{islam2026benchmarkingarxiv}, adds further controls and exhibits.\fi

\appendices

\section{The Split-Protocol Control, per Partition}\label{app:split}

\begin{table*}[!t]
\centering
\begin{threeparttable}
\caption{Paired-protocol control per partition}
\label{tab:splitparts}
% auto-generated by qgridbench.eval.tables -- do not edit by hand
\begin{tabular}{llrrrrr}
\toprule
Partition & Held-out files & LightGBM (all features) & XGBoost & ZZ QSVM & Contrast & Floor \\
\midrule
Row-level & -- & 0.9053 & 0.5957 & 0.5643 & $+0.0314$ & 0.5005 \\
p0 & 1, 8, 13 & 0.6032 & 0.5347 & 0.5070 & $+0.0277$ & 0.4980 \\
p1 & 3, 7, 9 & 0.5837 & 0.5134 & 0.5253 & $-0.0119$ & 0.4979 \\
p2 & 1, 10, 11 & 0.6033 & 0.5393 & 0.5093 & $+0.0300$ & 0.5003 \\
p3 & 5, 9, 14 & 0.5937 & 0.5028 & 0.4872 & $+0.0156$ & 0.4995 \\
p4 & 1, 11, 12 & 0.5884 & 0.5259 & 0.5105 & $+0.0154$ & 0.4991 \\
\bottomrule
\end{tabular}

\begin{tablenotes}[flushleft]
\footnotesize
\setlength{\labelsep}{0pt}
\item[] It uses one training seed, and ten-seed
retraining moves the LightGBM column by about 0.003 per partition. The transforms are
identical, and only the split indices change. The contrast column is classical $-$
quantum at matched dimensionality under the 2000-row cap. Partition p1 is the sign flip
quoted in Section~\ref{sec:split}.
\end{tablenotes}
\end{threeparttable}
\end{table*}

Table~\ref{tab:splitparts} gives the per-partition detail behind
Table~\ref{tab:split}. It shows the two sentences the body rests on, rather than
asserting them. The level drop is uniform across partitions. The matched contrast flips
sign on partition p1, where the tuned QSVM scores above the tuned classical
model. Held-out file ids are listed so any partition can be reproduced.

\subsection{The White-Box Contrast Under Whole-File Training}
We refit both tuned kernel
models on partitions p0 and p2 and ran the identical white-box attack at the six grid radii. We compared the attack's gradient
with the reference implementation on each partition (gradient error $\leq 5\times10^{-7}$), with ten attack seeds each.
Read at its worst radius inside $0.5\sigma$, the QSVM falls to $0.031 \pm 0.005$ and
$0.013 \pm 0.003$ macro-F1 (at $0.2\sigma$ and $0.3\sigma$), which is
$0.47$ and $0.49$ below each partition's stratified-random floor. The RBF-SVM falls to
$0.140 \pm 0.005$ and $0.151 \pm 0.004$, both at $0.1\sigma$ ($0.36$ and $0.35$ below). The QSVM is lower in
all 20 attack-seed pairs, by $0.11$ and $0.14$ macro-F1 in the mean. These readings bound the
attacker from above. One who picks the radius per input, over the six stored radii, leaves the
QSVM at $0.011$ and $0.004$ and the RBF-SVM at $0.100$ and $0.095$. Read at $0.5\sigma$ itself
the gap looks about three times larger ($0.31$ and $0.38$), because the RBF-SVM climbs back to
$0.40$ there. Two fit-dependence details come with this. First, at $0.5\sigma$ the RBF-SVM
trained on whole files ends just below the constant-predictor floor ($0.397$ and $0.406$) rather than on it. Second, on p0 its
\emph{clean} macro-F1 (0.473) is itself below the stratified-random floor. So hyperparameters tuned under
the row-level split do not guarantee an above-chance classical model on held-out
files.

\section{Full-Feature Reference, Binary Task}\label{app:fullbinary}

Table~\ref{tab:binary_full} reports the classical models on the full 128 features for
the headline binary task. It is the reference behind the 0.906 ceiling
quoted in Section~\ref{sec:results}. That figure uses the row-level split, and
Section~\ref{sec:split} measures how much it depends on the split. The table is also the
source of the cap/PCA decomposition.

\begin{table*}[!t]
\centering
\caption{Binary task, full 128 features (test split, ten seeds)}
\label{tab:binary_full}
% auto-generated by qgridbench.eval.tables -- do not edit by hand
\begin{tabular}{lrrrrrr}
\toprule
Model & Macro-F1 & Balanced accuracy & AUPRC & ROC-AUC & Brier & ECE \\
\midrule
Logistic regression$^{b}$ & 0.598\,$\pm$\,0.000 & 0.630\,$\pm$\,0.000 & 0.849\,$\pm$\,0.000 & 0.692\,$\pm$\,0.000 & 0.454\,$\pm$\,0.000 & 0.045\,$\pm$\,0.000 \\
Random forest$^{b}$ & 0.884\,$\pm$\,0.001 & 0.862\,$\pm$\,0.001 & 0.986\,$\pm$\,0.000 & 0.966\,$\pm$\,0.000 & 0.147\,$\pm$\,0.000 & 0.075\,$\pm$\,0.001 \\
XGBoost$^{b}$ & 0.891\,$\pm$\,0.002 & 0.886\,$\pm$\,0.002 & 0.979\,$\pm$\,0.001 & 0.957\,$\pm$\,0.001 & 0.134\,$\pm$\,0.001 & 0.009\,$\pm$\,0.001 \\
LightGBM$^{a}$ & 0.906\,$\pm$\,0.001 & 0.896\,$\pm$\,0.002 & 0.986\,$\pm$\,0.001 & 0.970\,$\pm$\,0.001 & 0.126\,$\pm$\,0.001 & 0.051\,$\pm$\,0.001 \\
RBF-SVM$^{b}$ & 0.705\,$\pm$\,0.000 & 0.686\,$\pm$\,0.000 & 0.881\,$\pm$\,0.000 & 0.791\,$\pm$\,0.000 & 0.336\,$\pm$\,0.000 & 0.109\,$\pm$\,0.000 \\
MLP$^{b}$ & 0.766\,$\pm$\,0.015 & 0.760\,$\pm$\,0.017 & 0.921\,$\pm$\,0.003 & 0.849\,$\pm$\,0.008 & 0.285\,$\pm$\,0.009 & 0.077\,$\pm$\,0.007 \\
\bottomrule
\end{tabular}
% a = best model on mean macro-F1;
% b = significantly below the best (Holm-corrected p < 0.05).

\end{table*}

\section{The Matched Contest at 12 and 16 Dimensions}\label{app:rq1dims}
Tables~\ref{tab:rq1d12} and~\ref{tab:rq1d16} give the six metrics of Table~\ref{tab:rq1} at 12
and 16 dimensions, with the same marks. The MLP rows are the unweighted fit, because the
class-weighted fit was run at eight dimensions only. The VQC was not run at 16.

\begin{table*}[!t]
\centering
\caption{RQ1 at matched dimensionality (PCA-12, binary task, test split; ten seeds for the classical models)}
\label{tab:rq1d12}
% auto-generated by qgridbench.eval.tables -- do not edit by hand
\begin{tabular}{lrrrrrr}
\toprule
Model & Macro-F1 & Balanced accuracy & AUPRC & ROC-AUC & Brier & ECE \\
\midrule
Logistic regression$^{b}$ & 0.511\,$\pm$\,0.000 & 0.538\,$\pm$\,0.000 & 0.749\,$\pm$\,0.000 & 0.542\,$\pm$\,0.000 & 0.496\,$\pm$\,0.000 & 0.023\,$\pm$\,0.000 \\
RBF-SVM$^{b}$ & 0.575\,$\pm$\,0.000 & 0.581\,$\pm$\,0.000 & 0.761\,$\pm$\,0.000 & 0.601\,$\pm$\,0.000 & 0.487\,$\pm$\,0.000 & 0.096\,$\pm$\,0.000 \\
Random forest$^{b}$ & 0.599\,$\pm$\,0.002 & 0.594\,$\pm$\,0.002 & 0.802\,$\pm$\,0.001 & 0.651\,$\pm$\,0.001 & 0.400\,$\pm$\,0.000 & 0.034\,$\pm$\,0.002 \\
XGBoost$^{a}$ & 0.606\,$\pm$\,0.002 & 0.602\,$\pm$\,0.002 & 0.794\,$\pm$\,0.002 & 0.648\,$\pm$\,0.002 & 0.438\,$\pm$\,0.002 & 0.113\,$\pm$\,0.002 \\
LightGBM$^{b}$ & 0.604\,$\pm$\,0.001 & 0.600\,$\pm$\,0.001 & 0.793\,$\pm$\,0.001 & 0.649\,$\pm$\,0.001 & 0.488\,$\pm$\,0.001 & 0.198\,$\pm$\,0.002 \\
MLP$^{b}$ & 0.473\,$\pm$\,0.044 & 0.519\,$\pm$\,0.017 & 0.758\,$\pm$\,0.011 & 0.566\,$\pm$\,0.016 & 0.420\,$\pm$\,0.012 & 0.051\,$\pm$\,0.023 \\
\rowcolor{qgQuantumTint}QSVM (RY-angle) & 0.541\,$\pm$\,0.000 & 0.546\,$\pm$\,0.000 & 0.747\,$\pm$\,0.000 & 0.562\,$\pm$\,0.000 & 0.492\,$\pm$\,0.000 & 0.054\,$\pm$\,0.000 \\
\rowcolor{qgQuantumTint}QSVM (ZZ) & 0.564\,$\pm$\,0.000 & 0.562\,$\pm$\,0.000 & 0.755\,$\pm$\,0.000 & 0.586\,$\pm$\,0.000 & 0.473\,$\pm$\,0.000 & 0.086\,$\pm$\,0.000 \\
\rowcolor{qgQuantumTint}VQC$^{b}$ & 0.510\,$\pm$\,0.009 & 0.514\,$\pm$\,0.007 & 0.727\,$\pm$\,0.007 & 0.522\,$\pm$\,0.010 & 0.489\,$\pm$\,0.007 & 0.062\,$\pm$\,0.028 \\
\rowcolor{qgReferenceTint}Majority class & 0.415\,$\pm$\,0.000 & 0.500\,$\pm$\,0.000 & 0.710\,$\pm$\,0.000 & 0.500\,$\pm$\,0.000 & 0.579\,$\pm$\,0.000 & 0.290\,$\pm$\,0.000 \\
\rowcolor{qgReferenceTint}Stratified random & 0.501\,$\pm$\,0.005 & 0.501\,$\pm$\,0.005 & 0.710\,$\pm$\,0.000 & 0.501\,$\pm$\,0.000 & 0.412\,$\pm$\,0.000 & 0.000\,$\pm$\,0.000 \\
\bottomrule
\end{tabular}
% a = best model on mean macro-F1;
% b = significantly below the best (Holm-corrected p < 0.05).

\end{table*}

\begin{table*}[!t]
\centering
\caption{RQ1 at matched dimensionality (PCA-16, binary task, test split, ten seeds)}
\label{tab:rq1d16}
% auto-generated by qgridbench.eval.tables -- do not edit by hand
\begin{tabular}{lrrrrrr}
\toprule
Model & Macro-F1 & Balanced accuracy & AUPRC & ROC-AUC & Brier & ECE \\
\midrule
Logistic regression$^{b}$ & 0.513\,$\pm$\,0.000 & 0.539\,$\pm$\,0.000 & 0.747\,$\pm$\,0.000 & 0.544\,$\pm$\,0.000 & 0.497\,$\pm$\,0.000 & 0.034\,$\pm$\,0.000 \\
RBF-SVM$^{b}$ & 0.581\,$\pm$\,0.000 & 0.586\,$\pm$\,0.000 & 0.756\,$\pm$\,0.000 & 0.597\,$\pm$\,0.000 & 0.504\,$\pm$\,0.000 & 0.125\,$\pm$\,0.000 \\
Random forest$^{b}$ & 0.598\,$\pm$\,0.002 & 0.593\,$\pm$\,0.001 & 0.801\,$\pm$\,0.001 & 0.648\,$\pm$\,0.001 & 0.405\,$\pm$\,0.000 & 0.030\,$\pm$\,0.001 \\
XGBoost$^{a}$ & 0.607\,$\pm$\,0.001 & 0.603\,$\pm$\,0.001 & 0.801\,$\pm$\,0.001 & 0.654\,$\pm$\,0.001 & 0.432\,$\pm$\,0.001 & 0.110\,$\pm$\,0.001 \\
LightGBM & 0.606\,$\pm$\,0.001 & 0.602\,$\pm$\,0.001 & 0.801\,$\pm$\,0.001 & 0.654\,$\pm$\,0.000 & 0.452\,$\pm$\,0.001 & 0.146\,$\pm$\,0.001 \\
MLP$^{b}$ & 0.459\,$\pm$\,0.033 & 0.513\,$\pm$\,0.011 & 0.754\,$\pm$\,0.015 & 0.561\,$\pm$\,0.023 & 0.424\,$\pm$\,0.023 & 0.060\,$\pm$\,0.045 \\
\rowcolor{qgQuantumTint}QSVM (RY-angle) & 0.574\,$\pm$\,0.000 & 0.580\,$\pm$\,0.000 & 0.756\,$\pm$\,0.000 & 0.589\,$\pm$\,0.000 & 0.499\,$\pm$\,0.000 & 0.095\,$\pm$\,0.000 \\
\rowcolor{qgQuantumTint}QSVM (ZZ) & 0.567\,$\pm$\,0.000 & 0.565\,$\pm$\,0.000 & 0.751\,$\pm$\,0.000 & 0.585\,$\pm$\,0.000 & 0.463\,$\pm$\,0.000 & 0.074\,$\pm$\,0.000 \\
\rowcolor{qgReferenceTint}Majority class & 0.415\,$\pm$\,0.000 & 0.500\,$\pm$\,0.000 & 0.710\,$\pm$\,0.000 & 0.500\,$\pm$\,0.000 & 0.579\,$\pm$\,0.000 & 0.290\,$\pm$\,0.000 \\
\rowcolor{qgReferenceTint}Stratified random & 0.501\,$\pm$\,0.005 & 0.501\,$\pm$\,0.005 & 0.710\,$\pm$\,0.000 & 0.501\,$\pm$\,0.000 & 0.412\,$\pm$\,0.000 & 0.000\,$\pm$\,0.000 \\
\bottomrule
\end{tabular}
% a = best model on mean macro-F1;
% b = significantly below the best (Holm-corrected p < 0.05).

\end{table*}

\section{Positive Control: Per-Cell Evidence}\label{app:relabel}

Table~\ref{tab:relabel} gives the per-cell evidence behind the positive control in
Section~\ref{sec:results}. The labels are engineered from each Gram's own geometric
eigenvector~\cite{huang2021power,egginger2024hyperparameter}. On these labels, the identical
pipeline detects a quantum win in every cell. The engineered labels raise KTA by one to two
orders of magnitude over the grid labels on the same Grams. The win size tracks $\log g$.
``Quantum win'' is QSVM macro-F1 minus the best freshly retuned classical opponent
(RBF-SVM or XGBoost), mean~$\pm$~std over five train/test resamples.

\begin{table*}[!t]
\centering
\caption{Relabel control per cell: $g$, KTA on grid vs.\ engineered labels, and the
quantum win (five resamples)}
\label{tab:relabel}
% auto-generated by qgridbench.eval.tables -- do not edit by hand
\begin{tabular}{lrrrr}
\toprule
Cell & $g$ & KTA (grid $y$) & KTA (engineered $y$) & Quantum win \\
\midrule
RY-angle, 8 qubits, binary & 13.43 & 0.0023 & 0.183 & 0.027\,$\pm$\,0.007 \\
ZZ, 8 qubits, binary & 6.34 & 0.0035 & 0.230 & 0.016\,$\pm$\,0.007 \\
RY-angle, 12 qubits, binary & 17.21 & 0.0022 & 0.500 & 0.004\,$\pm$\,0.003 \\
ZZ, 12 qubits, binary & 73.04 & 0.0145 & 0.055 & 0.053\,$\pm$\,0.006 \\
RY-angle, 16 qubits, binary & 5.60 & 0.0031 & 0.233 & 0.026\,$\pm$\,0.009 \\
ZZ, 16 qubits, binary & 70.12 & 0.0143 & 0.046 & 0.092\,$\pm$\,0.010 \\
RY-angle, 8 qubits, three-class & 5.39 & 0.0033 & 0.480 & 0.008\,$\pm$\,0.004 \\
ZZ, 8 qubits, three-class & 101.71 & 0.0113 & 0.120 & 0.082\,$\pm$\,0.018 \\
\midrule
\multicolumn{5}{l}{Pearson $r(\log g,\ \mathrm{win}) = +0.85$, $p = 0.007$ (8 cells)} \\
\bottomrule
\end{tabular}

\end{table*}

\section{Three-Class Formulation}\label{app:threeclass}

The three-class task (no-event / natural / attack, 5.6\%/23.4\%/71.0\%) repeats the
structure of the binary findings. Under the min-max scaling, the best tuned classical model leads at
matched dimensionality (XGBoost 0.516 against the ZZ-map QSVM's 0.486 at eight dimensions). The full-feature ceiling is high (LightGBM 0.917 macro-F1,
row-level split). The whole-file control of Section~\ref{sec:split} was run on the binary
task only. The kernel diagnostics repeat: the tuned ZZ kernel reaches $g = 101.7$ with low KTA,
and the ``KTA predicts, $g$ flips'' pattern holds in every sweep cell
(Section~\ref{sec:diag}). Tables~\ref{tab:triple_full} and \ref{tab:triple_pca8} report
the classical models on both regimes.

\begin{table*}[t]
\centering
\caption{Three-class task, full 128 features (test split, ten seeds)}
\label{tab:triple_full}
% auto-generated by qgridbench.eval.tables -- do not edit by hand
\begin{tabular}{lrrrrrr}
\toprule
Model & Macro-F1 & Balanced accuracy & AUPRC & ROC-AUC & Brier & ECE \\
\midrule
Logistic regression$^{b}$ & 0.496\,$\pm$\,0.000 & 0.667\,$\pm$\,0.000 & 0.563\,$\pm$\,0.000 & 0.772\,$\pm$\,0.000 & 0.532\,$\pm$\,0.000 & 0.035\,$\pm$\,0.000 \\
Random forest$^{b}$ & 0.887\,$\pm$\,0.001 & 0.865\,$\pm$\,0.001 & 0.952\,$\pm$\,0.000 & 0.973\,$\pm$\,0.000 & 0.162\,$\pm$\,0.000 & 0.091\,$\pm$\,0.001 \\
XGBoost$^{b}$ & 0.910\,$\pm$\,0.002 & 0.904\,$\pm$\,0.002 & 0.959\,$\pm$\,0.001 & 0.975\,$\pm$\,0.000 & 0.124\,$\pm$\,0.001 & 0.014\,$\pm$\,0.001 \\
LightGBM$^{a}$ & 0.917\,$\pm$\,0.001 & 0.908\,$\pm$\,0.001 & 0.967\,$\pm$\,0.000 & 0.980\,$\pm$\,0.000 & 0.114\,$\pm$\,0.001 & 0.024\,$\pm$\,0.001 \\
RBF-SVM$^{b}$ & 0.678\,$\pm$\,0.000 & 0.637\,$\pm$\,0.000 & 0.685\,$\pm$\,0.000 & 0.808\,$\pm$\,0.000 & 0.344\,$\pm$\,0.000 & 0.068\,$\pm$\,0.000 \\
MLP$^{b}$ & 0.763\,$\pm$\,0.030 & 0.755\,$\pm$\,0.033 & 0.811\,$\pm$\,0.019 & 0.894\,$\pm$\,0.010 & 0.288\,$\pm$\,0.012 & 0.073\,$\pm$\,0.011 \\
\bottomrule
\end{tabular}
% a = best model on mean macro-F1;
% b = significantly below the best (Holm-corrected p < 0.05).

\end{table*}

\begin{table*}[t]
\centering
\caption{Three-class task, PCA-8 with the 2000-row cap (test split, ten seeds)}
\label{tab:triple_pca8}
% auto-generated by qgridbench.eval.tables -- do not edit by hand
\begin{tabular}{lrrrrrr}
\toprule
Model & Macro-F1 & Balanced accuracy & AUPRC & ROC-AUC & Brier & ECE \\
\midrule
Logistic regression$^{b}$ & 0.343\,$\pm$\,0.000 & 0.513\,$\pm$\,0.000 & 0.393\,$\pm$\,0.000 & 0.635\,$\pm$\,0.000 & 0.648\,$\pm$\,0.000 & 0.073\,$\pm$\,0.000 \\
RBF-SVM$^{b}$ & 0.490\,$\pm$\,0.000 & 0.554\,$\pm$\,0.000 & 0.450\,$\pm$\,0.000 & 0.677\,$\pm$\,0.000 & 0.549\,$\pm$\,0.000 & 0.113\,$\pm$\,0.000 \\
Random forest$^{b}$ & 0.498\,$\pm$\,0.003 & 0.485\,$\pm$\,0.003 & 0.497\,$\pm$\,0.001 & 0.700\,$\pm$\,0.001 & 0.455\,$\pm$\,0.001 & 0.075\,$\pm$\,0.002 \\
XGBoost$^{a}$ & 0.516\,$\pm$\,0.002 & 0.517\,$\pm$\,0.002 & 0.501\,$\pm$\,0.001 & 0.708\,$\pm$\,0.001 & 0.471\,$\pm$\,0.001 & 0.059\,$\pm$\,0.001 \\
LightGBM$^{b}$ & 0.510\,$\pm$\,0.002 & 0.504\,$\pm$\,0.002 & 0.509\,$\pm$\,0.001 & 0.707\,$\pm$\,0.000 & 0.467\,$\pm$\,0.001 & 0.081\,$\pm$\,0.001 \\
MLP$^{b}$ & 0.305\,$\pm$\,0.010 & 0.344\,$\pm$\,0.004 & 0.373\,$\pm$\,0.007 & 0.575\,$\pm$\,0.013 & 0.440\,$\pm$\,0.003 & 0.046\,$\pm$\,0.012 \\
\bottomrule
\end{tabular}
% a = best model on mean macro-F1;
% b = significantly below the best (Holm-corrected p < 0.05).

\end{table*}

\FloatBarrier
\subsection{Where the Three-Class Errors Fall}
Macro-F1 cannot show which
confusion a detector makes. Two off-diagonal cells matter operationally, and they point
in opposite directions. Calling a \emph{natural} disturbance an attack is a false alarm on
an event that already demands operator attention. Calling an attack natural is a
miss. Table~\ref{tab:confusion} reports both, pooling counts over all ten seeds before
forming rates. Eleven of the twelve model-regime cells err toward the false alarm, several
by an order of magnitude. On all 128 features the strongest model (LightGBM, 0.917 macro-F1)
calls 19.7\% of natural disturbances attacks while missing 3.5\% of real attacks. Logistic
regression on full features is the single exception (0.253 against 0.376), and it is also
the weakest full-feature model. The natural class is the hardest of the three for every
model. LightGBM reaches per-class F1 0.963 and 0.948 on no-event and attack, against
0.840 on natural. This is the expected shape, because an injection attack is designed
to resemble a legitimate disturbance.

\begin{table*}[!t]
\centering
\begin{threeparttable}
\caption{Three-class confusion, pooled over ten seeds}
\label{tab:confusion}
% auto-generated by qgridbench.eval.tables -- do not edit by hand
\begin{tabular}{@{}lrrrrr@{}}
\toprule
\multicolumn{6}{@{}l}{\textit{\textbf{(a) All 128 features}}} \\
Model & Natural$\to$attack & Attack$\to$natural & F1 no-event & F1 natural & F1 attack \\
\midrule
Logistic regression & 0.253 & 0.376 & 0.432 & 0.433 & 0.622 \\
Random forest & 0.302 & 0.027 & 0.944 & 0.782 & 0.935 \\
XGBoost & 0.199 & 0.045 & 0.960 & 0.825 & 0.943 \\
LightGBM & 0.197 & 0.035 & 0.963 & 0.840 & 0.948 \\
RBF-SVM & 0.367 & 0.034 & 0.900 & 0.724 & 0.920 \\
MLP & 0.436 & 0.097 & 0.820 & 0.596 & 0.871 \\
\midrule
\multicolumn{6}{@{}l}{\textit{\textbf{(b) PCA-8, the matched-dimensionality regime}}} \\
Model & Natural$\to$attack & Attack$\to$natural & F1 no-event & F1 natural & F1 attack \\
\midrule
Logistic regression & 0.364 & 0.270 & 0.224 & 0.290 & 0.513 \\
RBF-SVM & 0.562 & 0.250 & 0.427 & 0.314 & 0.711 \\
Random forest & 0.690 & 0.145 & 0.387 & 0.318 & 0.790 \\
XGBoost & 0.610 & 0.199 & 0.429 & 0.352 & 0.767 \\
LightGBM & 0.633 & 0.179 & 0.406 & 0.347 & 0.777 \\
MLP$^{a}$ & 0.950 & 0.019 & 0.000 & 0.089 & 0.827 \\
\bottomrule
\end{tabular}
% a = model collapsed onto one class (an F1 below 0.10), not read as cautious.

\begin{tablenotes}[flushleft]
\footnotesize
\setlength{\labelsep}{0pt}
\item[] Natural$\to$attack is the fraction
of natural disturbances called attacks (false alarm); attack$\to$natural the fraction of
attacks called natural (miss). $^{a}$Marks a model that has collapsed onto one class
(per-class F1 below $0.10$). Its miss rate is degeneracy rather than robustness.
\end{tablenotes}
\end{threeparttable}
\end{table*}

The matched-dimensionality regime sharpens this pattern. Under PCA-8
the same model's false-alarm rate rises from 0.197 to 0.633, while its miss rate rises only
from 0.035 to 0.179. The two minority classes lose most of their per-class F1 (no-event
$0.963 \to 0.406$, natural $0.840 \to 0.347$), while the majority attack class keeps far
more ($0.948 \to 0.777$). So compression does not degrade the task evenly. It pushes
every family toward predicting the majority class, which is the same direction as the
false-alarm asymmetry. The PCA-8 MLP is marked because it has collapsed outright. It
predicts attack for 95\% of natural disturbances and scores per-class F1 $0.000$ on
no-event. Its low miss rate is therefore degeneracy, not caution. We report it rather than
drop it.

\section{The 37-Scenario Task}\label{app:multiclass}

Fig.~\ref{fig:multiclass} repeats the comparison on the release's 37 scenario labels. The
hyperparameters are the three-class winners; no 37-class search was run. On all 128 features
the nonlinear classical models score 0.59--0.82 macro-F1 (LightGBM 0.815). At PCA-8 under the
2000-row cap every model is at or below 0.171, the QSVM at 0.147, against 0.027 for a uniform
guess; the MLP falls below that guess (0.024). So the cost of the matched regime repeats at 37 classes. The VQC is left out because its
readout uses one qubit per class, and 37 classes exceed eight qubits. The full-feature random
forest is not reported: it exceeded the memory budget of our machine.

\begin{figure*}[!t]
\centering
\includegraphics[width=\textwidth]{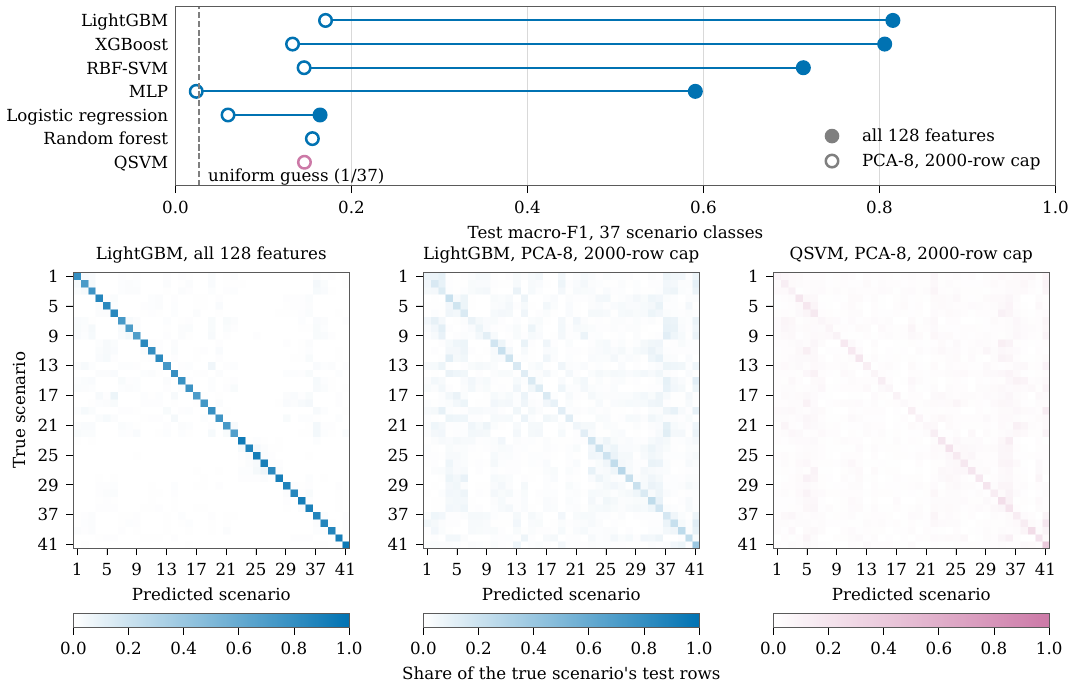}
\caption{Scores on the 37-scenario task (ten seeds, three-class hyperparameters reused). Top: test
macro-F1 per model on all 128 features (filled) and at PCA-8 under the 2000-row cap (hollow).
Bottom: row-normalized confusion matrices for LightGBM in both regimes and for the QSVM; the
axes carry the release's scenario numbers.}
\label{fig:multiclass}
\end{figure*}

\section{Data-Efficiency Curves}\label{app:dataeff}

Fig.~\ref{fig:dataeff} plots test macro-F1 against training-set size. The arms are
XGBoost on all 128 features at fixed settings, and the untuned RY-angle QSVM and the VQC at
eight qubits (Section~\ref{sec:rq3}). The classical lead over the RY-angle QSVM is 0.116
macro-F1 at 250 training samples and 0.121 at 2000; over the VQC at Table~\ref{tab:rq1}'s
configuration it is 0.125 and 0.118. The arms are not matched in features, so the direction
holds and the size does not carry over to the matched regime.

\begin{figure}[!t]
\centering
\includegraphics[width=\columnwidth]{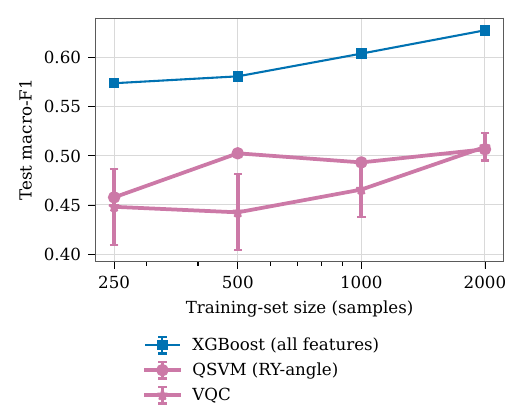}
\caption{Data efficiency: XGBoost on all 128 features at fixed, untuned settings leads the
untuned eight-qubit quantum models at every training-set size (ten seeds, mean $\pm$ std). The
arms are not matched in features.}
\label{fig:dataeff}
\end{figure}

\section{Bandwidth Sweep Diagnostics}\label{app:bwsweep}

Fig.~\ref{fig:bandwidth} shows the 36-point binary bandwidth sweep behind
Section~\ref{sec:diag}. The three-class variant adds 18 configurations with the same
pattern. The spectral-shape panels locate the two degenerate regimes: near-constant
at small bandwidth, near-identity at large. The performance panels show validation macro-F1
tracking KTA, while the geometric difference $g$ peaks in the degenerate regions. The dotted
vertical line marks each cell's Optuna-tuned bandwidth. A bandwidth of 3.0 sits beyond the
tuning box and loses everywhere. Table~\ref{tab:diag} lists the diagnostics for every
encoding and qubit count.

\begin{figure*}[!t]
\centering
\includegraphics[width=\textwidth]{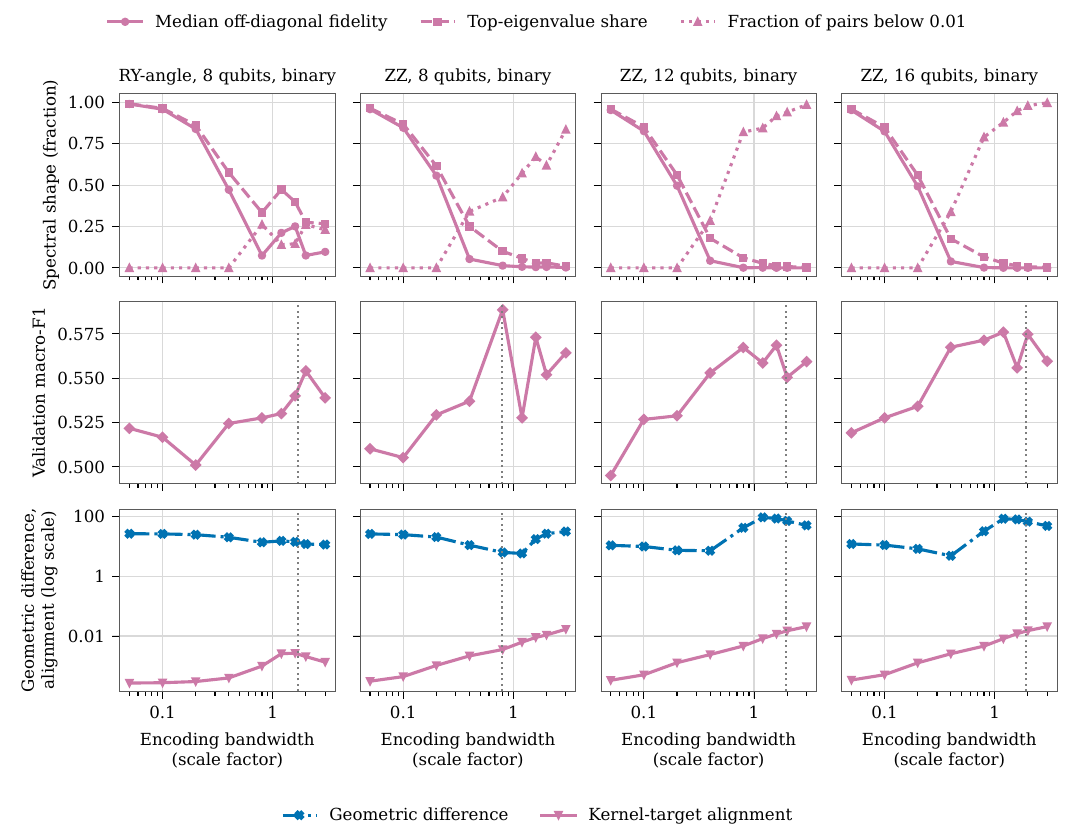}
\caption{Bandwidth sweep per cell (columns). Top: Gram spectral shape. Middle: validation
macro-F1. Bottom: $g$ and KTA on a log scale. The dotted line marks each cell's tuned
bandwidth.}
\label{fig:bandwidth}
\end{figure*}

\begin{table}[!t]
\centering
\setlength{\tabcolsep}{3.5pt}
\begin{threeparttable}
\caption{Kernel diagnostics across encodings and qubit counts}
\label{tab:diag}
% auto-generated by qgridbench.eval.tables -- do not edit by hand
\begin{tabular}{llrrrr}
\toprule
Encoding & Qubits & Test macro-F1 & KTA$_q$ & KTA$_{\mathrm{RBF}}$ & $g(K_C\!\to\!K_Q)$ \\
\midrule
RY-angle & 8 & 0.527\,$\pm$\,0.000 & 0.002 & 0.013 & 13.43 \\
ZZ & 8 & 0.564\,$\pm$\,0.000 & 0.003 & 0.013 & 6.34 \\
RY-angle & 12 & 0.541\,$\pm$\,0.000 & 0.002 & 0.019 & 17.21 \\
ZZ & 12 & 0.564\,$\pm$\,0.000 & 0.015 & 0.005 & 73.04 \\
RY-angle & 16 & 0.574\,$\pm$\,0.000 & 0.003 & 0.006 & 5.60 \\
ZZ & 16 & 0.567\,$\pm$\,0.000 & 0.014 & 0.006 & 70.12 \\
\bottomrule
\end{tabular}

\begin{tablenotes}[flushleft]
\footnotesize
\setlength{\labelsep}{0pt}
\item[] A large geometric
difference $g$ never coincides with label alignment (KTA), and accuracy follows KTA.
Section~\ref{sec:diag} quotes its numbers.
\end{tablenotes}
\end{threeparttable}
\end{table}

\section{Bandwidth Fragility: Accuracy and Fragility Co-Peak}\label{app:fragility}

Fig.~\ref{fig:fragility} is the exhibit behind the fragility co-peak
(Section~\ref{sec:blackbox}). We rebuilt the ZZ-map QSVM at eight qubits across encoding
bandwidths. The random-perturbation flip rate at $0.5\sigma$ peaks at the accuracy-optimal
bandwidth. It falls only where the clean score sits at the stratified-random floor.

\begin{figure}[!t]
\centering
\includegraphics[width=\columnwidth]{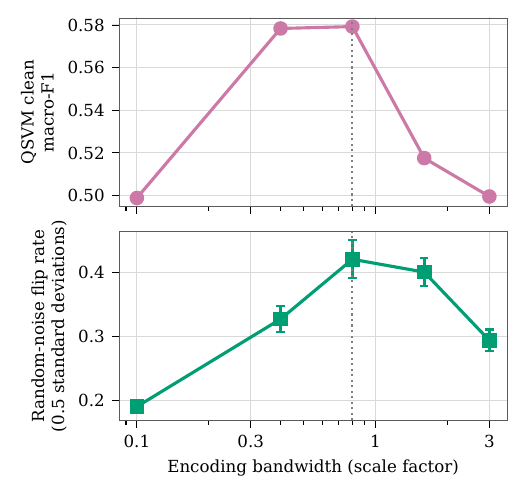}
\caption{Clean macro-F1 (top) and random flip rate at $0.5\sigma$ (bottom, mean $\pm$ std
over three draws, 100 test points) vs.\ encoding bandwidth. The dotted line marks the
accuracy-optimal bandwidth in both panels.}
\label{fig:fragility}
\end{figure}

\subsection{The Matched Classical Control}
Read alone, a quantum-only sweep cannot tell a
property of this feature map from a property of kernel machines. So we ran the identical
protocol on the classical RBF-SVM. We generated the same perturbation set once and reused it
across widths. We reselected $C$ per width by the same three-fold cross-validation over the same
grid, and used the same 100 evaluation points. We swept $\gamma$ about its tuned value in
place of bandwidth. At $\gamma = 419.1$ ($\times100$ the tuned value) the clean macro-F1 is 0.474 and
the flip rate at $0.5\sigma$ is 0.04. At $41.91$ ($\times10$) the two are 0.534 and 0.15, at the
tuned $4.191$ they are 0.558 and 0.23, at $0.4191$ ($\times0.1$) they are 0.618 and 0.18, and at
$0.04191$ ($\times0.01$) they are 0.536 and 0.42. The accuracy optimum, 0.618, sits at
$\gamma$ $\times0.1$; the flip-rate maximum, 0.42, at $\times0.01$.

The accuracy optimum ($\gamma = 0.419$) and the fragility maximum ($\gamma = 0.042$) are
different settings. Flip rate falls monotonically as the kernel narrows
($0.23 \to 0.15 \to 0.04$). This is the published classical direction~\cite{fawzi2015analysis},
reproduced here on grid data with a different robustness measure. The quantum sweep does the
opposite: clean macro-F1 $0.499/0.578/\mathbf{0.579}/0.517/0.499$ against flip
$0.190/0.327/\mathbf{0.420}/0.400/0.293$, so its accuracy optimum \emph{is} its fragility
maximum. One caveat is recorded rather than hidden. The narrowest classical cell
($\gamma = 419$, flip $0.04$) is partly collapsed, predicting the majority class on 96\% of
points, so it is not the basis of the comparison. The accuracy optimum sits at a healthy 78\%
predicted-positive rate against a 71\% true rate.

\subsection{A Second Control, Because the First Varies Two Things at Once}
The RBF-SVM differs
from the fidelity kernel on two axes at once: classical versus quantum, and \emph{decaying}
versus \emph{bounded and periodic}. So the contrast above cannot say which axis carries the
reversal. We therefore added a third arm that is bounded and periodic but unentangled, and
needs no simulator. The product RY feature map has the closed form
$k_b(x,y) = \prod_i \cos^2\!\big(b(x_i - y_i)/2\big)$. We verified it against the shipped
statevector path to a maximum absolute deviation of $3\times10^{-15}$ before using it. Under the
identical protocol and the same perturbation set, it does \emph{not} co-peak: clean macro-F1
$0.548/0.536/\mathbf{0.585}/0.554/0.569$ against flip
$0.127/0.200/0.210/\mathbf{0.413}/0.410$ at bandwidths $0.1$--$3.0$. Its accuracy optimum
($b = 0.8$) sits at half the sweep's maximum flip rate. We then ranked the arms by flip rate at
the accuracy optimum, as a fraction of each sweep's own maximum: RBF $0.43$, product RY $0.51$,
ZZ $1.00$. So boundedness and periodicity are not sufficient. The sharpest form is a
matched pair. At the same bandwidth $0.8$, with the same 100 points and the same
perturbations, the unentangled kernel matches the entangled one on clean accuracy
($0.585$ versus $0.579$) at half its flip rate ($0.210$ versus $0.420$). No cell here is
stable by collapse: predicted-positive rates run $0.45$--$0.62$ against a $0.71$ true rate.

\subsection{Entanglement Strength Does Not Order the Co-Peak}
The two maps
above differ in entanglement, rotation structure, repetition count, and the quadratic bandwidth
scaling on the pair terms. We isolated the first of these. The ZZ ring block is a
controlled-NOT gate (\texttt{CNOT}), then $R_Z(2 s\, x_i x_j)$, then another \texttt{CNOT}.
At $s = 0$ the rotation is the identity and the two \texttt{CNOT} gates cancel exactly. This
leaves the same map with the same repetitions and scaling but no entanglement. We swept
$s \in \{0, 0.25, 0.5, 0.75, 1\}$ across the same bandwidth grid. The ratio runs $0.959$,
$1.000$, $0.766$, $0.752$, $1.000$. This is nonmonotone, with the \emph{zero}-entanglement
end among the highest, so we claim nothing from it. At $s = 1$ the parameterized circuit
reproduces the deployed one exactly (all five bandwidths, both quantities, to four decimals).
This is what licenses reading the $s < 1$ rows at all. The $s = 0$ map carries no pair term yet
still reaches $0.959$, while the product RY map reaches $0.51$. But the ratio does not
distinguish an aligned peak from a flat, uniformly fragile profile, and the $s = 0$ sweep is the
flat case ($0.39$, $0.39$, $0.40$, $0.41$ above the degenerate end), where the deployed map has
an interior peak ($0.33$, $\mathbf{0.42}$, $0.40$, $0.29$). So the $s = 0$ arm shows only that
a map without entanglement is as fragile at its accuracy optimum ($0.39$) as the deployed one
($0.42$). It cannot say what shapes the peak. A re-score that also measures how far the peak
stands out reads the five values of $s$ as tied top (the flat profile), co-peak, misaligned,
misaligned and, at $s = 1$, a tie: two bandwidths differ by 0.0009 in accuracy, and the map
co-peaks at one and is misaligned at the other. Entanglement strength does not order that
sequence, so we neither credit the co-peak to entanglement nor rule it out. For the deployed
map, read at its validation-tuned bandwidth, the flip rate at the operating point
stands at least $0.127$ above both ends of its sweep, on a two-cell plateau ($0.420$ and
$0.400$). Both ends are cells at chance accuracy; measured only against cells above chance, the
peak is a tie with its neighbor. The classical RBF's accuracy pick sits $0.24$ below its most
fragile cell. So accuracy tuning puts the deployed map at a setting on its fragility peak,
which is what the headline claim says, and the cause stays open.

\subsection{The Reversal Tracks Stationarity}
Two other differences between the maps, rotation
structure and repetition count, are not independent, and the algebra collapses them before any
further measurement. At one repetition the Hadamard-plus-$R_Z$ (H+RZ) overlap is
$\prod_i \cos^2\!\big(b(x_i-y_i)\big)$. This is a function of $x-y$ alone, and hence the RY
kernel at twice the bandwidth (verified against the simulator to $2.2\times10^{-15}$,
shift-invariant to $1.6\times10^{-15}$). At two repetitions the per-qubit state
$e^{-it}\cos t\,\ket{0} - i e^{it}\sin t\,\ket{1}$ gives an overlap that depends on $t_x$ and
$t_y$ separately, violating shift-invariance by $0.492$. Repeating RY cannot do this, since
$R_Y(t)R_Y(t) = R_Y(2t)$. The contrast is therefore \emph{stationary} versus \emph{nonstationary}.
The decisive cell is H+RZ at one repetition: same gates as the co-peaking arm, but stationary. It
scores $0.460$. Ranking every arm by the same ratio gives two groups with no overlap. The three
stationary kernels group at $0.43$ (classical RBF), $0.460$ (H+RZ $\times$1), and $0.51$
(RY $\times$1). The nonstationary ones group at $0.713$ (H+RZ $\times$3) and $1.00$ (the deployed
entangled map). The
H+RZ $\times$2 arm is excluded here because its flip profile is the flat one identified above
(range $0.020$). Every retained arm has a genuine profile (range $0.127$--$0.266$). Under
the re-score every stationary arm reads misaligned. So do the two partly entangled maps
($s = 0.5$ and $0.75$), which are nonstationary, and H+RZ $\times$3 ties between misaligned and
tied top. On these arms nonstationarity is at most necessary for the co-peak, not sufficient. The
grouping is not quantum versus classical: the classical RBF and two quantum product maps sit
together on the stationary side. Two limits are worth stating. The relationship is graded rather
than a switch (a three-repetition map reaches only $0.713$, where we had expected it to match the
two-repetition arm). And there are three arms per class, so we report an association with a
checkable property rather than a mechanism. One untested account fits all six arms: a stationary
kernel looks the same everywhere, so accuracy tuning could not concentrate fragility at one
operating point. We do not rest anything on it.

We also tested the most natural mechanism, and it is insufficient. Treating fragility as
first-order, a perturbation $\delta \sim U(-\varepsilon\sigma, \varepsilon\sigma)$ shifts the
decision value by $\nabla f \cdot \delta$. This gives a predicted flip rate
$2\Phi(-|f|/(\varepsilon\|\nabla f\|_2/\sqrt{3}))$. Across 20 cells and four feature maps this
predicts the measured random-flip rate well ($r = 0.90$, $0.95$ excluding cells at chance
accuracy). It overestimates the level by a near-constant factor of $\approx\!2$. But it cannot
produce the co-peak. The predicted rate is monotone in bandwidth for \emph{every} arm, because
$\|\nabla f\|$ grows with bandwidth (1.9 to 78 for the deployed map) while the margin does not
keep pace. The deployed map's measured rate instead peaks in the interior and falls
($0.19/0.33/\mathbf{0.42}/0.40/0.29$). The bandwidth minimizing $|f|/\|\nabla f\|$ is the widest
for all four arms. But the bandwidth maximizing measured flips coincides with it in only one arm,
the flat-profile arm, where an argmax means least. So the co-peak is not a first-order effect.
One account is consistent with both this and the HopSkipJump overstatement. If decision regions
interleave at fine scale, a wide-bandwidth perturbation crosses many boundaries, and an even
number of crossings restores the original label. Under that account the measured rate would
saturate and fall, while a single-crossing model keeps rising. The first half of this account is now measured: crossings
per ray do rise with bandwidth, to 3.84 at the widest setting
(Appendix~\ref{app:crossings}). The second half remains a hypothesis. The obvious test of it is not
one, because crossing parity is algebraically identical to a flip and so confirms itself on any
data. What the measurement does add is that the counts are not independent. At 3.84 crossings per
ray the flip rate is 0.340, where independence would give 0.500. So the saturation account needs
the clustering structure, which a later count model does not settle
(Appendix~\ref{app:crossings}).

This runs against the expected ordering. Robustness to random noise normally
\emph{exceeds} adversarial robustness by a factor scaling as $\sqrt{d}$. This is proven for
linear classifiers and observed to persist for polynomial SVMs~\cite{fawzi2015analysis},
though not established for kernels of the type used here. Our QSVM runs the
other way: random $\ell_\infty$ noise at $0.5\sigma$ flips 45\% of its predictions, more
than any other family, while a decision-based search at about 600 queries per point reports a
median distortion of $3.41\sigma$, averaged over restarts (Section~\ref{sec:blackbox} reads it by the model's clean label and at larger budgets). We
therefore report the random control as a required companion to any minimum-distortion
number, rather than as a curiosity.

Repeating the sweep on three-class labels replicates the conclusion but not the exact
alignment. We report the difference rather than the agreement alone. Flip rate there is
$0.33$, $0.58$, $0.58$, $0.46$, $0.36$ at bandwidths $0.1$--$3.0$ against clean macro-F1
$0.300$, $0.380$, $0.442$, $0.540$, $0.460$. Accuracy peaks at $1.6$ while fragility peaks at
$0.4$--$0.8$, so the argmax coincidence is a binary-task property. What holds on both label
sets is the part the finding rests on. Fragility is elevated across the useful-accuracy
range and lowest exactly where accuracy is worst, so no bandwidth buys good accuracy together
with a low flip rate.

\section{Deployability: Training and Inference Cost}\label{app:cost}

Table~\ref{tab:cost} gives the per-model timings behind the deployability sentence in
Section~\ref{sec:rq3}. Quantum rows are statevector-simulator measurements, not device
timings, and are labeled as such.

\begin{table}[!t]
\centering
\setlength{\tabcolsep}{3.5pt}
\begin{threeparttable}
\caption{Deployability at PCA-8: training wall-clock and per-sample inference latency}
\label{tab:cost}
% auto-generated by qgridbench.eval.tables -- do not edit by hand
\begin{tabular}{lrrr}
\toprule
Model & Train (s) & Inference ($\mu$s/sample) & Relative to fastest \\
\midrule
Logistic regression & 0.01 & 0.07 & 1$\times$ \\
MLP & 1.85 & 2.37 & 33$\times$ \\
XGBoost & 0.78 & 5.40 & 76$\times$ \\
LightGBM & 3.76 & 12.18 & 171$\times$ \\
Random forest & 2.46 & 26.03 & 365$\times$ \\
VQC$^{a}$ & 92.81 & 136.72 & 1918$\times$ \\
RBF-SVM & 0.40 & 242.73 & 3405$\times$ \\
QSVM (RY-angle)$^{a}$ & 4.81 & 2233.24 & 31\,327$\times$ \\
QSVM (ZZ)$^{a}$ & 10.82 & 5352.39 & 75\,080$\times$ \\
\bottomrule
\end{tabular}
% a = simulated quantum model; latency is statevector-simulator time,
% not a hardware measurement, and is not a deployment claim.

\begin{tablenotes}[flushleft]
\footnotesize
\setlength{\labelsep}{0pt}
\item[] $^{a}$Simulated quantum model: statevector-simulator timings, not device measurements.
\end{tablenotes}
\end{threeparttable}
\end{table}

\section{First Account of the White-Box Gap: Kernel Decay (Superseded in \texorpdfstring{Appendix~\ref{app:objreg}}{a Later Appendix})}\label{app:exhaustion}

This appendix gives the first account of the white-box gap. Appendices~\ref{app:lengthscale}
and~\ref{app:objreg} correct it in turn, and the last gives the reading the body
uses. All of these appendices read an attack at its largest radius ($2\sigma$). They describe where
a full-budget attack ends, which is not the lowest score inside that ball. The worst-radius
readings, themselves upper bounds on what the attacker can do, are
in Appendices~\ref{app:lengthscale} and~\ref{app:qbw}, and the body reads every white-box attack
that way (Section~\ref{sec:rq2}). Table~\ref{tab:exhaustion} gives the kernel-mass measurements behind this first
account; Section~\ref{sec:blackbox} gives the current reading. Local input-sensitivity does \emph{not} separate the two kernel
machines. The $\sigma$-weighted gradient norm differs by only $2.4\times$ in probability
space. Single-step FGSM at $\varepsilon = 0.01\sigma$ moves them almost identically
(mean $|\Delta p|$ 0.183 vs.\ 0.173, against 0.004 for the MLP). The separation is a
large-radius property. As the
budget grows, the tuned RBF kernel decays toward zero, its predictions become constant,
and macro-F1 returns to the constant-predictor floor. At the full budget the attacker has disabled
it, not steered it; at smaller radii the same attack inverts it
(Appendix~\ref{app:lengthscale}). Neither quantum model does this \emph{under a loss-maximizing attack}. The
fidelity kernel's mass bottoms at $0.145$, and the VQC's $|\langle Z\rangle|$ \emph{rises} above
clean ($1.10$ of clean at $0.1\sigma$). Both end far below the constant-predictor floor, so the attacker keeps
steering authority in both families. The qualifier matters and is quantified in
Appendix~\ref{app:objreg}: the absence of decay is a property of that objective, not of the
bounded periodic encoding we first attributed it to.

\begin{table}[!t]
\centering
\setlength{\tabcolsep}{3.5pt}
\begin{threeparttable}
\caption{Behavior under PGD as the budget grows}
\label{tab:exhaustion}
% auto-generated by qgridbench.eval.tables -- do not edit by hand
\begin{tabular}{lrrrrrr}
\toprule
& \multicolumn{2}{c}{Quantum kernel} & \multicolumn{2}{c}{Variational (VQC)} & \multicolumn{2}{c}{Classical RBF} \\
\cmidrule(lr){2-3}\cmidrule(lr){4-5}\cmidrule(lr){6-7}
$\varepsilon$ ($\sigma$) & $m/m_0$ & F1$-$const. & $m/m_0$ & F1$-$const. & $m/m_0$ & F1$-$const. \\
\midrule
0 & 1.000 & +0.168 & 1.000 & +0.062 & 1.000 & +0.191 \\
0.01 & 0.997 & $-$0.151 & 1.002 & +0.055 & 0.990 & $-$0.067 \\
0.05 & 0.964 & $-$0.372 & 1.033 & $-$0.042 & 0.860 & $-$0.245 \\
0.1 & 0.897 & $-$0.391 & 1.102 & $-$0.091 & 0.763 & $-$0.252 \\
0.2 & 0.769 & $-$0.405 & 1.357 & $-$0.188 & 0.640 & $-$0.256 \\
0.3 & 0.632 & $-$0.396 & 1.723 & $-$0.288 & 0.460 & $-$0.212 \\
0.5 & 0.468 & $-$0.372 & 2.515 & $-$0.385 & 0.128 & $-$0.025 \\
1 & 0.267 & $-$0.290 & 3.598 & $-$0.415 & 0.006 & $-$0.002 \\
2 & 0.179 & $-$0.235 & 3.625 & $-$0.415 & 0.000 & +0.000 \\
\bottomrule
\end{tabular}

\begin{tablenotes}[flushleft]
\footnotesize
\setlength{\labelsep}{0pt}
\item[] $m/m_0$ is bounded-output mass relative
to clean: kernel-row mass for the kernel machines, $|\langle Z\rangle|$ for the VQC.
F1$-$const. is macro-F1 minus the constant-predictor floor (about 0.415; the end-state
thresholds are those of Table~\ref{tab:regime}). At $2\sigma$ the predicted-probability spread
reads 0.453 (quantum kernel), 0.144 (VQC), and 0.000 (RBF).
\end{tablenotes}
\end{threeparttable}
\end{table}

One qualification on the VQC row: its mass ratio is measured against a small base. Its
clean mean $|\langle Z\rangle|$ is $0.064$ of the attainable $1.0$, so the $3.6\times$
growth reaches only $\approx 0.23$. The key evidence is therefore the
\emph{absence of decay} together with macro-F1 below the constant-predictor floor, not the magnitude gain.
The VQC's bounded head caps it well below the fidelity kernel's $10.4\times$.

\section{Length Scale, Not Model Family, Selects the Full-Budget Failure Mode}\label{app:lengthscale}

Table~\ref{tab:regime} and Fig.~\ref{fig:widthsweep} give the control that fixes the scope of
the claim above, and it changes its shape. We hold $C$, the subset, the evaluation points, the radii, and the PGD
budget fixed, and move \emph{only} the RBF bandwidth. This takes a single classical model
through both end states and the partial label between them. Its clean macro-F1 stays within
0.533--0.607 throughout. The
tuned kernel ($\gamma = 4.19$) returns to the constant-predictor floor exactly. At $\gamma = 0.126$ the same
model lands $0.173$ below that floor, although its clean macro-F1 of 0.607 is
indistinguishable from the tuned kernel's 0.606 on these 200 points. At $\gamma = 0.0042$ it is
fully inverted. So ``the classical kernel can be disabled'' is a statement about the
bandwidth that tuning selected, not about classical kernels. Bandwidth is already known to move RBF-SVM
adversarial robustness. The study in~\cite{fawzi2015analysis} reports roughly
double the robustness at $\sigma^2 = 0.1$ against $\sigma^2 = 1$ on the Modified National
Institute of Standards and Technology (MNIST) handwritten-digit images, which is the same
direction. What is new here is the \emph{end state} it selects, and that an
accuracy-matched classical kernel inverts rather than shutting off.

These are readings at the full $2\sigma$ budget, and an attacker bounded by $2\sigma$ may stop
earlier. Read at its worst radius inside that ball, every one of the eight RBF widths is inverted,
0.27 to 0.42 below the constant-predictor floor, and each holds that label over at least four
of the eight radii. The tuned kernel reads $-0.273$ at $0.1\sigma$,
and the $\gamma = 0.126$ kernel $-0.386$ at $0.5\sigma$. So the width decides what the largest
radius shows, and a disabled reading there is not protection. We first listed the kernel length
scale among the evaluation choices; under the worst-radius reading its two readings carry
the same label, and the radius at which a white-box attack is read took its place
(Section~\ref{sec:rq2}).

\begin{table}[!t]
\centering
\setlength{\tabcolsep}{3.5pt}
\begin{threeparttable}
\caption{End state under a full-budget ($2\sigma$) white-box attack}
\label{tab:regime}
% auto-generated by qgridbench.eval.tables -- do not edit by hand
\begin{tabular}{@{}lrrrl@{}}
\toprule
Configuration & \begin{tabular}[b]{@{}r@{}}Clean\\ macro-F1\end{tabular} & $m/m_0$ & F1$-$const. & End state \\
\midrule
RBF $\gamma=4.191$ (tuned) & 0.606 & 0.000 & +0.000 & Disabled \\
RBF $\gamma=1.257$ ($\times$0.3) & 0.587 & 0.002 & +0.003 & Disabled \\
RBF $\gamma=0.4191$ ($\times$0.1) & 0.578 & 0.023 & +0.036 & Disabled \\
RBF $\gamma=0.2095$ ($\times$0.05) & 0.582 & 0.076 & $-$0.041 & Partial \\
RBF $\gamma=0.1257$ ($\times$0.03) & 0.607 & 0.158 & $-$0.173 & Partial \\
RBF $\gamma=0.08382$ ($\times$0.02) & 0.563 & 0.238 & $-$0.286 & Inverted \\
RBF $\gamma=0.04191$ ($\times$0.01) & 0.544 & 0.399 & $-$0.368 & Inverted \\
RBF $\gamma=0.004191$ ($\times$0.001) & 0.533 & 0.708 & $-$0.415 & Inverted \\
QSVM ZZ, 8 qubits & 0.583 & 0.179 & $-$0.235 & Inverted \\
QSVM RY, 8 qubits & 0.525 & 0.550 & $-$0.415 & Inverted \\
QSVM ZZ, 12 qubits & 0.575 & 0.103 & $-$0.125 & Partial \\
VQC, 8 qubits & 0.478 & 3.625 & $-$0.415 & Inverted \\
VQC, 12 qubits & 0.517 & 5.007 & $-$0.415 & Inverted \\
\bottomrule
\end{tabular}

\begin{tablenotes}[flushleft]
\footnotesize
\setlength{\labelsep}{0pt}
\item[] $m/m_0$ is bounded-output
mass relative to clean, and F1$-$const. is macro-F1 minus the constant-predictor floor (about
0.415). Labels are derived from fixed thresholds, not assigned by inspection. Disabled:
$m/m_0 \le 0.05$ and $|$F1$-$const.$| \le 0.05$. Inverted: F1$-$const. $\le -0.2$. Partial:
neither disabled nor inverted. Only $\gamma$ varies across the RBF rows; $C$ and every protocol element are held.
\end{tablenotes}
\end{threeparttable}
\end{table}

\begin{figure}[!t]
\centering
\includegraphics[width=\columnwidth]{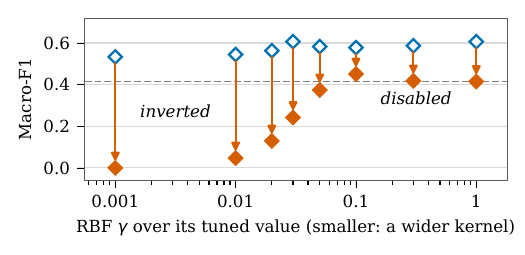}
\caption{RBF rows of Table~\ref{tab:regime}: one RBF-SVM as its $\gamma$ varies (a smaller
$\gamma$ is a wider kernel), clean (hollow) and under the full-budget ($2\sigma$) white-box attack
(filled), on 200 test points. The dashed line is the constant-predictor floor; the two points
with no word are partial, between the two end states.}
\label{fig:widthsweep}
\end{figure}

The classical behavior follows the closed form. A full-budget displacement suppresses a
kernel row by roughly $\exp(-\gamma\,\varepsilon^2\|\sigma\|^2)$, with
$\varepsilon^2\|\sigma\|^2 = 308$ at $\varepsilon = 2\sigma$ over eight PCA dimensions,
which places the crossover where it is measured. We present that half as a sanity check
rather than a result. For the RBF the group $u = \gamma\varepsilon^2\|\sigma\|^2$ is the
kernel's own exponent, so its collapse is close to analytic. The claim rests on the
\emph{quantum} kernel, which has no such closed form, collapsing onto the same group with
leave-one-$\gamma$-out median error $0.024$.

What survives, \emph{for this attack}, is that no tested quantum configuration is disabled.
The ZZ and RY fidelity kernels and the VQC all end inverted at
eight qubits. At 12 qubits the fidelity kernel retains $0.103$ of its mass against the
RBF's $0.000$. We originally read that as a structural consequence of a bounded, periodic
encoding having no decay regime. Appendix~\ref{app:objreg} shows it is not: change the
attacker's objective and the same kernels decay. Two limits are worth stating plainly.
First, the width axis was remeasured in one dedicated ladder at the full 2$\sigma$ budget,
with 100 evaluation points and one fixed tuned configuration across widths. Its absolute
masses therefore differ slightly from the per-width runs above. Retained mass runs $0.164$ at
8 qubits, $0.096$ at 12, and $0.111$ at 16. So the decline from 8 to 12 qubits does
\emph{not} continue, and the disabled regime is
not reached on either half of its criterion at any tested width. A convergence control
rules out a weaker attack (doubling the budget at 16 qubits moves mass by $0.0035$). With one
seed per width, we claim ``does not continue to decline,'' never a rise, since 8/12/16 could
be read as noise around $\approx 0.12$. Second, the VQC rows vary qubit count only. Its encoding
is a fixed \texttt{AngleEmbedding} ansatz, so on the end-state axis the feature-map choice
is tested for the kernel and not for the variational model. The re-uploading control below
tests a VQC encoding variant for accuracy and white-box retention only.

\section{The Quantum Knob Moves the End State Too}\label{app:qbw}

We ran the same sweep on the quantum side, since bandwidth is the fidelity
kernel's own tuning knob (Table~\ref{tab:qbw}). Under the loss-maximizing attack, the kernel
does not reach the disabled state at any
setting. Retained mass is \emph{U-shaped} in bandwidth: $0.985$ at $0.05$, a minimum of
$0.145$ at $1.2$, back to $0.596$ at $3.0$. It never approaches the $0.05$ that would mark
a collapsed kernel. Both extremes are degenerate in opposite directions, and neither permits
decay \emph{to that attack}. At small bandwidth every state crowds toward $\ket{0\ldots0}$
and the Gram saturates
near all-ones, so an attack cannot reduce what is already at ceiling. At large bandwidth the
clean Gram has already fragmented to the $\sim 2^{-n}$ baseline, so the attacked kernel sits
at that same baseline.

\begin{table}[!t]
\centering
\begin{threeparttable}
\caption{Quantum encoding bandwidth swept over the same range as the classical control, at
fixed $C$ and identical attack budget, under the loss-maximizing attack, read at the full
$2\sigma$ budget}
\label{tab:qbw}
% auto-generated by qgridbench.eval.tables -- do not edit by hand
\begin{tabular}{lrrrrl}
\toprule
Bandwidth & \begin{tabular}[b]{@{}r@{}}Clean\\ macro-F1\end{tabular} & $m/m_0$ & Std $p$ & F1$-$const. & End state \\
\midrule
0.05 & 0.521 & 0.985 & 0.403 & $-$0.404 & Inverted \\
0.1 & 0.557 & 0.926 & 0.446 & $-$0.407 & Inverted \\
0.2 & 0.569 & 0.710 & 0.451 & $-$0.414 & Inverted \\
0.4 & 0.555 & 0.358 & 0.456 & $-$0.371 & Inverted \\
0.793\,(tuned) & 0.583 & 0.164 & 0.447 & $-$0.199 & Partial \\
0.8 & 0.583 & 0.155 & 0.434 & $-$0.210 & Inverted \\
1.2 & 0.587 & 0.145 & 0.334 & $-$0.043 & Partial \\
1.6 & 0.558 & 0.219 & 0.262 & +0.046 & Partial \\
2 & 0.580 & 0.263 & 0.252 & +0.047 & Partial \\
3 & 0.601 & 0.596 & 0.108 & +0.039 & Partial \\
\bottomrule
\end{tabular}

\begin{tablenotes}[flushleft]
\footnotesize
\setlength{\labelsep}{0pt}
\item[] F1$-$const. is macro-F1 minus the constant-predictor
floor (about 0.415). Retained mass
never reaches the $0.05$ that would mark a collapsed kernel, but see
Appendix~\ref{app:objreg}, where changing the objective alone reaches $0.027$ at the tuned
bandwidth. The end state still moves from inverted to near the constant-predictor floor. Every row is
nondegenerate (clean macro-F1 at least $0.105$ above the constant-predictor floor). End-state labels use the
same fixed thresholds as Table~\ref{tab:regime}. Partial means neither disabled (retained mass at most $0.05$
and macro-F1 within $0.05$ of the constant-predictor floor) nor inverted (macro-F1 at least $0.2$ below it).
Rows straddling one threshold differ less than the
labels suggest: bandwidths $0.793$ and $0.8$ sit at $-0.199$ and $-0.210$, either side of
the $-0.2$ inversion cut. Read the F1$-$const. column, not the label, for near-boundary rows.
Clean macro-F1 here is not comparable to the bandwidth-fragility sweep, which reselects $C$
per bandwidth by cross-validation and evaluates on 100 points. This table holds $C$ at the
tuned value so that only bandwidth varies, and evaluates on 200.
\end{tablenotes}
\end{threeparttable}
\end{table}

The end state at the full budget nevertheless moves. The accuracy-tuned bandwidth ($0.793$) ends $0.199$ below
the constant-predictor floor. In contrast, $\varepsilon$-matched attacks on bandwidths $1.6$--$3.0$ land
\emph{at} that floor. Bandwidth $3.0$ does so at clean macro-F1 $0.601$ on this table's
200 evaluation points. On a 2000-point validation sample the same setting costs about
0.019 clean against the accuracy optimum, so the price is small rather than zero. Selected
against an adversarial objective at $0.5\sigma$, the one radius that selection ran, the same
width gives $+0.42$ macro-F1
under the white-box attack (0.010 to 0.427), an upper bound on its worst case. For the classical RBF at that
radius, the accuracy-optimal and robustness-optimal widths coincide. Neither is a safe operating
point. Read at its
worst radius inside the $2\sigma$ ball, every bandwidth in the table sits 0.28 to 0.42
below the constant-predictor floor, in the inverted band. That is a single-radius reading. By
the rule that a label holds only across a run of at least three radii, nine of the ten
bandwidths are inverted. Bandwidth $3.0$ is the exception: it reads 0.28 and 0.21 below that
floor at $0.01\sigma$ and $0.05\sigma$ and within 0.06 of it from $0.1\sigma$ on. Its
$0.01\sigma$ reading is lower than every reading at a larger radius; it comes from one run and
we have not reproduced it. So the bandwidth,
like the RBF width, changes what a full-budget reading shows; for nine of the ten settings it
does not change whether the attacker can invert the detector. The quantum failure form at the full budget differs from the classical one.
Prediction spread never falls below $0.108$, against $0.000$ for a disabled RBF, so the
failure is uninformative output rather than constant output. The setting also lies outside the
bandwidth box the accuracy protocol searches, so accuracy tuning alone would never find it.

The security reading is that the two end states are not equally bad. A disabled detector
emits a constant prediction, which an operator can monitor for directly. An inverted
detector reports the wrong class confidently and looks like it is working. At the full budget
both families let the defender move between those regimes by tuning, and in both the accuracy-optimal setting
is not the robustness-optimal one. At its worst radius inside that budget, every setting in both tables
reads in the inverted band, and all but bandwidth $3.0$ hold that label over a run of radii. We
do \emph{not} claim the families differ in the endpoint
reachable. Appendix~\ref{app:objreg} reaches the disabled endpoint on the quantum side too.

\section{Measuring Multiple Crossings Directly}\label{app:crossings}

The likely cause in Section~\ref{sec:blackbox}, a decision boundary that a path crosses more
than once, was inferred from two indirect signals: the
random-perturbation control and the white-box ranking reversal. Table~\ref{tab:crossings}
measures it. We sample the decision function at 200 points along a fixed
perturbation ray inside the $0.5\sigma$ ball for each of 100 test points, and count sign changes. We do this for the
fidelity kernel across bandwidth, and for the tuned RBF-SVM across a matched $\gamma$ span on the
\emph{identical} rays.

\begin{table*}[!t]
\centering
\begin{threeparttable}
\caption{Decision-region crossings along a fixed $0.5\sigma$ perturbation ray (binary, PCA-8,
200 samples per ray, 100 test points, identical rays for both families)}
\label{tab:crossings}
% auto-generated by qgridbench.eval.tables -- do not edit by hand
\begin{tabular}{llrrrr}
\toprule
Kernel & Length scale & Clean macro-F1 & Crossings/ray & Max & $>$1 crossing \\
\midrule
Fidelity (ZZ) & $\sigma_{\mathrm{bw}}=0.1$ & 0.499 & 0.19 & 1 & 0\% \\
Fidelity (ZZ) & $\sigma_{\mathrm{bw}}=0.4$ & 0.578 & 0.41 & 1 & 0\% \\
Fidelity (ZZ) & $\sigma_{\mathrm{bw}}=0.8$$^{a}$ & 0.579 & 1.01 & 3 & 26\% \\
Fidelity (ZZ) & $\sigma_{\mathrm{bw}}=1.6$ & 0.517 & 2.35 & 8 & 63\% \\
Fidelity (ZZ) & $\sigma_{\mathrm{bw}}=3.0$ & 0.499 & 3.84 & 19 & 69\% \\
\midrule
RBF (classical) & $\gamma\times0.01$ & 0.546 & 0.42 & 2 & 2\% \\
RBF (classical) & $\gamma\times0.1$ & 0.533 & 0.65 & 3 & 14\% \\
RBF (classical) & $\gamma\times1$$^{a}$ & 0.576 & 0.58 & 2 & 17\% \\
RBF (classical) & $\gamma\times10$ & 0.541 & 0.18 & 2 & 2\% \\
RBF (classical) & $\gamma\times100$ & 0.479 & 0.03 & 1 & 0\% \\
\bottomrule
\end{tabular}
% a = that family's accuracy optimum, the comparison the paper quotes.

\begin{tablenotes}[flushleft]
\footnotesize
\setlength{\labelsep}{0pt}
\item[] ``$>$1 crossing'' is
the fraction of rays that cross the decision boundary more than once, where a bisection such as
HopSkipJump's can stop at a far crossing.
$^{a}$Marks each family's accuracy optimum, which is the comparison quoted in
Section~\ref{sec:blackbox}. The wide-length-scale rows are the larger effect, but they are not
where the deployed model sits. Parity of the crossing count is deliberately not reported. It is
identical to ``the endpoints differ in sign'' and therefore tests nothing.
\end{tablenotes}
\end{threeparttable}
\end{table*}

There are two separations of different sizes, and the smaller one matters more. At each
family's own accuracy optimum, where the deployed model sits and where HopSkipJump actually
runs, the fidelity kernel crosses the boundary more than once on 26\% of rays against the
classical kernel's 17\%. That is a real but modest $1.7\times$. The larger divergence is at
wide length scale. There the quantum count climbs monotonically to 3.84 crossings per ray (one
ray crosses 19 times), while the classical count is nonmonotone and falls to 0.03. A
degenerate control passes: at bandwidth 0.1 the model scores at the stratified-random floor and shows 0.19
crossings with a maximum of 1, so the scan reads decision geometry rather than numerical noise.

What this does \emph{not} settle is the accuracy--fragility co-peak. We preregistered a parity
test: a flip should occur exactly when a ray crosses an odd number of boundaries. The test is
vacuous. Counting sign changes along a path and asking whether the count is odd is
algebraically identical to asking whether the endpoints differ in sign, so it reproduces the
measured flip rate exactly on any data, including noise. The live clue is that the flip rate
falls to 0.340 at bandwidth 3.0 while crossings climb to 3.84, where independent crossings would
give 0.500. A count model fitted afterwards to a new scan of 400 rays per cell compares paired
with independent crossings. Paired crossings fit better at bandwidth 3.0 (Akaike information
criterion lower by 60) and worse at the deployed bandwidth (higher by 42); at the registered
cell, bandwidth 1.6, the two fit alike (a difference of 0.3). The registered test is therefore
neither passed nor failed, and the comparison at 3.0 was not registered, so the co-peak's
mechanism stays open.

\section{The Surrogate Control}\label{app:surrogate}

Table~\ref{tab:surrogate} is the evidence behind Section~\ref{sec:surrogate}, and
Fig.~\ref{fig:transfer} shows the asymmetry that the control removes. In the first two
rows the surrogate is refitted with balanced sample weights and nothing else changes. The
estimator class, tuned hyperparameters, seeds, and evaluation subset stay the same, and so does
the hand-derived input gradient used for every published transfer number. Read the
classical-damage column first. It is flat across those two fits (0.208, 0.204), so the
corrected surrogate is not simply a weaker attack. The VQC column is where the change lands.

\begin{table*}[!t]
\centering
\begin{threeparttable}
\caption{Surrogate control (binary, PCA-8, ZZ, $\varepsilon = 0.5\sigma$, ten seeds),
ordered by how far each surrogate clears the stratified-random floor}
\label{tab:surrogate}
\footnotesize
\setlength{\tabcolsep}{2.5pt}
% auto-generated by qgridbench.eval.tables -- do not edit by hand
\begin{tabular}{lrrrrrrrr}
\toprule
Surrogate & Clean macro-F1 & $-$floor & Positive rate & VQC damage & Classical damage & Reverse damage & Forward/reverse & QSVM retention \\
\midrule
MLP, as tuned & 0.465 & $-$0.031 & 0.966 & 0.268 & 0.208 & 0.058 & 4.62 & 0.867 \\
MLP, balanced weights & 0.532 & +0.035 & 0.547 & 0.211 & 0.204 & 0.242 & 0.87 & 0.741 \\
RBF-SVM (other class) & 0.587 & +0.090 & --- & 0.001 & 0.123 & 0.172 & 0.01 & 0.704 \\
\bottomrule
\end{tabular}

\begin{tablenotes}[flushleft]
\footnotesize
\setlength{\labelsep}{0pt}
\item[] Rows 1--2 change only
the fit: same estimator class, tuned hyperparameters, seeds, evaluation subset, and
input-gradient path. Their classical-damage column is flat, so the corrected surrogate is not
a weaker attack, and the VQC column is where the change lands. Row 3 changes the function
class (tuned RBF-SVM, exact verified input gradient, no imbalance defect by construction).
It is white-box against its own row, so its classical column excludes itself. It is also a
weaker attacker in absolute terms. Normalizing by classical damage gives $1.29$, $1.04$,
$0.008$, the same monotone decline. ``Positive rate'' is the fraction of the evaluation set the
surrogate calls positive against a true rate of 0.710. It is an MLP diagnostic and is blank
for row 3. ``Forward/reverse'' is the surrogate$\to$VQC damage over the VQC$\to$surrogate
damage, the ``reverse damage'' column. ``QSVM retention'' is the share of its clean macro-F1 the
QSVM keeps under each surrogate's transferred attack: a competent surrogate leaves it near 0.7,
far from its white-box 0.023.
\end{tablenotes}
\end{threeparttable}
\end{table*}

\begin{figure*}[!t]
\centering
\includegraphics[width=\textwidth]{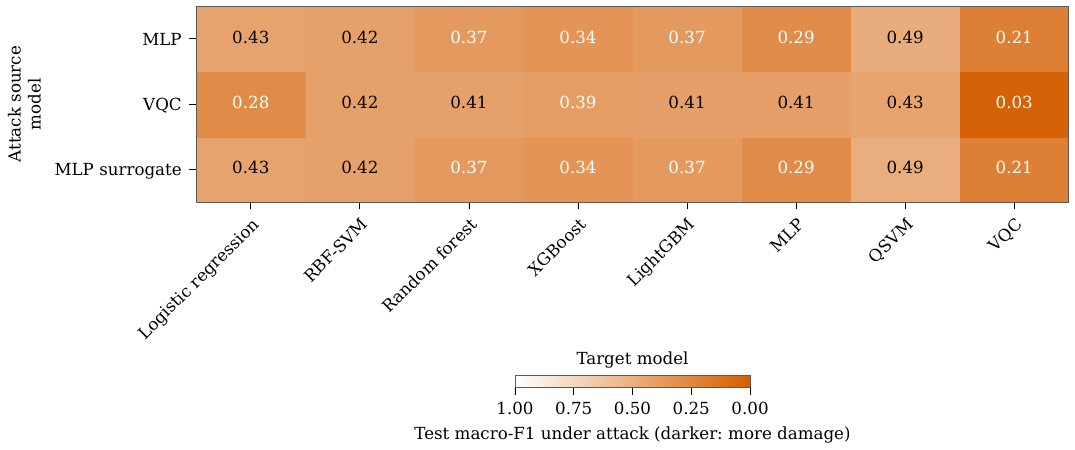}
\caption{Asymmetry that a mis-specified surrogate manufactures. MLP-crafted attacks
(row ``MLP''; row ``MLP surrogate'' is the same set on nondifferentiable targets)
appear to break the VQC harder than any classical target. The MLP here is fitted without
the imbalance policy, because its class carries no argument for it. Refitting it removes the
asymmetry (Section~\ref{sec:surrogate}) without changing any classical column.}
\label{fig:transfer}
\end{figure*}

Two fits of one estimator cannot separate ``this defect caused the artifact'' from ``the
artifact is specific to neural-net surrogates,'' so the third row changes the function class.
The tuned RBF-SVM is a kernel machine, and its estimator accepts \texttt{class\_weight}, so it
carries no imbalance defect by construction. Its input gradient is hand-derived and
verified against central finite differences. This makes it a genuine white-box surrogate rather
than a score estimate. It is white-box against its own row, so its classical column excludes
itself. That arm is a weaker attacker in absolute terms (0.123 against 0.204), which is why the
normalized column matters. Quantum damage per unit classical damage falls $1.29 \to 1.04 \to 0.008$ across the three rows, the same monotone decline the raw ratio shows.
Refitting the MLP surrogate barely moves the other families' transferred retention: the trees
keep 0.60, 0.55, and 0.59 (random forest, XGBoost, LightGBM) under the as-tuned fit and 0.63,
0.59, and 0.59 under the balanced one, and logistic regression 0.85 and 0.79. The QSVM moves the
most, from 0.87 to 0.74, and under either fit every family keeps far more than the QSVM's
white-box 0.023.

The correction also fixes a baseline. The unweighted surrogate predicts the positive class
96.6\% of the time against a true rate of 71.0\%, and its clean macro-F1 sits 0.031 below the
stratified-random floor. Balanced weights move it to 0.035 above, and the RBF-SVM sits 0.090
above. An independently written model reaches 0.552 clean and a
forward/reverse ratio of $1.01$. This model is a two-layer network of the same shape, trained
with class-weighted cross-entropy instead of sample weights. The sample-weight fit, recomputed
under that same implementation, gives $0.87$. Those two disagree on which side of unity the
residual sits, which is why we claim no direction rather than a reduced one.

\subsection{The Mechanism Is Open}
 Our first account was a class-prior shift. An unweighted fit
on an imbalanced task has a loss gradient dominated by one global direction, so sign-based PGD
hands every point nearly the same corner of the ball. Two tests reject it. First, artifact damage
does not track a target's distance from the stratified-random floor (Spearman $0.107$, $p = 0.82$ over seven
targets). Logistic regression sits at the stratified-random floor with $-0.028$ artifact damage, while the VQC sits
at that floor with $+0.057$. Second, stripping the attack to its global component at full budget
moves the VQC by $-0.032$, where the point-specific residual moves it by $0.224$. The same
decomposition separates the two kernel machines cleanly. A single constant displacement reproduces
94\% of the RBF-SVM's damage, and a \emph{random} one reproduces 83\% of it.

\subsection{A Mechanism We Tested and Rejected}
 The natural explanation for the transfer
asymmetry, when we still read it as an asymmetry, is gradient alignment. A transferred
perturbation should damage a target in proportion
to its component along that target's own loss-ascent direction. It does not hold here. Across the
off-diagonal source--target cells, mean cosine alignment at the clean points is uncorrelated with
measured damage (Spearman $\rho = 0.20$, $p = 0.70$), with $0.398$ for MLP\,$\to$\,VQC against
$0.364$ for VQC\,$\to$\,MLP. The metric is sound. Self-transfer is the
maximum-alignment cell for each differentiable target ($0.712$, $0.554$), and matched-norm random
displacements score $\approx\!0$. Alignment to either kernel machine is indistinguishable from
random from both sources, yet transferred attacks still move them.

In hindsight, this null was the first, unrecognized sign of the surrogate artifact. If the
surrogate's fit manufactured the damage asymmetry, no directional effect existed for a
first-order directional measure to track, and the measure should then come out null. That is
what it did, before we had any reason to doubt the surrogate. The two instruments give one
answer.

\subsection{Adding the VQC to an Ensemble Does Not Help}
 A natural use for a weak but
``different'' model is diversity inside an ensemble. We tested it. We built equal-weight
soft-vote ensembles over the three and five strongest classical models, with and without each
quantum member, and evaluated them on four attack sets over ten seeds. Adding the VQC changes
adversarial macro-F1 by $-0.003$ to $-0.010$ and clean macro-F1 by $-0.003$ to $-0.005$, never
once helping. Adding the fidelity kernel does not help against VQC-crafted attacks either ($-0.010$
to $-0.003$) and costs $0.015$ clean macro-F1 in the three-model ensemble. The result is limited to equal-weight
soft voting. A learned combiner was not tested.

\section{At an Operator's Threshold, the Failure Depends on the Alarm Budget}\label{app:console}

Every metric in the body uses the argmax threshold. Here we fit a threshold on validation for
a 1\% false-positive budget instead. The full-feature models are usable clean (LightGBM
0.652 and random forest 0.663 detection), and they diverge under transferred PGD at $0.5\sigma$.
Random forest goes \emph{silent}: detection falls to 0.013 while its realized false-alarm
rate stays at 0.005. LightGBM, XGBoost, and logistic regression \emph{flood}. Their realized
false-alarm rate rises from 0.007, 0.002, and 0.014 to 0.920, 0.940, and 0.924 at the same
fixed threshold. At a 5\% budget all four flood: the false-alarm rates are 0.963, 0.961, and
0.926, and 0.795 for the random forest, so its silent reading holds at the 1\% budget only.
The full-feature MLP, the surrogate the attack is crafted on, has no finite threshold that
meets the 1\% budget and is left out. The matched-dimensionality models are
not deployable at any budget (0.003 to 0.041 detection at 1\%), so this reading needs
the full-feature regime to mean anything. These operating points are measured under the
row-level split (Section~\ref{sec:split}), and a leakage-free deployment estimate would sit
lower still.

\section{Is Our Own Headline Objective-Limited? A Control}\label{app:objctl}

Appendix~\ref{app:objreg} argues that an end-state claim is a claim about the attacker's objective.
That invites the same question about our headline: every retention number here comes from PGD
ascending cross-entropy. We tested it. At the headline budget, with ten attack seeds and nothing
retuned, cross-entropy is the strongest of three objectives for both kernel machines. The QSVM
falls to $0.033 \pm 0.006$ macro-F1 under cross-entropy and the margin loss alike, and only to
$0.501 \pm 0.025$ under the distance-seeking objective. The RBF-SVM falls to $0.401 \pm 0.012$,
$0.401 \pm 0.012$, and $0.415$. The ordering and its significance are unchanged
($p = 0.002$, paired over seeds).

The margin result is an identity, not a coincidence. For binary labels $y \in \{0,1\}$, let
$L_{\mathrm{BCE}}$ be the binary cross-entropy (BCE) loss of the decision value $f$, and $s(f)$ the logistic function. Then
$\partial L_{\mathrm{BCE}}/\partial f = s(f) - y = -(2y-1)\,|s(f) - y|$, so the
cross-entropy and margin gradients differ by a strictly positive per-point scalar, which
$\mathrm{sign}(\cdot)$ discards. Under sign-based PGD they are the same attack. We verified
componentwise sign agreement of $1.000000$ on both models. The equivalence fails only if the
cross-entropy factor underflows to exactly zero, so we measure it rather than assume it. At
most $0.2\%$ of points come within $10^{-6}$, and $\max|f|$ is $18.4$ and $9.24$.

The distance-seeking result mirrors Appendix~\ref{app:objreg}. There, the loss objective
under-attacks kernel mass by $6.2\times$. Here, the distance objective under-attacks macro-F1 by
more. Neither dominates, since each is strongest only for the quantity it targets. That is why we
frame objective-dependence as a property of the evaluation rather than as having found a better
attack.

\section{The End State Belongs to the Objective and the Fit, Not to the
Encoding}\label{app:objreg}

Every attack above ascends classification loss, and that is not a neutral choice.
For a decaying kernel, loss ascent implies distance: moving a point away from the training
set is \emph{how} the attacker raises its loss. So the kernel row collapses as a side
effect, and ``disabling'' comes for free. A bounded, periodic kernel breaks that coupling. The
attacker can raise loss by crossing into the opposite class's region while remaining
kernel-close, so it never buys distance. The lack of decay we reported is therefore a statement
about what the attacker wanted, not about what the encoding permits. Table~\ref{tab:objreg}(a)
tests this by changing the objective alone: same budget, steps, step size, seed, evaluation
points, and fitted model, but descending mean kernel mass instead of ascending loss. At the
tuned bandwidth the mass falls to $0.027$ of clean against $0.167$ for the loss objective, a
factor of $6.2$. That is below the $0.05$ we named above as the mark of a collapsed kernel. It is
also below the unrelated-pair baseline of $0.613$, precommitted as the mass of evaluation points
whose columns are independently permuted. At bandwidth $0.4$ and $8\sigma$ the attacker
reaches an absolute fidelity of $0.0014$, beneath the $2^{-8}$ random-state floor. The two
objectives converge ($1.0\times$) at the extremes, where the clean Gram is already saturated
or already fragmented and there is no mass left to remove. This is why the sweep that
varied bandwidth alone could not separate the objective from the encoding.

\begin{table}[!t]
\centering
\begin{threeparttable}
\caption{End state belongs to the objective and the fit}
\label{tab:objreg}
% auto-generated by qgridbench.eval.tables -- do not edit by hand
\begin{tabular}{@{}lrrrr@{}}
\toprule
\multicolumn{5}{@{}l}{\textit{\textbf{(a) Objective only: same budget, steps, seed, and model}}} \\
Bandwidth & Loss obj. & Mass obj. & Factor & Baseline \\
\midrule
0.05 & 0.985 & 0.971 & 1.0$\times$ & 1.000 \\
0.1 & 0.932 & 0.894 & 1.0$\times$ & 1.000 \\
0.2 & 0.712 & 0.559 & 1.3$\times$ & 0.989 \\
0.4 & 0.335 & 0.073 & 4.6$\times$ & 0.693 \\
0.793\,(tuned) & 0.167 & 0.027 & 6.2$\times$ & 0.613 \\
0.8 & 0.159 & 0.026 & 6.1$\times$ & 0.610 \\
1.2 & 0.155 & 0.094 & 1.7$\times$ & 0.511 \\
1.6 & 0.205 & 0.164 & 1.3$\times$ & 0.485 \\
2 & 0.261 & 0.269 & 1.0$\times$ & 0.496 \\
3 & 0.599 & 0.585 & 1.0$\times$ & 0.685 \\
\midrule
\multicolumn{5}{@{}l}{\textit{\textbf{(b) Regularization only: attacked mass held fixed, tuned bandwidth}}} \\
$C$ & Clean macro-F1 & $m/m_0$ & Spread & F1$-$const. \\
\midrule
1 & 0.593 & 0.028 & 0.119 & +0.000$\pm$0.000 \\
10 & 0.599 & 0.028 & 0.207 & +0.005$\pm$0.010 \\
100 & 0.573 & 0.027 & 0.407 & +0.011$\pm$0.019 \\
835.2\,(tuned) & 0.583 & 0.026 & 0.642 & +0.022$\pm$0.003 \\
\bottomrule
\end{tabular}
% a = clean model degenerate or already constant, so not read as disabled by the attack.

\begin{tablenotes}[flushleft]
\footnotesize
\setlength{\labelsep}{0pt}
\item[] (a) Objective is the only
variable: identical budget ($2\sigma$), steps, step size, seed, evaluation points, and
fitted model. ``Loss obj.'' ascends classification loss (every other attack in this paper),
and ``Mass obj.'' descends mean kernel mass. Both are reported relative to clean mass, against
the precommitted unrelated-pair baseline (column-permuted evaluation points). The two coincide at
the extremes, where the Gram is already saturated or already fragmented.
(b) Regularization is the only variable, at the tuned bandwidth with attacked mass
held at $0.026$--$0.028$. Clean macro-F1 is read on the 200 attacked test points. ``Spread'' is the attacked decision spread over the same model's
clean spread, which is invariant to a uniform rescaling of the duals and therefore comparable
across $C$. F1$-$const. is attacked macro-F1 minus the constant-predictor floor (about 0.415), averaged over three
attack seeds. No row is marked $^{a}$, because every clean model clears that floor and none predicts a constant class before the attack.
\end{tablenotes}
\end{threeparttable}
\end{table}

Panel (b) asks what then keeps the output alive. We hold the attacked mass fixed at
$0.026$--$0.028$ and vary only the QSVM's regularization. The decision-spread ratio (attacked
spread over that same model's clean spread, which is invariant to a uniform rescaling of the
duals) rises monotonically $0.119 \to 0.207 \to 0.407 \to 0.642$ as $C$ runs $1 \to 835$. At
$C \leq 10$ the model reaches modal-prediction fraction $\geq 0.995$ with macro-F1 at the
constant-predictor floor: kernel decayed, predictions constant, accuracy at chance. This is a
constant output on the quantum side, at this radius; the later sweep below reads the disabled
label at $1\sigma$ and $2\sigma$ only, one radius short of the three we ask of an end state. Those cells are not degenerate models: on these 200 test points $C = 10$
scores $0.599$ against the tuned model's $0.583$. The setting is not free, though. On 2000
validation rows $C = 10$ scores $0.526$ against $0.595$ for the tuned $C$, so accuracy tuning
does not pick it, and the constant output appears only under the distance-seeking objective. What
distinguishes the tuned fit is that its duals are pinned at a bound $835\times$ larger. So
residual kernel structure far below the random-state floor is still amplified into a varying
decision.

There are three limits. First, the difference between $C = 100$ and the tuned $C$ is
small ($+0.011$ vs $+0.022$ above the constant-predictor floor). We read the spread ratio, which separates cleanly, as
the primary evidence. The tuned model sits \emph{near} the line for a constant output rather than
safely away from it. Second, the classical positive control passed: the same optimizer drives the
tuned RBF to $0.000$ mass and a constant output, so a failure to collapse the quantum
kernel would have been readable. The optimizer also beat random perturbation at matched radius
in every cell inside the tested budget. Third, the VQC was tested on the objective half of this
axis (Section~\ref{sec:blackbox}). Its lack of decay is objective-independent: a readout-descending
attack suppresses $|\langle Z\rangle|$ to 0.357 of clean against 1.10 under loss ascent, and
never approaches a constant output. The objective still separates the outcomes: loss ascent
inverts it (macro-F1 0.000 at $2\sigma$), while magnitude descent leaves it 0.074 above the
constant-predictor floor (0.489 against 0.415). The regularization half does not apply to it.

Panel (b) is a reading at the full $2\sigma$ budget, and an attacker bounded by $2\sigma$ may
stop earlier. A later sweep ran both objectives at the eight tested radii, $0.01\sigma$ to
$2\sigma$, with ten attack starts for each $C$. The full-budget reading reproduces: under the
distance-seeking objective the decision-spread ratio is 0.113, 0.202, 0.433, and 0.696 for
$C = 1$, 10, 100, and 835. That objective never takes any $C$ more than 0.001 below the
constant-predictor floor at any radius. Inside the ball, loss ascent inverts the QSVM at every
$C$: at $0.3\sigma$ it reads 0.370, 0.392, 0.401, and 0.402 below that floor, and it holds the
inverted label at five, six, six, and seven of the eight radii. Random noise of $0.3\sigma$
leaves 0.49 to 0.53. So the regularization decides what a full-budget distance-seeking
reading shows, and a small $C$ is not protection. We first listed it among the evaluation
choices; under the worst-radius reading it left the list, as the kernel length scale did
(Section~\ref{sec:widthreg}).

\section{Evasion Degradation Curves}\label{app:evasion}

Fig.~\ref{fig:pgd} gives the full PGD degradation curves behind the
$\varepsilon = 0.5\sigma$ column of Table~\ref{tab:robust}. Note that the QSVM curve is
the transferred attack, whose white-box replacement is discussed in
Section~\ref{sec:rq2}.

\begin{figure}[!t]
\centering
\includegraphics[width=\columnwidth]{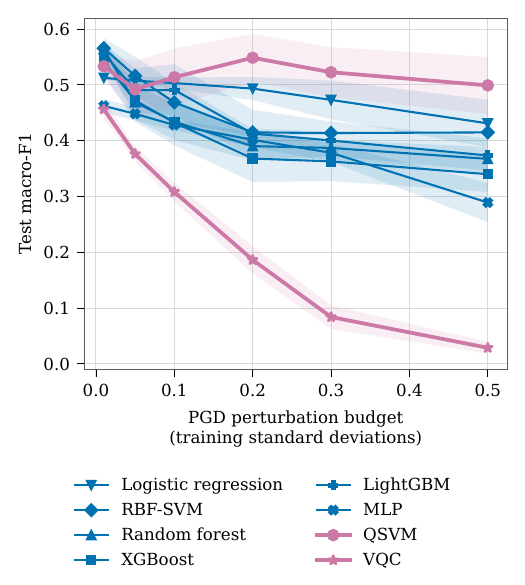}
\caption{PGD evasion degradation (ten seeds, mean $\pm$ std; white-box for MLP/VQC,
MLP-surrogate transfer otherwise).}
\label{fig:pgd}
\end{figure}

Single-step FGSM was run over the same grid and adds nothing. Across all 48 model
$\times$ $\varepsilon$ cells the two attacks agree to a mean absolute difference of 0.0075
macro-F1 (maximum 0.058, Spearman 0.994), so we report PGD only.

\section{Poisoning and Benign Sensor Noise}\label{app:poisonnoise}

Fig.~\ref{fig:poisonnoise} shows the two nonevasion ablations of
Section~\ref{sec:rq3}. Over five seeds, label-flip poisoning up to 20\% moves no classical
family other than the MLP by more than about 0.023 in mean macro-F1. On average it moves the MLP and the RY-angle QSVM by about 0.03, against a
seed spread of 0.02, and the VQC by 0.012, against 0.013. Under benign Gaussian sensor noise the strongest models lose
the most, so we report retention rather than raw degradation throughout.

\begin{figure*}[!t]
\centering
\includegraphics[width=\textwidth]{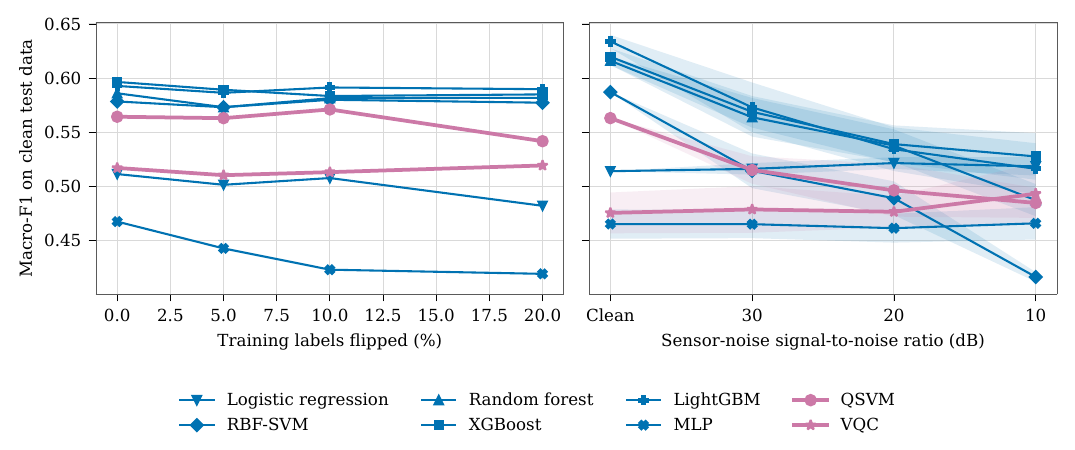}
\caption{Left: clean-test macro-F1 after training-label flips (the one-seed exhibit;
the five-seed spread is quoted in Section~\ref{sec:rq3}). Right: macro-F1 under additive
Gaussian sensor noise (ten seeds, mean
$\pm$ std; clean reference at left).}
\label{fig:poisonnoise}
\end{figure*}

\section{Trainability: The Plateau Is on the Qubit Axis}\label{app:trainability}

Fig.~\ref{fig:plateau} shows the trainability evidence behind Section~\ref{sec:rq3}.
The two panels measure different quantities. At initialization, single-parameter
gradient variance decays as $\approx 2^{-n}$ with qubit count $n$ (barren-plateau
signature; 40 initializations, 64 samples, depth 6). During training at eight qubits,
gradient-norm variance stays within 2\% of its mean across depths 2, 4, and 6. So the VQC's weak
accuracy at 8--12 qubits is model-class fit, not trainability collapse.

\begin{figure*}[!t]
\centering
\includegraphics[width=\textwidth]{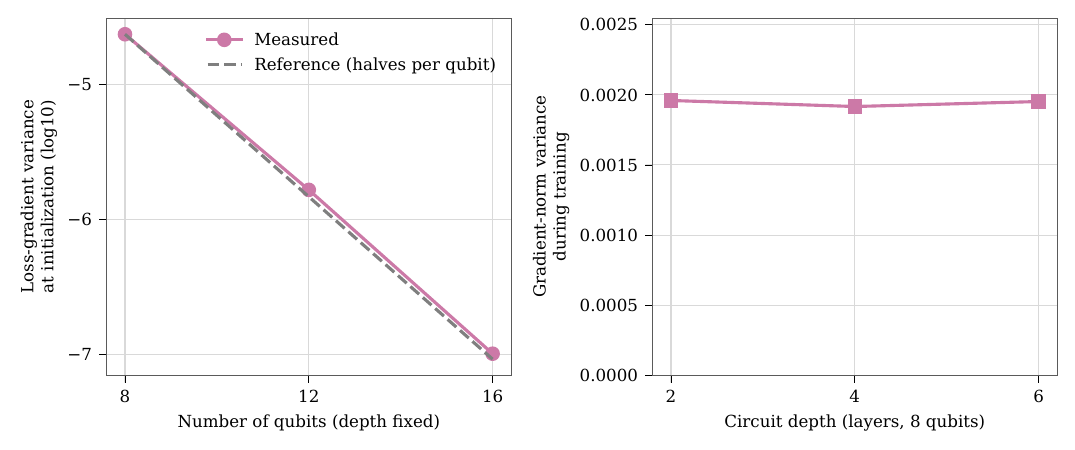}
\caption{Left: variance of the loss gradient for a single parameter at initialization vs.\ qubit count,
with a reference line that halves per qubit. Right: gradient-norm variance during training vs.\ depth
at eight qubits.}
\label{fig:plateau}
\end{figure*}

\subsection{Testing Whether the Ansatz Is the Problem}
 \texttt{AngleEmbedding} with strongly entangling
layers is the older variational form, and data re-uploading is the current default. So we
tested whether the choice carries our VQC results. We reencoded the data before every
trainable layer, and held fixed the weight tensor, parameter count, optimizer, loss, early
stopping, seeds, and training subset. Six configurations span 0.513--0.537 validation
macro-F1. This is entirely inside the deployed model's own 50-trial spread (0.498--0.540, best
0.540), and a matched-budget control on the deployed ansatz is indistinguishable. Fragility
is unchanged too: white-box retention at $0.5\sigma$ is $0.053 \pm 0.034$ for the
re-uploading model against $0.064 \pm 0.033$ for the deployed one. This is six
configurations at one seed against the deployed model's 50 tuning trials, so it is
evidence, not proof. A lift would have been decisive, and none appeared.

\section{Further Controls}\label{app:further}

Table~\ref{tab:further} lists checks the body has no room for. Each row gives what was asked,
what the run read, and the section it applies to. The rows are grouped as the body is: split
protocol and accuracy, attack protocol, physical consistency, and replication data and
reproducibility. A white-box row gives the radius its control ran; where that is one radius, the
reading is an upper bound on the worst case inside it.

\begin{table*}[p]
\centering
\caption{Further controls, one row per check}
\label{tab:further}
\fontsize{7.5}{8.7}\selectfont
% auto-generated by experiments/build_further_controls.py -- do not edit by hand
\begin{tabularx}{\textwidth}{@{}>{\raggedright\arraybackslash}p{0.27\textwidth}Y>{\raggedright\arraybackslash}p{0.2\textwidth}@{}}
\toprule
Check & Reading & Where it applies \\
\midrule
\rowcolor{qgChoiceTint}[0pt][0pt]\multicolumn{3}{@{}l@{}}{\textbf{Split protocol and accuracy}} \\
Do duplicated rows carry conflicting labels? & Among 78\,368 pooled rows the number that duplicate another row's features is 0, so no duplicate carries a conflicting label. Of the test rows with a training neighbor closer than 0.1, 0.4\% have one of the opposite label & Section~\ref{sec:rq1}; the null is not row-level label ambiguity \\\addlinespace[2pt]
The projection and cap costs with the hyperparameters held fixed (LightGBM, 16 components, one seed) & Under the set tuned at PCA-8, PCA costs $0.063$ and the cap $0.197$; under the full-feature set, $0.051$ and $0.257$. On all features, changing from the full-feature set to the PCA-8 set costs $0.042$ & Section~\ref{sec:rq1} and Fig.~\ref{fig:bridge} \\\addlinespace[2pt]
The classical lead over ten draws of the 2000 training rows per dimensionality (the published draw and nine more; three model seeds, tuned parameters unchanged) & XGBoost minus the best quantum kernel: $0.030 \pm 0.005$, $0.040 \pm 0.008$, and $0.037 \pm 0.007$ macro-F1 at 8, 12, and 16 dimensions (lowest $0.023$, highest $0.055$); positive on 10, 10, and 10 of ten draws & Section~\ref{sec:rq1}; the margin is not an artifact of one training draw \\\addlinespace[2pt]
The classical margin over the stratified-random floor, resampling files & Deleting one file at a time gives the interval $[0.020, 0.044]$ (one seed), wider than the interval over partitions, which share files & Section~\ref{sec:split}, matched contest \\\addlinespace[2pt]
Both matched models retuned on held-out files & The classical minus quantum contrast is $+0.016 \pm 0.017$ over five partitions; retuning moved it by $+0.0009$ & Section~\ref{sec:split}; the contrast is not stale tuning \\\addlinespace[2pt]
A time-ordered split inside each scenario block (first 60\% of rows train, last 20\% test) & LightGBM scores $0.601$, against $0.594$ with whole files held out, so time order alone removes almost all of the leak & Section~\ref{sec:split} \\\addlinespace[2pt]
Is a released file one event? & No. Each file holds one unbroken run of each of its scenarios (553 runs in 15 files), and 86\% of rows have their nearest neighbor in the same file & Section~\ref{sec:data} \\\addlinespace[2pt]
A projected quantum kernel, tuned at the same budget & $0.559$ macro-F1 at the bandwidth validation selects ($0.571$ at the best of three on the test split), against $0.564$ for the fidelity kernel and $0.597$ for tuned XGBoost & Section~\ref{sec:rq1}; the null covers both kernel constructions \\\addlinespace[2pt]
How similar is the deployed quantum Gram matrix to the tuned RBF Gram matrix? & Their centered kernel alignment is $0.36$ & Section~\ref{sec:rq1} \\\addlinespace[2pt]
Which features carry the signal that survives on held-out files? & On held-out files, dropping the 48 voltage phasors costs XGBoost $0.039$ of its $0.069$ margin over the stratified-random floor. Those 48 alone give $0.053$, and 48 random features $0.059$ & Section~\ref{sec:levers}; the signal is spread over many channels \\\addlinespace[2pt]
One to nine training files at a fixed budget of 2000 rows, by learner & XGBoost gains $0.015$ macro-F1 and the MLP loses $0.065$, as each file then contributes fewer rows & Section~\ref{sec:levers}; the file-count reading is for the tree model \\\addlinespace[2pt]
\rowcolor{qgAttackTint}[0pt][0pt]\multicolumn{3}{@{}l@{}}{\textbf{Attack protocol}} \\
The worst case over ten restarts instead of their mean (transferred attack, RY-angle QSVM, one model seed, 500 test rows) & Retention falls from $0.513$ to $0.246$ for the RY-angle QSVM and from $0.583$ to $0.402$ for LightGBM & Section~\ref{sec:rq2}; a mean over restarts is not the worst case \\\addlinespace[2pt]
The QSVM's white-box result at $0.5\sigma$ across the regularization, and with sampled circuits (three attack seeds) & White-box retention is $0.138$ at $C = 1$ and $0.069$ at the tuned $C$. Trained and read at 1024 shots, the deployed model's clean macro-F1 falls from $0.563$ to $0.505$ (standard deviation $0.030$ over three sampling draws) & Sections~\ref{sec:rq2} and~\ref{sec:rq3} \\\addlinespace[2pt]
The radius at which damage is read (PGD curves of the first full robustness run, whose VQC predates its final configuration) & It moves tree numbers by up to $0.107$ under the score-based attack, where damage is not monotone in the radius, and no family's by more than $0.007$ under the shared-protocol PGD. The kernel machines' own white-box attacks move far more (Section~\ref{sec:rq2}) & Section~\ref{sec:rq2} \\\addlinespace[2pt]
The Square search against LightGBM at $0.5\sigma$ & It does $0.481$ macro-F1 of damage, $0.081$ more than natural evolution strategies at the same budget & Section~\ref{sec:rq2}; tree retentions are upper bounds \\\addlinespace[2pt]
White-box attack at $0.5\sigma$ with whole files held out (MLP, all 128 features) & Retention is $0.303$ (attacked macro-F1 $0.174$), so the MLP's white-box fragility holds without the leak & Sections~\ref{sec:preempt} and~\ref{sec:limitations} \\\addlinespace[2pt]
What an attacked kernel detector still catches (its own white-box attack at $0.5\sigma$, 200 test rows) & The deployed fidelity kernel catches $0.05$ of attacks ($0.71$ clean) while flagging 98\% of normal rows. The tuned RBF kernel catches $0.85$ and flags 97\% & Section~\ref{sec:rq2} \\\addlinespace[2pt]
\rowcolor{qgControlTint}[0pt][0pt]\multicolumn{3}{@{}l@{}}{\textbf{Physical consistency on the MSU/ORNL data}} \\
Does the transfer to the trees depend on which channels the attacker moves? & Moving only the 56 channels the attacker controls directly, with the derived channels frozen, keeps $0.95$ to $1.22$ of the full excess damage on the three tree models, so it does not & Section~\ref{sec:physics} \\\addlinespace[2pt]
The consistent four-relay attack with the budget in a physical voltage unit & The QSVM's white-box macro-F1 is $0.217$, $0.209$, and $0.270$ at 0.1, 0.25, and 0.5 of that unit, below the constant-predictor floor each time & Section~\ref{sec:physics} \\\addlinespace[2pt]
A relay-local consistency check on the sequence channels, at a 1\% false-alarm rate & It flags 100\% of the white-box attack rows that ignore the physics and 100\% of random rows of the same size, and at most 0.8\% of the physically consistent rows & Sections~\ref{sec:physics} and~\ref{sec:realizability} \\\addlinespace[2pt]
\rowcolor{qgReferenceTint}[0pt][0pt]\multicolumn{3}{@{}l@{}}{\textbf{Replication data and reproducibility}} \\
Gas: missing-value indicators in place of the median fill & No score moves by more than $0.0096$ with pressure log-scaled, or $0.0085$ under the main-study pipeline & Section~\ref{sec:replication} \\\addlinespace[2pt]
Gas: training inside the held-out time block (main-study pipeline) & It scores $0.563$, against $0.581$ for as many rows from the other blocks, the same direction as on the MSU/ORNL files & Section~\ref{sec:replication} \\\addlinespace[2pt]
Gas: the RBF kernel narrowed 100-fold (main-study pipeline) & Its white-box retention at $0.5\sigma$, the one radius this check ran, rises from $0.112$ to $0.263$. The MSU/ORNL RBF-SVM keeps $0.685$ at that radius, so the length scale accounts for part of that difference, not all of it; the MSU/ORNL reading at its worst radius is $0.243$ & Section~\ref{sec:replication} \\\addlinespace[2pt]
Gas: the Square search at $0.5\sigma$ & Every family falls to between $0.048$ and $0.245$ macro-F1 & Section~\ref{sec:replication} \\\addlinespace[2pt]
Simulated grids: the VQC's accuracy & $0.826 \pm 0.040$ (14-bus) and $0.702 \pm 0.015$ (118-bus) macro-F1, against $0.988$ and $0.993$ for the ZZ-map QSVM and $0.996$ and $0.997$ for the RBF-SVM & Section~\ref{sec:replication} \\\addlinespace[2pt]
XGBoost fitted with 1, 4, 10, and 24 threads & The ten-seed mean moves by $0.0012$ on all 128 features and $0.0011$ at PCA-8 & Reproducibility statement \\\addlinespace[2pt]
\bottomrule
\end{tabularx}

\end{table*}

\FloatBarrier

\section*{Acknowledgment}
During the preparation of this work, M. R. Islam used Claude (Anthropic)~\cite{anthropic2026claude}
in all sections to help organize the structure of the manuscript, to improve the clarity of
the writing, and to check references and formatting. The authors reviewed and edited all
content.

\bibliographystyle{IEEEtran}
\bibliography{refs}

\end{document}